\documentclass{article}

\usepackage{ifthen}

\usepackage{microtype}
\usepackage{graphicx}
\usepackage{subfigure}
\usepackage{booktabs}
\usepackage{listings}
\usepackage{hyperref}
\usepackage[frozencache]{minted}

\newboolean{useICML}
\setboolean{useICML}{true}

\ifthenelse{\boolean{useICML}}{
  \usepackage[accepted]{icml2024}
}{
  \usepackage{mathstyle}
}

\ifthenelse{\boolean{useICML}}{
}{
  \usepackage{natbib}
}

\usepackage{amsmath}
\usepackage{amssymb}
\usepackage{mathtools}
\usepackage{amsthm}
\usepackage[capitalise,noabbrev]{cleveref}

\usepackage{url}            
\usepackage{amsfonts}       
\usepackage{nicefrac}       
\usepackage{xcolor}         
\usepackage{enumitem}
\usepackage{graphics}

\newcommand{\papertitle}{On the Implicit Bias of Adam\internalComment{internal}}

\graphicspath{{pics/}}

\newboolean{isInternal}
\setboolean{isInternal}{false}

\ifthenelse{\boolean{isInternal}}{
  \newcommand{\internalComment}[1]{\textbf{\color{red}[}#1\textbf{\color{red}]}}
  \rfoot{\color{red} internal}
  \renewcommand{\headrulewidth}{0pt} 
  \renewcommand{\footrulewidth}{0pt} 
}{
  \newcommand{\internalComment}[1]{}
}

\newcounter{relctr} 
\everydisplay\expandafter{\the\everydisplay\setcounter{relctr}{0}} 

\newcommand\labelrel[2]{%
  \begingroup
  \refstepcounter{relctr}%
  \stackrel{\textnormal{(\alph{relctr})}}{\mathstrut{#1}}%
  \originallabel{#2}%
  \endgroup
}
\AtBeginDocument{\let\originallabel\label} 

\newcommand\MTkillspecial[1]{
  \bgroup
  \catcode`\&=9
  \let\\\relax%
  \scantokens{#1}%
  \egroup
}
\DeclarePairedDelimiter\multlineparen\lparen\rparen
\reDeclarePairedDelimiterInnerWrapper\multlineparen{star}{
  \mathopen{#1\vphantom{\MTkillspecial{#2}}\kern-\nulldelimiterspace\right.}
#2
\mathclose{\left.\kern-\nulldelimiterspace\vphantom{\MTkillspecial{#2}}#3}}

\DeclarePairedDelimiter\multlinebrace\{\}
\reDeclarePairedDelimiterInnerWrapper\multlinebrace{star}{
  \mathopen{#1\vphantom{\MTkillspecial{#2}}\kern-\nulldelimiterspace\right.}
#2
\mathclose{\left.\kern-\nulldelimiterspace\vphantom{\MTkillspecial{#2}}#3}}

\newcommand*\makeAlph[1]{\symbol{\numexpr96+#1}}
\NewDocumentCommand{\labrel}{m o }{%
  \IfNoValueTF{#2}{(\makeAlph{#1})}{\stackrel{\mathrm{(\makeAlph{#1})}}{#2}}%
}

\newcounter{ccnt}
\NewDocumentCommand{\newc}{o}{%
  \refstepcounter{ccnt}\ensuremath{c_{\theccnt}}\IfValueT{#1}{\label{#1}}%
}

\newcounter{bigccnt}
\NewDocumentCommand{\newbigc}{o}{%
  \refstepcounter{bigccnt}\ensuremath{C_{\thebigccnt}}\IfValueT{#1}{\label{#1}}%
}
\newcommand{\oldbigc}[1]{\ensuremath{C_{\ref{#1}}}}
\newcounter{dcnt}
\NewDocumentCommand{\newd}{o}{%
  \refstepcounter{dcnt}\ensuremath{d_{\thedcnt}}\IfValueT{#1}{\label{#1}}%
}
\newcommand{\oldd}[1]{\ensuremath{d_{\ref{#1}}}}
\newcounter{bigdcnt}
\NewDocumentCommand{\newbigd}{o}{%
  \refstepcounter{bigdcnt}\ensuremath{D_{\thebigdcnt}}\IfValueT{#1}{\label{#1}}%
}

\newcommand{\bv}{\mathbf{v}}
\newcommand{\bof}{\mathbf{f}}
\newcommand{\btheta}{\boldsymbol{\theta}}
\newcommand{\bdelta}{\boldsymbol{\delta}}

\makeatletter
\newcommand*\bigcdot{\mathpalette\bigcdot@{.5}}
\newcommand*\bigcdot@[2]{\mathbin{\vcenter{\hbox{\scalebox{#2}{$\m@th#1\bullet$}}}}}
\makeatother

\let\originalleft\left
\let\originalright\right
\renewcommand{\left}{\mathopen{}\mathclose\bgroup\originalleft}
\renewcommand{\right}{\aftergroup\egroup\originalright}

\DeclareMathOperator{\sign}{sign}
\newcommand{\param}{\btheta}
\newcommand{\modparam}{\tilde{\btheta}}
\newcommand{\paramsc}{\theta}
\newcommand{\loss}{E}
\newcommand{\modloss}{\widetilde{E}}
\newcommand{\trans}{^\intercal}
\newcommand{\timeder}[1]{\ensuremath\left(#1\right)^{\boldsymbol{\cdot}}}
\newcommand{\timederder}[1]{\ensuremath\left(#1\right)^{\boldsymbol{\cdot} \boldsymbol{\cdot}}}
\definecolor{highlight}{HTML}{6E4196}

\theoremstyle{plain}
\newtheorem{theorem}{Theorem}[section]
\newtheorem{result}{Result}[section]

\newtheorem{lemma}[theorem]{Lemma}

\theoremstyle{definition}
\newtheorem{definition}[theorem]{Definition}
\newtheorem{assumption}[theorem]{Assumption}
\newtheorem{example}[theorem]{Example}
\theoremstyle{remark}
\newtheorem{remark}[theorem]{Remark}

\ifthenelse{\boolean{useICML}}{
  \newcommand{\namePar}[1]{\textbf{#1}}  
}{
  \newcommand{\namePar}[1]{\paragraph{#1}}  
}

\ifthenelse{\boolean{useICML}}{}{%
  \title{\papertitle}

  \author{
    Matias D. Cattaneo\thanks{Equal contribution} \\
    Princeton University\\
    \texttt{cattaneo@princeton.edu} \\
    \and
    Jason M. Klusowski\footnotemark[1] \\
    Princeton University\\
    \texttt{jason.klusowski@princeton.edu} \\
    \and
    Boris Shigida\footnotemark[1] \\
    Princeton University\\
    \texttt{bs1624@princeton.edu} \\
  }
}

\allowdisplaybreaks[4]

\begin{document}

\ifthenelse{\boolean{useICML}}{
  \twocolumn[
  \icmltitle{\papertitle}
  \icmlsetsymbol{equal}{*}

  \begin{icmlauthorlist}
    \icmlauthor{Matias D. Cattaneo}{equal,orfe}
    \icmlauthor{Jason M. Klusowski}{equal,orfe}
    \icmlauthor{Boris Shigida}{equal,orfe}
  \end{icmlauthorlist}

  \icmlaffiliation{orfe}{Department of Operations Research and Financial Engineering, Princeton University, Princeton, NJ, USA}

  \icmlcorrespondingauthor{Boris Shigida}{bs1624@princeton.edu}

  \icmlkeywords{theory, optimization, implicit bias, generalization, backward error analysis, modified equations, Adam, adaptive methods, gradient descent, sharpness}

  \vskip 0.3in
  ]

  \printAffiliationsAndNotice{\icmlEqualContribution}
}{}

\begin{abstract}
In previous literature, backward error analysis was used to find ordinary differential equations (ODEs) approximating the gradient descent trajectory. It was found that finite step sizes implicitly regularize solutions because terms appearing in the ODEs penalize the two-norm of the loss gradients. We prove that the existence of similar implicit regularization in RMSProp and Adam depends on their hyperparameters and the training stage, but with a different ``norm'' involved: the corresponding ODE terms either penalize the (perturbed) one-norm of the loss gradients or, conversely, impede its reduction (the latter case being typical). We also conduct numerical experiments and discuss how the proven facts can influence generalization.
\end{abstract}

\section{Introduction}

Gradient descent (GD) can be seen as a numerical method solving the ordinary differential equation (ODE) $\dot{\param} = - \nabla \loss(\param)$, where $\loss(\cdot)$ is the loss function and $\nabla \loss(\param)$ is its gradient. Starting at $\param^{(0)}$, it creates a sequence of guesses $\param^{(1)}, \param^{(2)}, \ldots$, which lie close to the solution trajectory $\param(t)$ governed by the aforementioned ODE. Since the step size $h$ is finite, one could search for a modified differential equation $\dot{\modparam} = - \nabla \modloss(\modparam)$ such that $\param^{(n)} - \modparam(n h)$ is exactly zero, or at least closer to zero than $\param^{(n)} - \param(n h)$, that is, all the guesses of the descent lie exactly on the new solution curve or closer compared to the original curve. This approach to analysing properties of a numerical method is sometimes called backward error analysis in the numerical integration literature (see Chapter IX in~\citet{hairer2006} and references therein).

\citet{barrett2021implicit} used this idea for full-batch GD and found that the modified loss function $\modloss(\modparam) = \loss(\modparam) + (h / 4) \| \nabla \loss(\modparam) \|^2$ makes the trajectory of the solution to $\dot{\modparam} = - \nabla \modloss(\modparam)$ approximate the sequence $\{ \param^{(n)} \}_{n = 0}^{\infty}$ one order of $h$ better than the original ODE, where $\|\cdot\|$ is the Euclidean norm. In related work, \citet{miyagawa2022toward} obtained the correction term for full-batch GD up to any chosen order, also studying the global error (uniform in the iteration number) as opposed to the local (one-step) error.

The analysis was later extended to mini-batch GD in~\citet{smith2021on}. Assume that the training set is split into batches of size $B$ and there are $m$ batches per epoch (so the training set size is $m B$). The cost function is rewritten as $\loss(\param) = (1 / m) \sum_{k = 0}^{m - 1} \hat{\loss}_k(\param)$ with mini-batch costs denoted $\hat{\loss}_k(\param) = (1 / B) \sum_{j = k B + 1}^{k B + B} \loss_j(\param)$. It was obtained in that work that after one epoch, the mean iterate of the algorithm, averaged over all possible shuffles of the batch indices, is close to the solution to $\dot{\param} = - \nabla \modloss_{SGD}(\param)$, where the modified loss is given by $\modloss_{SGD}(\param) = \loss(\param) + h / (4 m)\cdot \sum_{k = 0}^{m - 1} \big\| \nabla \hat{\loss}(\param) \big\|^2$.

Modified equations have also been derived for GD with heavy-ball momentum $\param^{(n + 1)} = \param^{(n)} - h \nabla \loss ( \param^{(n)} ) + \beta ( \param^{(n)} - \param^{(n - 1)} )$, where $\beta$ is the momentum parameter. In the full-batch setting, it turns out that for $n$ large enough it is close to the continuous trajectory solving
\begin{equation}\label{eq:heavy-ball-regularization}
  \dot{\param} = - \frac{\nabla E(\param)}{1 - \beta} - \underbrace{h \frac{1 + \beta}{(1 - \beta)^3} \frac{\nabla \left\| \nabla E(\param) \right\|^2}{4}}_{\text{implicit regularization}}.
\end{equation}
Versions of this general result were proven in \citet{farazmand2020multiscale}, \citet{kovachki2021continuous}, \citet{ghosh2023implicit} under different assumptions. The focus of the latter work is the closest to ours since they interpret the correction term as implicit regularization. Their main theorem also provides the analysis for the general mini-batch case.

In another recent work, \citet{zhao2022penalizing} introduce a regularization term $\lambda \cdot \left\| \nabla \loss(\param) \right\|$ to the loss function as a way to ensure finding flatter minima, improving generalization. The only difference between their term and the first-order correction coming from backward error analysis (up to a coefficient) is that the norm is not squared and regularization is applied on a per-batch basis.

The application of backward error analysis for approximating the discrete dynamics of adaptive algorithms such as RMSProp~\citep{tieleman2012lecture} and Adam~\citep{kingma2015adam} is currently missing in the literature. \citet{barrett2021implicit} note that ``it would be interesting to use backward error analysis to calculate the modified loss and implicit regularization for other widely
used optimizers such as momentum, Adam and RMSprop''. \citet{smith2021on} reiterate that they ``anticipate that backward error analysis could also be used to clarify the role of finite learning rates in adaptive optimizers like Adam''. \citet{ghosh2023implicit} agree that ``RMSProp ... and Adam ...,  albeit being powerful alternatives to SGD
with faster convergence rates, are far from well-understood in the aspect of implicit regularization''. In a similar context, in Appendix G to~\citet{miyagawa2022toward} it is mentioned that ``its [Adam's] counter term and discretization error are open questions''.

This work fills the gap by conducting backward error analysis for (mini-batch, and full-batch as a special case) Adam and RMSProp. Our main contributions are listed below.

\begin{itemize}[leftmargin=*]
\item In \cref{th:backward-error-analysis-modified-adam}, we provide a global second-order in $h$ continuous piecewise ODE approximation to Adam in the general mini-batch setting. (A similar result for RMSProp is moved to \cref{sec:appendix-about-rmsprop-with-eps-inside}.) For the full-batch special case, it was shown in prior work \citet{ma2022qualitative} that the continuous-time limit of both these algorithms is a (perturbed by the numerical stability parameter $\varepsilon$) signGD flow $\dot{\param} = - \nabla \loss(\param) / (\left| \nabla \loss(\param) \right| + \varepsilon)$
component-wise; we make this more precise by finding a linear in $h$ correction term on the right.

\item We analyze the full-batch case in the context of regularization (see the summary in \cref{sec:summary}). In contrast to the case of GD, where the two-norm of the loss gradient is implicitly penalized, Adam typically \textit{anti}-penalizes the perturbed one-norm of the loss gradient $\| \bv \|_{1, \varepsilon} = \sum_{i = 1}^p \sqrt{v_i^2 + \varepsilon}$ (i.\,e., penalizes the negative norm), as specified in~\eqref{eq:informal-summary-ode-for-small-eps}. Thus, the implicit bias of Adam that we identify serves as
\textit{anti-regularization} (except for the unusual case $\beta \geq \rho$, large $\varepsilon$ or very late at training).

\item We provide numerical evidence consistent with our theoretical results by training various vision models on CIFAR-10 using full-batch Adam. In particular, we observe that the stronger the implicit anti-regularization effect predicted by our theory, the worse the generalization. This pattern holds across different architectures: ResNets, simple convolutional neural networks (CNNs) and Vision Transformers. Thus, we propose a novel possible explanation for often-reported poor generalization of adaptive gradient algorithms. The code used for training the models is available at \url{https://github.com/borshigida/implicit-bias-of-adam}.

\end{itemize}

\subsection{Related Work}\label{sec:related-work}

\namePar{Backward error analysis of first-order methods.} We outlined the history of finding ODEs approximating different algorithms above in the introduction. Recently, there have been other applications of backward error analysis related to machine learning. \citet{kunin2020neural} show that the approximating continuous-time trajectories satisfy conservation laws that are broken in discrete time. \citet{francca2021dissipative} use backward error analysis while studying how to discretize continuous-time dynamical systems preserving stability and convergence rates. \citet{rosca2021discretization} find continuous-time approximations of discrete two-player differential games.

\namePar{Approximating gradient methods by differential equation trajectories.} Under the assumption that the hyperparameters $\beta, \rho$ of the Adam algorithm (see \cref{def:adam}) tend to 1 at a certain rate as $h \to 0$, a first-order continuous ODE approximation to this algorithm was derived in \citet{barakat2021convergence}. On the other hand, if $\beta, \rho$ are kept fixed, \citet{ma2022qualitative} prove that the trajectories of Adam and RMSProp are close to signGD dynamics, and investigate different training regimes of these algorithms empirically. SGD is approximated by stochastic differential equations and novel adaptive parameter adjustment policies are devised in~\citet{pmlr-v70-li17f}. \citet{malladi2022sdes} derive stochastic differential equations that are order-1 weak approximations of RMSProp and Adam. We go in a different direction: instead of clarifying the previously obtained continuous ODE approximations by taking gradient noise into account, we take a deterministic approach but go one order of $h$ further. In particular, we keep $\beta$, $\rho$ fixed (thus generalizing the analysis for SGD with momentum), whereas \citet{malladi2022sdes} take $\beta, \rho \to 1$.

\namePar{Connection with signGD.} The connection of adaptive gradient methods with sign(S)GD is extensively discussed in \citet{bernstein2018signsgd}. \citet{balles2020geometry} study a version of signGD with an update proportional to $- \|\nabla E(\param)\|_1 \sign{\nabla E(\param)}$ as a special case of steepest descent, and discuss when sign-based methods are preferable to GD.

\namePar{Implicit bias of first-order methods.} \citet{soudry2018implicit} prove that GD trained to classify linearly separable data with logistic loss converges to the direction of the max-margin vector (the solution to the hard margin SVM). This result has been extended to different loss functions in \citet{nacson2019convergence}, to SGD in \citet{nacson2019stochastic}, AdaGrad in \citet{qian2019implicit}, (S)GD with momentum, deterministic Adam and stochastic RMSProp in \citet{wang2022does}, more generic optimization methods in \citet{gunasekar2018characterizing}, to the nonseparable case in \citet{ji2018risk}, \citet{ji2019implicit}. This line of research has been generalized to studying implicit biases of linear networks \citep{ji2018gradient, gunasekar2018implicit}, homogeneous neural networks \citep{ji2020directional, nacson2019lexicographic, lyu2019gradient}. \citet{woodworth2020kernel} study the gradient flow of a diagonal linear network with squared loss and show that large initializations lead to minimum two-norm solutions while small initializations lead to minimum one-norm solutions. \citet{even2023s} extend this work to the case of non-zero step sizes and mini-batch training. \citet{pmlr-v139-wang21q} prove that Adam and RMSProp maximize the margin of homogeneous neural networks. Our perspective on the implicit bias is different since we are considering a generic loss function without any assumptions on the network architecture. \citet{beneventano2023trajectories} proves that in expectation over batch sampling the trajectory of SGD without replacement differs from that of SGD with replacement by an additional step on a regularizer. As opposed to the work on backward error analysis for SGD discussed above, they do not assume the largest eigenvalue of the hessian to be bounded.

\namePar{Generalization of adaptive methods.} \citet{cohen2022adaptive} investigate the edge-of-stability regime of adaptive gradient algorithms and the effect of sharpness (the largest eigenvalue of the hessian) on generalization. \citet{granziol2020flatness,chen2021vision} observe that adaptive methods find sharper minima than SGD and \citet{zhou2020towards,xie2022adaptive} argue theoretically that it is the case. \citet{jiang2022does} introduce a statistic that measures the uniformity of the hessian diagonal and argue that adaptive gradient algorithms are biased towards making this statistic smaller. \citet{keskar2017improving} propose to improve generalization of adaptive methods by switching to SGD in the middle of training.

\subsection{Notation}

We denote the loss of the $k$th minibatch as a function of the network parameters $\param \in \mathbb{R}^p$ by $E_k(\param)$, and in the full-batch setting we omit the index and write $E(\param)$. $\nabla E$ means the gradient of $E$, and $\nabla$ with indices denotes partial derivatives, e.\,g. $\nabla_{i j s} E$ is a shortcut for $\frac{\partial^3 E}{\partial \paramsc_i \partial \paramsc_j \partial \paramsc_s}$. The norm notation without indices $\left\| \cdot \right\|$ is the two-norm of a vector, $\left\| \cdot \right\|_1$ is the one-norm and $\left\| \cdot \right\|_{1, \varepsilon}$ is the perturbed one-norm defined as $\left\| \bv \right\|_{1, \varepsilon} = \sum_{i = 1}^p \sqrt{v_i^2 + \varepsilon}$. (Of course, if $\varepsilon > 0$ the perturbed one-norm is not really a norm, but taking $\varepsilon = 0$ makes it the one-norm.) For a real number $a$ the floor $\lfloor a \rfloor$ is the largest integer not exceeding $a$.

To provide the names and notations for hyperparameters, we define the algorithm below.

\begin{definition}\label{def:adam}
  The \emph{Adam} algorithm \citep{kingma2015adam} is an optimization algorithm with numerical stability hyperparameter $\varepsilon > 0$, squared gradient momentum hyperparameter $\rho \in (0, 1)$, gradient momentum hyperparameter $\beta \in (0, 1)$, initialization $\param^{(0)} \in \mathbb{R}^p$, $\boldsymbol{\nu}^{(0)} = \mathbf{0} \in \mathbb{R}^p$, $\mathbf{m}^{(0)} = \mathbf{0} \in \mathbb{R}^p$ and the following update rule: for each $n \geq 0$, $j \in \left\{ 1, \ldots, p \right\}$
  \begin{align*}
    \nu_j^{(n + 1)} &= \rho \nu_j^{(n)} + (1 - \rho) \big( \nabla_j E_n(\param^{(n)}) \big)^2,\\
    m_j^{(n + 1)} &= \beta m_j^{(n)} + (1 - \beta) \nabla_j E_n ( \param^{(n)} ),\\
    \paramsc_j^{(n + 1)} &= \paramsc_j^{(n)} - h \frac{m_j^{(n + 1)} / (1 - \beta^{n + 1})}{\sqrt{\nu_j^{(n + 1)} / ( 1 - \rho^{n + 1}) + \varepsilon}}.
  \end{align*}
\end{definition}

\begin{remark}
Note that the numerical stability hyperparameter $\varepsilon > 0$, which is introduced in these algorithms to avoid division by zero, is inside the square root in our definition. This way we avoid division by zero in the derivative too: the first derivative of $x \mapsto \left(\sqrt{x + \varepsilon}\right)^{-1}$ is bounded for $x \geq 0$. This is useful for our analysis. In \cref{th:global-error-bound,th:adam-global-error-bound}, the original versions of RMSProp and Adam are also tackled, though with an additional assumption which requires that no component of the gradient can come very close to zero in the region of interest. This is true only for the initial period of learning (whereas \cref{th:backward-error-analysis-modified-adam} tackles the whole period). Practitioners do not seem to make a distinction between the version with $\varepsilon$ inside vs. outside the square root: tutorials with both versions abound on machine learning related websites. Moreover, the popular \href{https://github.com/keras-team/keras/blob/f9336cc5114b4a9429a242deb264b707379646b7/keras/optimizers/rmsprop.py\#L190}{Tensorflow} and \href{https://github.com/google-deepmind/optax/blob/8a3ee74de2d58e027c5df91a4d797b3c64a012cd/optax/_src/transform.py\#L252}{Optax} variants of RMSProp have $\varepsilon$ inside the square root. Empirically we also observed that moving $\varepsilon$ inside or outside the square root does not change the behavior of Adam or RMSProp qualitatively.
\end{remark}

\section{Implicit Bias of Full-Batch Adam: an Informal Summary}\label{sec:summary}

We are ready to describe our theoretical result (\cref{th:backward-error-analysis-modified-adam} below) in the full-batch special case. Assume $E(\param)$ is the loss, whose partial derivatives up to the fourth order are bounded. Let $\{\param^{(n)}\}$ be iterations of Adam as defined in~\cref{def:adam}. We find an ODE whose solution trajectory $\tilde{\param}(t)$ is $h^2$-close to $\{\param^{(n)}\}$, meaning that for any time horizon $T > 0$ there is a constant $C$ such that for any step size $h \in (0, T)$ we have $\| \tilde{\param}(n h) - \param^{(n)} \| \leq C h^2$ (for $n$ between $0$ and $\lfloor T / h \rfloor$). The ODE is written the following way (up to terms that rapidly go to zero as $n$ grows): for the component number $j \in \{ 1, \ldots, p \}$
\begin{equation}\label{eq:informal-summary-ode-def}
  \dot{\tilde{\paramsc}}_j(t) = - \frac{\nabla_j E\bigl(\tilde{\param}(t)\bigr) + \text{correction}_j\bigl(\tilde{\param}(t)\bigr)}{\sqrt{\bigl| \nabla_j E\bigl(\tilde{\param}(t)\bigr)\bigr|^2 + \varepsilon}}
\end{equation}
with initial conditions $\tilde{\param}_j(0) = \param_j^{(0)}$ for all $j$, where the correction term is
\begin{multline}\label{eq:bias}
  \text{correction}_j(\param)\\
  := \frac{h}{2} \biggl\{ \frac{1 + \beta}{1 - \beta} - \frac{1 + \rho}{1 - \rho} + \frac{1 + \rho}{1 - \rho} \cdot \frac{\varepsilon}{| \nabla_j E(\param)|^2 + \varepsilon} \biggr\}\\
  \times \nabla_j \big\| \nabla E(\param) \big\|_{1, \varepsilon}.
\end{multline}
Depending on hyperparameters and the training stage, the correction term can take two extreme forms listed below. The reality is in between, but typically much closer to the first case.

\begin{itemize}[leftmargin=*]
    \item If $\sqrt{\varepsilon}$ is \textbf{small} compared to all components of $\nabla E\bigl(\tilde{\param}(t)\bigr)$, i.\,e. $\min_{j}\bigl| \nabla_j E\bigl(\tilde{\param}(t)\bigr) \bigr| \gg \sqrt{\varepsilon}$, which \textbf{is usually the case during most of the training}, then we can write
      \begin{equation}\label{eq:bias-if-eps-is-small}
        \text{correction}_j(\param) \approx \frac{h}{2} \left\{ \frac{1 + \beta}{1 - \beta} - \frac{1 + \rho}{1 - \rho} \right\} \nabla_j \| \nabla E(\param) \|_{1, \varepsilon}.
      \end{equation}
For small $\varepsilon$, the perturbed one-norm is indistinguishable from the usual one-norm, and for $\beta > \rho$ it is penalized (in much the same way as the squared two-norm is implicitly penalized in the case of GD), but for the typical case $\rho > \beta$ its decrease is actually hindered by this term (so the bias is \textit{anti}-regularization). The ODE in~\eqref{eq:informal-summary-ode-def} approximately becomes
\begin{equation}\label{eq:informal-summary-ode-for-small-eps}
  \begin{aligned}
    &\dot{\tilde{\paramsc}}_j(t) = - \frac{\nabla_j \widetilde{E}\bigl(\tilde{\param}(t)\bigr)}{\big| \nabla_j E\bigl(\tilde{\param}(t)\bigr) \big|},\quad\text{with}\\
    &\widetilde{E}(\param) = E(\param) + \underbrace{\frac{h}{2} \left\{ \frac{1 + \beta}{1 - \beta} - \frac{1 + \rho}{1 - \rho} \right\} \| \nabla E(\param) \|_{1}}_{\text{implicit \textbf{anti-regularization} (if $\rho > \beta$)}}.
  \end{aligned}
\end{equation}

    \item If $\sqrt{\varepsilon}$ is \textbf{large} compared to all gradient components, i.\,e. $\max_{j}\big| \nabla_j E\bigl(\tilde{\param}(t)\bigr) \big| \ll \sqrt{\varepsilon}$ (which may happen during the late learning stage, or if non-standard hyperparameter values are chosen), the fraction in~\eqref{eq:bias} with $\varepsilon$ in the numerator approaches one, the dependence on $\rho$ cancels out, and
      \begin{multline}\label{eq:perturbed-one-norm-linearization}
          \big\| \nabla E\bigl(\tilde{\param}(t)\bigr) \big\|_{1, \varepsilon} \approx \sum_{i = 1}^p \sqrt{\varepsilon} \bigl( 1 + \bigl| \nabla_i E\bigl(\tilde{\param}(t)\bigr) \bigr|^2 / (2 \varepsilon) \bigr)\\
          = p \sqrt{\varepsilon} + \frac{1}{2 \sqrt{\varepsilon}} \bigl\|  \nabla E\bigl(\tilde{\param}(t)\bigr) \bigr\|^2.
      \end{multline}
In other words, $\| \cdot \|_{1, \varepsilon}$ becomes $\| \cdot \|^2 / (2 \sqrt{\varepsilon})$
up to an additive constant, giving
\begin{align*}
  &\text{correction}_j(\param)\\
  &\quad \approx \bigl(4 \sqrt{\varepsilon}\bigr)^{-1} (1 - \beta)^{-1} (1 + \beta) \nabla_j \| \nabla E(\param) \|^2.
\end{align*}
The form of the ODE in this case is
\begin{equation}\label{eq:ode-form-if-eps-is-large}
  \begin{aligned}
    &\dot{\tilde{\paramsc}}_j(t) = -\nabla_j \widetilde{E}\bigl(\tilde{\param}(t)\bigr),\quad\text{with}\\
    &\widetilde{E}(\param) = \frac{1}{\sqrt{\varepsilon}} \biggl( E(\param) + \frac{h}{4 \sqrt{\varepsilon}} \frac{1 + \beta}{1 - \beta} \| \nabla E(\param) \|^2 \biggr).
  \end{aligned}
\end{equation}
\end{itemize}

\begin{figure}[htb!]
  \centerline{\includegraphics[width=\columnwidth]{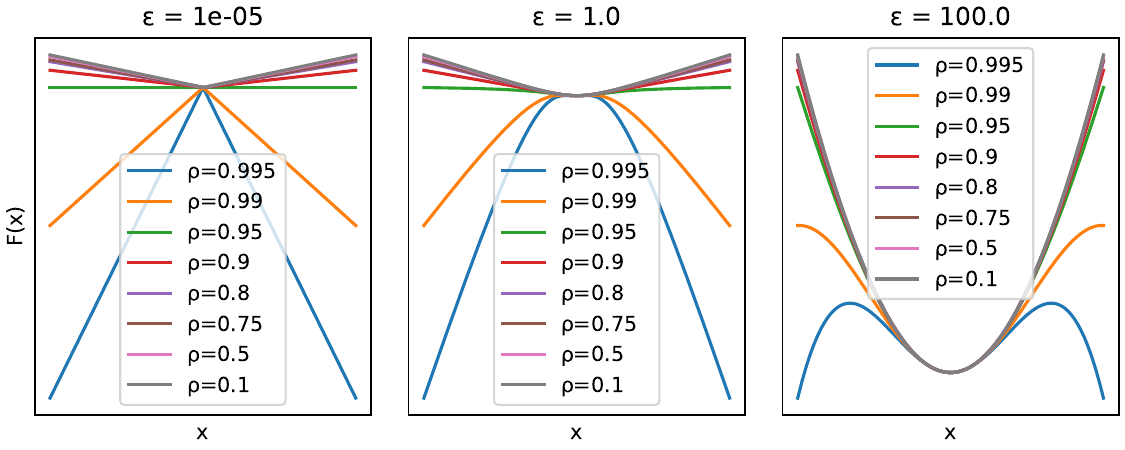}}
  \caption{\label{fig:illustration-what-is-penalized} To illustrate what term is being implicitly penalized in the simple case $p = 1$, we plot the graphs of $x \mapsto F(x) := \frac{h}{2} \int_0^x \bigl\{ \frac{1 + \beta}{1 - \beta} - \frac{1 + \rho}{1 - \rho} + \frac{1 + \rho}{1 - \rho} \cdot \frac{\varepsilon}{y^2 + \varepsilon} \bigr\}\,\mathrm{d}\sqrt{\varepsilon + y^2}$ with $\beta = 0.95$. In this case, the correction term in~\eqref{eq:bias} is itself the gradient of the function $F(E'(\paramsc))$, where $E'$ is the derivative (=gradient) of the loss: specifically, $\text{correction} = \frac{\mathrm{d}}{\mathrm{d} \paramsc} F(E'(\paramsc))$. Hence, Adam's iteration penalizes $F(E'(\paramsc))$. If $\varepsilon$ is small and $\rho > \beta$, the \textit{negative} one-norm of the gradient is penalized (leftmost picture, highest values of $\rho$); in other words, the one-norm is \textit{anti}-penalized.}
\end{figure}

These two extreme cases are summarized in \cref{table:Implicit Bias of Adam: Special Cases}. In \cref{fig:illustration-what-is-penalized}, we use the one-dimensional ($p = 1$) case to illustrate what kind of term is being implicitly penalized.

\begin{table}[h]
\caption{Implicit bias of Adam: special cases. ``Small'' and ``large'' are in relation to squared gradient components (Adam in the latter case is close to GD with momentum).}
\label{table:Implicit Bias of Adam: Special Cases}
\vskip 0.15in
\begin{center}
  \begin{small}
    \begin{tabular}{|c|c|c|}
      \hline
      & $\varepsilon$ ``small'' & $\varepsilon$ ``large'' \\
      \hline
      $\rho > \beta$ & \textbf{$-\| \nabla E(\param) \|_{1}$-penalized} & $\| \nabla E(\param) \|_{2}^2$-penalized \\
      \hline
      $\beta \geq \rho$ & $\| \nabla E(\param) \|_{1}$-penalized & $\| \nabla E(\param) \|_{2}^2$-penalized \\
      \hline
    \end{tabular}
  \end{small}
\end{center}
\vskip -0.1in
\end{table}

Usually, $\varepsilon$ is chosen to be small, and during most of the training Adam is much better described by the first extreme case.
It is clear from \eqref{eq:informal-summary-ode-for-small-eps} that, if $\rho > \beta$, the correction term provides the opposite of regularization, in contrast to~\eqref{eq:heavy-ball-regularization}. The larger $\rho$ compared to $\beta$, the stronger the anti-regularization effect is.

This finding may partially explain why adaptive gradient methods have been reported to generalize worse than non-adaptive ones \citep{chen2018closing,wilson2017marginal},
as it offers a previously unknown perspective on why they are biased towards ``higher-curvature'' regions and find ``sharper'' minima.
Indeed, note that standard (non-adaptive) $\ell_{\infty}$-sharpness at $\param$ can be defined by $\max_{\|\bdelta\|_{\infty} \leq r} E(\param + \bdelta) - E(\param)$ for some radius $r$.
This or similar definitions have been considered often in literature, see, e.\,g., \citet{andriushchenko2023ModernLook}, \citet{foret2021sharpnessaware}. Replacing the difference of the losses with its first-order approximation under the maximum \citep{foret2021sharpnessaware,ghosh2023implicit}
\begin{multline*}
  \max_{\|\bdelta\|_{\infty} \leq r} E(\param + \bdelta) - E(\param)\\
  \approx \max_{\|\bdelta\|_{\infty} \leq r} \nabla E(\param)\trans \bdelta = r \|\nabla E(\param)\|_1,
\end{multline*}
we see that Adam typically anti-penalizes the approximation of $\ell_{\infty}$-sharpness.
Although the connection between sharpness and generalization is not clear-cut \citep{andriushchenko2023ModernLook}, our empirical results (\cref{sec:numerical})
are consistent with this theory.

This overview also applies to RMSProp by setting $\beta = 0$; see \cref{th:mod-rmsprop-global-error-bound} for the formal result.

\begin{example}[Backward Error Analysis for GD with Heavy-ball Momentum]\label{ex:heavy-ball}
    Assume $\varepsilon$ is large compared to all squared gradient components during the whole training process, so that the form of the ODE is approximated by~\eqref{eq:ode-form-if-eps-is-large}. Since Adam with a large $\varepsilon$ and after a certain number of iterations approximates SGD with heavy-ball momentum with step size $h (1 - \beta) / \sqrt{\varepsilon}$, a linear step size change (and corresponding time change) gives exactly the equations in Theorem~4.1 of~\citet{ghosh2023implicit}. Taking $\beta = 0$ (no momentum), we get the implicit regularization of GD from~\citet{barrett2021implicit}.
\end{example}

\section{Main Result: ODE Approximating Mini-Batch Adam}\label{sec:main-result}

We only make one assumption, which is standard in the literature: the loss $E_k$ for each mini-batch is 4 times continuously differentiable, and partial derivatives of $E_k$ up to order 4 are bounded, i.\,e. there is a positive constant $M$ such that for $\param$ in the region of interest
\begin{multline}\label{eq:bounded-smoothness-assumption}
  \sup_k \biggl\{ \sup_{i} | \nabla_{i} E_k(\param) | \vee \sup_{i, j} | \nabla_{i j} E_k(\param) |\\
  \vee \sup_{i, j, s} | \nabla_{i j s} E_k(\param) | \vee \sup_{i, j, s, r} | \nabla_{i j s r} E_k(\param) | \biggr\} \leq M.
\end{multline}

\begin{theorem}\label{th:backward-error-analysis-modified-adam}
Assume~\eqref{eq:bounded-smoothness-assumption} holds. Let $\{\param^{(n)}\}$ be iterations of Adam as defined in~\cref{def:adam}, $\tilde{\param}(t)$ be the continuous solution to the piecewise ODE
\begin{multline}\label{eq:modified-adam-nth-step-modified-equation}
  \dot{\tilde{\paramsc}}_j(t) = - \frac{M_j^{(n)}\bigl(\tilde{\param}(t))}{R^{(n)}_j(\tilde{\param}(t)\bigr)}\\
  + h \biggl( \frac{M_j^{(n)}\bigl(\tilde{\param}(t)\bigr)\big(2 P_j^{(n)}\bigl(\tilde{\param}(t)\bigr) + \bar{P}_j^{(n)}\bigl(\tilde{\param}(t)\bigr)\big)}{2 R_j^{(n)}\bigl(\tilde{\param}(t)\bigr)^3}\\
  - \frac{2 L_j^{(n)}\bigl(\tilde{\param}(t)\bigr) + \bar{L}_j^{(n)}\bigl(\tilde{\param}(t)\bigr)}{2 R_j^{(n)}\bigl(\tilde{\param}(t)\bigr)} \biggr)
\end{multline}
for $t \in [n h, (n + 1) h]$ with the initial condition $\tilde{\param}(0) = \param^{(0)}$, where
\begin{align*}
  &R^{(n)}_j(\param) := \sqrt{\frac{\sum_{k = 0}^n \rho^{n - k} (1 - \rho) \left( \nabla_j E_k(\param) \right)^2}{1 - \rho^{n + 1}} + \varepsilon},\\
    &M^{(n)}_j(\param) := \frac{\sum_{k = 0}^n \beta^{n - k} \left(1 - \beta\right) \nabla_j E_k \left( \param \right)}{1 - \beta^{n + 1}},\\
    &L_j^{(n)}(\param) := \frac{1}{1 - \beta^{n + 1}} \sum_{k = 0}^n \beta^{n - k} (1 - \beta)\\
    &\qquad \times \sum_{i = 1}^p \nabla_{i j} E_k(\param) \sum_{l = k}^{n - 1} \frac{M^{(l)}_i(\param)}{R_i^{(l)}(\param)},\\
    &\bar{L}_j^{(n)}(\param) := \frac{1}{1 - \beta^{n + 1}} \sum_{k = 0}^n \beta^{n - k} (1 - \beta)\\
    &\qquad \times \sum_{i = 1}^p \nabla_{i j} E_k(\param) \frac{M^{(n)}_i(\param)}{R_i^{(n)}(\param)},\\
    &P_j^{(n)}(\param) := \frac{1}{1 - \rho^{n + 1}} \sum_{k = 0}^n \rho^{n - k} (1 - \rho) \nabla_j E_k(\param)\\
    &\qquad \times \sum_{i = 1}^p \nabla_{i j} E_k(\param) \sum_{l = k}^{n - 1} \frac{M^{(l)}_i(\param)}{R_i^{(l)}(\param)},\\
    &\bar{P}_j^{(n)}(\param) := \frac{1}{1 - \rho^{n + 1}} \sum_{k = 0}^n \rho^{n - k} (1 - \rho) \nabla_j E_k(\param)\\
    &\qquad \times \sum_{i = 1}^p \nabla_{i j} E_k(\param) \frac{M^{(n)}_i(\param)}{R_i^{(n)}(\param)}.
  \end{align*}
  Then, for any fixed positive time horizon $T > 0$ there exists a constant $C$ (depending on $T$, $\rho$, $\beta$, $\varepsilon$) such that for any step size $h \in (0, T)$ we have $\big\| \tilde{\param}(n h) - \param^{(n)} \big\| \leq C h^2$ for $n \in \left\{0, \ldots, \lfloor T / h \rfloor\right\}$.
\end{theorem}

\begin{remark}
In the \textit{full-batch} setting $E_k \equiv E$, the terms above simplify to
\begin{align*}
  &R_j^{(n)}(\param) = (| \nabla_j E(\param) |^2 + \varepsilon)^{1 / 2},\\
  &M^{(n)}_j(\param) = \nabla_j E(\param),\\
  &L_j^{(n)}(\param) = \left[ \frac{\beta}{1 - \beta} - \frac{(n + 1) \beta^{n + 1}}{1 - \beta^{n + 1}} \right] \bar{L}_j^{(n)}(\param),\\
  &\bar{L}_j^{(n)}(\param) = \nabla_j \| \nabla E(\param) \|_{1, \varepsilon},\\
  &P_j^{(n)}(\param) = \left[ \frac{\rho}{1 - \rho} - \frac{(n + 1) \rho^{n + 1}}{1 - \rho^{n + 1}} \right] \bar{P}_j^{(n)}(\param),\\
  &\bar{P}_j^{(n)}(\param) = \nabla_j E(\param) \nabla_j \| \nabla E(\param) \|_{1, \varepsilon}.
\end{align*}
If the iteration number $n$ is large, \eqref{eq:modified-adam-nth-step-modified-equation} rapidly becomes as described in~\eqref{eq:informal-summary-ode-def} and~\eqref{eq:bias}.
\end{remark}

\begin{proof}[Derivation sketch]
The proof is in the appendix (this is \cref{th:mod-adam-inside-global-error-bound}; see \cref{sec:appendix-overview} for the overview of the appendix). To help the reader understand how the ODE~\eqref{eq:modified-adam-nth-step-modified-equation} is obtained, apart from the full proof, we include an informal derivation in \cref{sec:informal-derivation}, and provide an even briefer sketch of this derivation here.

Our goal is to find such a trajectory $\tilde{\param}(t)$ that, denoting $t_n := n h$, we have
\begin{align}
  &\tilde{\paramsc}_j(t_{n + 1}) = \tilde{\paramsc}_j(t_n) - h \frac{T_{\beta, j}^{(n)}}{\sqrt{T_{\rho, j}^{(n)}}} + O(h^3)\quad\text{with}\label{eq:derivation-goal}\\
  &T_{\beta, j}^{(n)} := \frac{1}{1 - \beta^{n + 1}} \sum_{k = 0}^n \beta^{n - k} \left(1 - \beta\right) \nabla_j E_k \bigl( \tilde{\param}(t_k) \bigr),\nonumber\\
  &T_{\rho, j}^{(n)} := \frac{1}{1 - \rho^{n + 1}} \sum_{k = 0}^n \rho^{n - k} (1 - \rho) \bigl( \nabla_j E_k \bigl( \tilde{\param}(t_k) \bigr) \bigr)^2 + \varepsilon.\nonumber
\end{align}
Ignoring the terms of order higher than one in $h$, we can take a first-order approximation for granted: $\tilde{\paramsc}_j(t_{n + 1}) = \tilde{\paramsc}_j(t_n) - h A^{(n)}_j(\tilde{\param}(t_n)) + O( h^2)$ with $A^{(n)}_j(\param) := M_j^{(n)}(\param) / R_j^{(n)}(\param)$.
The challenge is to make this more precise by finding an equality of the form
\begin{multline}\label{eq:second-order-iteration-form-with-a-and-b}
    \tilde{\paramsc}_j\left(t_{n + 1}\right) = \tilde{\paramsc}_j(t_n)\\
    - h A^{(n)}_j\bigl(\tilde{\param}(t_n)\bigr) + h^2 B^{(n)}_j\bigl(\tilde{\param}(t_n)\bigr) + O(h^3),
\end{multline}
where $B^{(n)}_j(\cdot)$ is a known function. This is a numerical iteration to which standard backward error analysis (Chapter IX in~\citet{hairer2006}) can be applied.

Using the Taylor series, we can write
\begin{align*}
  &\nabla_j E_k \bigl( \tilde{\param}(t_{n - 1}) \bigr) = \nabla_j E_k \bigl( \tilde{\param}(t_n) \bigr)\\
  &\qquad + \sum_{i = 1}^p \nabla_{i j} E_k \bigl( \tilde{\param}(t_n) \bigr) \bigl\{ \tilde{\paramsc}_i(t_{n - 1}) - \tilde{\paramsc}_i(t_{n}) \bigr\} + O ( h^2 )\\
  &\quad = \nabla_j E_k \bigl( \tilde{\param}(t_n) \bigr)\\
  &\qquad + h \sum_{i = 1}^p \nabla_{i j} E_k \bigl( \tilde{\param}(t_n) \bigr) \frac{M_i^{(n - 1)} \bigl( \tilde{\param}(t_{n - 1}) \bigr)}{R_i^{(n - 1)}\bigl(\tilde{\param}(t_{n - 1})\bigr)} + O(h^2)\\
  &\quad = \nabla_j E_k \bigl( \tilde{\param}(t_n) \bigr)\\
  &\qquad + h \sum_{i = 1}^p \nabla_{i j} E_k \bigl( \tilde{\param}(t_n) \bigr) \frac{M_i^{(n - 1)} \bigl( \tilde{\param}(t_{n}) \bigr)}{R_i^{(n - 1)}\bigl(\tilde{\param}(t_{n})\bigr)} + O( h^2),
\end{align*}
where in the last equality we just replaced $t_{n - 1}$ with $t_n$ in the $h$-term since it only affects higher-order terms. Doing this again for steps $n - 2$, $n - 3$, $\ldots$, and adding the resulting equations will give for $k < n$
\begin{multline*}
  \nabla_j E_k \bigl( \tilde{\param}(t_k) \bigr) = \nabla_j E_k \bigl( \tilde{\param}(t_n) \bigr)\\
  + h \sum_{i = 1}^p \nabla_{i j} E_k \bigl( \tilde{\param}(t_n) \bigr) \sum_{l = k}^{n - 1} \frac{M_i^{(l)} \bigl( \tilde{\param}(t_n) \bigr)}{R_i^{(l)} \bigl( \tilde{\param}(t_n) \bigr)} + O( h^2),
\end{multline*}
where we could safely ignore that $n - k$ is not bounded because of exponential averaging. Taking the square of this formal power series in $h$, multiplying this square by $\rho^{n - k} (1 - \rho)$ and summing up over $k$ will give
\begin{align*}
  &\frac{1}{1 - \rho^{n + 1}} \sum_{k = 0}^n \rho^{n - k} (1 - \rho) \bigl[\nabla_j E_k \bigl( \tilde{\param}(t_k) \bigr) \bigr]^2 + \varepsilon\\
  &\quad = R_j^{(n)}\bigl(\tilde{\param}(t_n)\bigr)^2 + 2 h P_j^{(n)}\bigl(\tilde{\param}(t_n)\bigr) + O(h^2),
\end{align*}
which, using the expression for the inverse square root $\bigl(\sum_{r = 0}^\infty a_r h^r\bigr)^{- 1 / 2}$ of a formal power series $\sum_{r = 0}^\infty a_r h^r$, gives us an expansion
\begin{equation*}
  \frac{1}{\sqrt{T^{(n)}_{\rho, j}}} = \frac{1}{R_j^{(n)}\bigl(\tilde{\param}(t_n)\bigr)} - h \frac{P_j^{(n)}\bigl(\tilde{\param}(t_n)\bigr)}{R_j^{(n)}\bigl(\tilde{\param}(t_n)\bigr)^3} + O ( h^2 ).
\end{equation*}
A similar process provides an expansion for $T^{(n)}_{\beta, j}$:
\begin{equation*}
  T^{(n)}_{\beta, j} = M_j^{(n)}\bigl(\tilde{\param}(t_n)\bigr) + h L_j^{(n)} \bigl( \tilde{\param}(t_n) \bigr) + O(h^2).
\end{equation*}
Inserting these two expansions into \eqref{eq:derivation-goal} leads to an expression for $B^{(n)}_j(\cdot)$:
\begin{equation*}
B^{(n)}_j(\param) = \frac{M_j^{(n)}(\param) P_j^{(n)}(\param)}{R_j^{(n)}(\param)^3} - \frac{L_j^{(n)}(\param)}{R_j^{(n)}(\param)}.
\end{equation*}
We are now ready to find an ODE for $t \in [t_n, t_{n + 1}]$ of the form $\dot{\tilde{\param}} = \tilde{\bof} \bigl( \tilde{\param}(t) \bigr)$ whose discretization is \eqref{eq:second-order-iteration-form-with-a-and-b}. This is a task for standard backward error analysis: expand $\tilde{\bof}(\cdot)$ into $\tilde{\bof}(\param) = \bof(\param) + h \bof_1(\param) + O(h^2)$. By Taylor expansion, we have
\begin{align*}
  &\tilde{\param}(t_{n + 1}) = \tilde{\param}(t_n) + h \dot{\tilde{\param}}(t_n^+) + \frac{h^2}{2} \ddot{\tilde{\param}}(t_n^+) + O(h^3)\\
  &\quad = \tilde{\param}(t_n) + h \bigl[ \bof\bigl(\tilde{\param}(t_n)\bigr) + h \bof_1 \bigl( \tilde{\param}(t_n) \bigr) + O(h^2) \bigr]\\
  &\qquad + \frac{h^2}{2} \bigl[ \nabla \bof \bigl( \tilde{\param}(t_n) \bigr) \bof \bigl( \tilde{\param}(t_n) \bigr) + O(h) \bigr] + O(h^3)\\
  &\quad = \tilde{\param}(t_n) + h \bof \bigl( \tilde{\param}(t_n) \bigr)\\
  &\qquad + h^2 \biggl[ \bof_1 \bigl( \tilde{\param}(t_n) \bigr) + \frac{\nabla \bof \bigl( \tilde{\param}(t_n) \bigr) \bof \bigl( \tilde{\param}(t_n) \bigr)}{2} \biggr] + O(h^3).
\end{align*}
It is left to equate the terms before the corresponding powers of $h$ here and in \eqref{eq:second-order-iteration-form-with-a-and-b}, giving $f_j(\param) = - A^{(n)}_j(\param)$ and $f_{1, j}(\param) = - \frac{1}{2} \sum_{i = 1}^p \nabla_i f_j(\param) f_i(\param) + B^{(n)}_j(\param)$. Omitting some algebra, the piecewise ODE~\eqref{eq:modified-adam-nth-step-modified-equation} is derived.
\end{proof}

\section{Illustration: Simple Bilinear Model}

\begin{figure*}[h]
  \centerline{\includegraphics[width=0.9\linewidth]{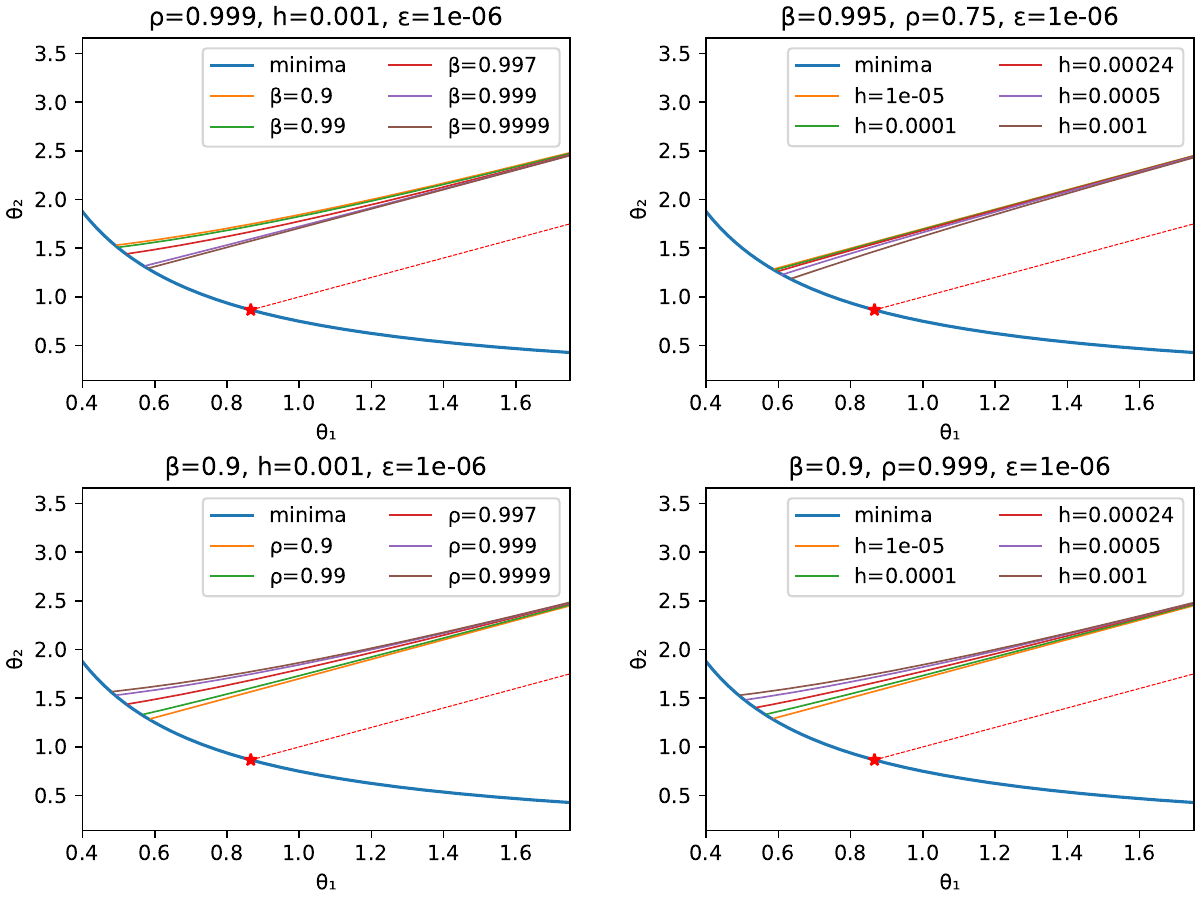}}
  \caption{Left: increasing $\beta$ moves the trajectory of Adam towards the regions with smaller one-norm of the gradient (if $\varepsilon$ is sufficiently small); increasing $\rho$ does the opposite. Right: increasing the learning rate moves the Adam trajectory towards the regions with smaller one-norm of the gradient if $\beta$ is significantly larger than $\rho$ and does the opposite if $\rho$ is larger than $\beta$. The cross denotes the limit point of gradient one-norm minimizers on the level sets $4 \theta_1 \theta_2 - 3 = c$. The minimizers are drawn with a dashed line. All Adam trajectories start at $(2.8, 3.5)$.}
  \label{fig:two-d-illustrations}
\end{figure*}

We now analyze the effect of the first-order term for Adam in the same model as \citet{barrett2021implicit} and \citet{ghosh2023implicit} have studied. Namely, assume the parameter $\param = (\paramsc_1, \paramsc_2)\trans$ is 2-dimensional, and the loss is given by $E(\param) := 1 / 2 ( 3 / 2 - 2 \paramsc_1 \paramsc_2 )^2$. The loss is minimized on the hyperbola $\paramsc_1 \paramsc_2 = 3 / 4$. We graph the trajectories of Adam in this case: the left part of \cref{fig:two-d-illustrations} shows that increasing $\beta$ forces the trajectory to the region with smaller $\| \nabla E(\param) \|_1$, and increasing $\rho$ does the opposite. The right part shows that increasing the learning rate moves Adam towards the region with smaller $\left\| \nabla E(\param) \right\|_1$ if $\beta > \rho$ (just like in the case of GD, except the norm is different if $\varepsilon$ is small compared to gradient components), and does the opposite if $\rho > \beta$. All these observations are exactly what \cref{th:backward-error-analysis-modified-adam} predicts.

\section{Numerical Experiments}\label{sec:numerical}

\begin{figure}[htb!]
  \centerline{\includegraphics[width=0.8\linewidth]{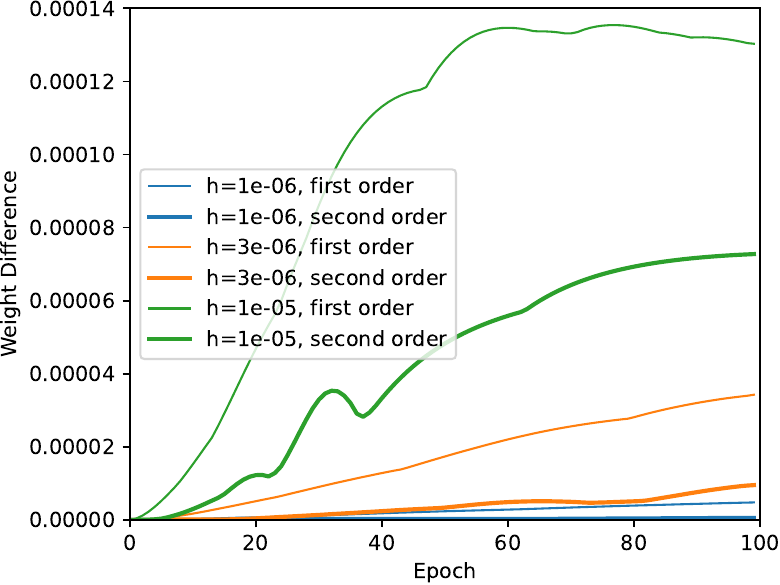}}
  \caption{$\|\param^{(n)} - \tilde{\param}(t_n)\|_{\infty}$ for a MLP trained with full-batch Adam on truncated MNIST, where $\tilde{\param}(t_n)$ is either first (signGD perturbed by $\varepsilon$) or second order approximation to Adam; $\beta = 0.9$, $\rho = 0.95$, $\varepsilon = 10^{-6}$. Precise definitions are provided in \cref{sec:numerical-experiments}, specifically \eqref{eq:bea-closeness-precise-iterations}.}
  \label{fig:mlp_gelu_bea_trajectories}
\end{figure}

As a first sanity check, we train a relatively small fully-connected neural network with around $10^5$ parameters on the first 10,000 images of MNIST with full-batch Adam for 100 epochs and plot the value $\|\param^{(n)} - \tilde{\param}(t_n)\|_{\infty}$, i.\,e. the maximal weight difference between the Adam iteration and the piecewise ODE solution.\footnote{Since it makes little sense to numerically solve an ODE by further discretization, $\tilde{\param}(t_n)$ is estimated using the iteration~\eqref{eq:second-order-iteration-form-with-a-and-b} with $O(h^3)$ ignored. Strictly speaking, this is not the trajectory obtained by the final backward error analysis step but rather the step immediately preceding it (after removing long-term memory but before converting the iteration to an ODE).} We see in \cref{fig:mlp_gelu_bea_trajectories} that even on this very large time horizon the trajectories are close in infinity-norm.

Further, we offer some preliminary empirical evidence that Adam (anti-)penalizes the perturbed one-norm of the gradients, as discussed in \cref{sec:summary}.

\citet{ma2022qualitative} divide training regimes of Adam into three categories: the spike regime when $\rho$ is much larger than $\beta$, in which the training loss curve contains very large spikes and the training is obviously unstable; the (stable) oscillation regime when $\rho$ is sufficiently close to $\beta$, in which the loss curve contains fast and small oscillations; the divergence regime when $\beta$ is much larger than $\rho$, in which Adam diverges. We exclude the last regime. In the spike regime, the loss spikes to large values at irregular intervals. This has also been observed in the context of large transformers, and mitigation strategies have been proposed in \citet{chowdhery2022palm} and \citet{molybog2023theory}. Since it is unlikely that an unstable Adam trajectory can be meaningfully approximated by a smooth ODE solution, we reduce the incidence of large spikes by only considering $\beta$ and $\rho$ that are not too far apart, which is what \citet{ma2022qualitative} recommend to do in practice.

\begin{figure}[htb!]
  \centerline{\includegraphics[width=\linewidth]{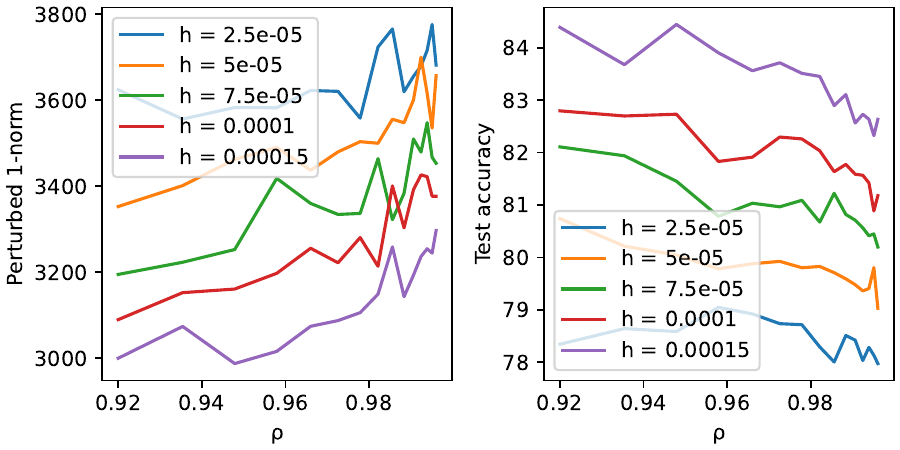}}
  \caption{Resnet-50 on CIFAR-10 trained with full-batch Adam, $\varepsilon = 10^{-8}$, $\beta = 0.99$. As $\rho$ increases, the norm rises and the test accuracy falls. We train longer than necessary for near-perfect classification on the train dataset (at least 2-3 thousand epochs), and the test accuracies plotted here are maximal. The perturbed norms are also maximal after excluding the initial training period (i.\,e., the plotted ``norms'' are at peaks of the ``hills'' described in \cref{sec:numerical}). All results are averaged across five runs with different initialization seeds. Additional evidence and more details are provided in \cref{sec:numerical-experiments}.}
  \label{fig:resnet-50-rho-increase}
\end{figure}

\begin{figure}[htb!]
  \centerline{\includegraphics[width=\linewidth]{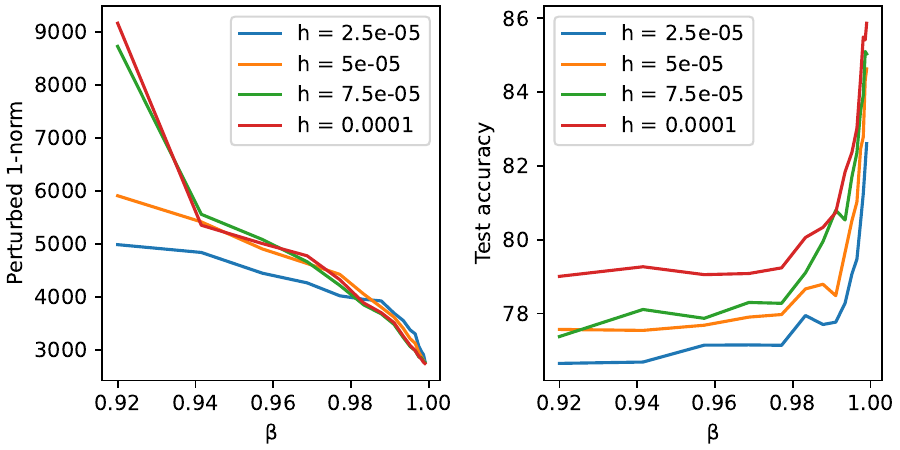}}
  \caption{Resnet-50 on CIFAR-10 trained with full-batch Adam, $\rho = 0.999$, $\varepsilon = 10^{-8}$. The perturbed one-norm falls as $\beta$ increases, and the test accuracy rises. Both metrics are calculated as in \cref{fig:resnet-50-rho-increase}. All results are averaged across three runs with different initialization seeds.}
  \label{fig:resnet-50-beta-increase}
\end{figure}

We train Resnet-50, CNNs and Vision Transformers \citep{dosovitskiy2020image} on the CIFAR-10 dataset with full-batch Adam. In this section, we provide the results for Resnet-50; the pictures for CNNs and Transformers are similar and are given in \cref{sec:additional-evidence}. \cref{fig:resnet-50-rho-increase} shows that in the stable oscillation regime increasing $\rho$ appears to increase the perturbed one-norm (consistent with our analysis: the smaller $\rho$, the more this ``norm'' is penalized) and decrease the test accuracy. \cref{fig:resnet-50-beta-increase} shows that increasing $\beta$ appears to decrease the perturbed one-norm (consistent with our analysis: the larger $\beta$, the more this norm is penalized) and increase the test accuracy. The picture confirms the finding in \citet{ghosh2023implicit} (for the momentum parameter in momentum GD).

\begin{figure}[h]
  \centerline{\includegraphics[width=\linewidth]{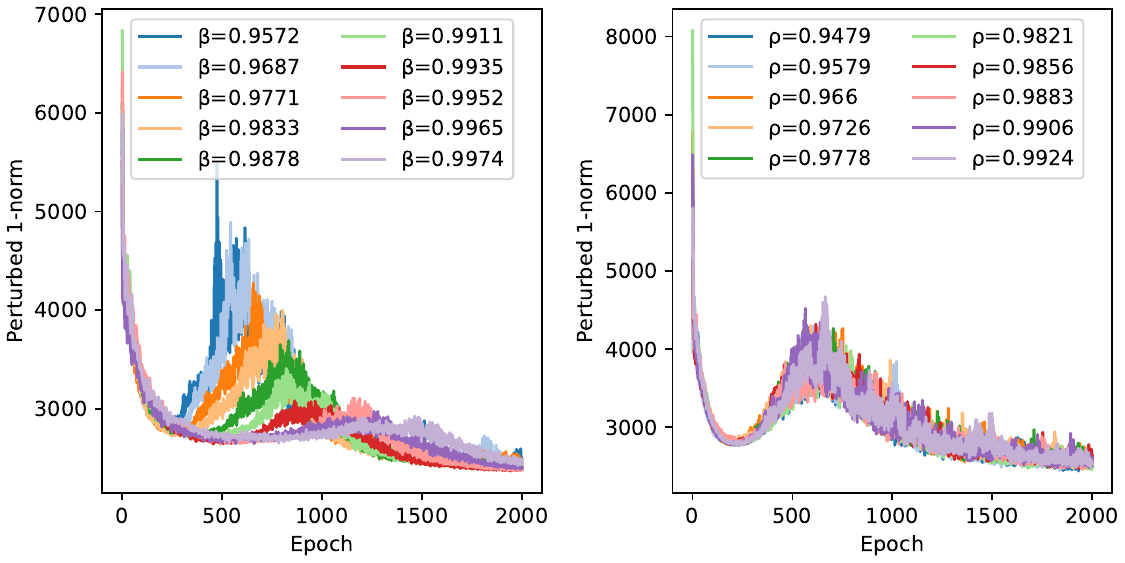}}
  \caption{Plots of $\| \nabla E \|_{1, \varepsilon}$ after each epoch for full-batch Adam, $h = 10^{-4}, \varepsilon = 10^{-8}$. Resnet-50 on CIFAR-10, left: $\rho = 0.999$, right: $\beta = 0.97$.}
  \label{fig:resnet-50-norm-curves}
\end{figure}

\cref{fig:resnet-50-norm-curves} shows the graphs of $\left\| \nabla E \right\|_{1, \varepsilon}$ as functions of the epoch number. The ``norm'' decreases, then rises again, and then decreases further until it flatlines.\footnote{Note that the perturbed one-norm cannot be near-zero at the end of training because it is bounded from below by $p \sqrt{\varepsilon}$.} Throughout most of the training, the larger $\beta$ the smaller the ``norm''. The ``hills'' of the ``norm'' curves are higher with smaller $\beta$ and larger $\rho$. This is consistent with our analysis because the larger $\rho$ compared to $\beta$, the more $\left\| \nabla E \right\|_{1, \varepsilon}$ is prevented from falling by the correction term.

\section{Limitations and Future Directions}

As far as we know, the assumption similar to~\eqref{eq:bounded-smoothness-assumption} is explicitly or implicitly present in all previous work on backward error analysis of gradient-based machine learning algorithms. (Recently, \citet{beneventano2023trajectories} weakened this assumption for SGD without replacement, but their focus is somewhat different.) There is evidence that large-batch algorithms often operate near or at the edge of stability \citep{cohen2021gradient, cohen2022adaptive}, in which the largest eigenvalue of the hessian can be large, making it unclear whether the higher-order partial derivatives can safely be assumed bounded near optimality. In addition, as \citet{smith2021on} point out, in the mini-batch setting backward error analysis can be more accurate. We leave a qualitative analysis of the behavior of first-order terms in \cref{th:backward-error-analysis-modified-adam} in the mini-batch case as a future direction.

Relatedly, Adam does not always generalize worse than SGD: for transformers, Adam often outperforms \citep{zhang2020adaptive,kumar2022fine}. Moreover, for NLP tasks a long time can be spent training close to an interpolating solution. Our analysis suggests that in the latter regime the anti-regularization effect disappears, which does indeed confirm the finding that generalization can be better. However, we believe this explanation is not complete, and more work is needed to connect the implicit bias to the training dynamics of transformers.

In addition, the constant $C$ in \cref{th:backward-error-analysis-modified-adam} goes to infinity as $\varepsilon$ goes to zero. Theoretically, our proof does not exclude the case where for very small $\varepsilon$ the trajectory of the piecewise ODE is only close to the Adam trajectory for small, suboptimal learning rates, at least at later stages of learning. (For the initial learning period, this is not a problem.) It appears to also be true of Proposition~1 in~\citet{ma2022qualitative} (zeroth-order approximation by sign-GD). This is especially noticeable in the large-spike regime of training (see~\cref{sec:numerical}) which, despite being obviously unstable, can still lead to acceptable test errors. It would be worthwhile to investigate this regime in detail.

\section*{Acknowledgments}

We extend our special thanks to Boris Hanin, Sam Smith, and the anonymous reviewers for their insightful comments and suggestions that greatly enhanced this work. We are also grateful to Jianqing Fan, Pier Beneventano, and Rae Yu for engaging and productive discussions. Cattaneo gratefully acknowledges financial support from the National Science Foundation through DMS-2210561 and SES-2241575. Klusowski gratefully acknowledges financial support from the National Science Foundation through CAREER DMS-2239448. Additionally, we acknowledge the Princeton Research Computing resources, coordinated by the Princeton Institute for Computational Science and Engineering (PICSciE) and the Office of Information Technology's Research Computing.

\ifthenelse{\boolean{useICML}}{
  \section*{Impact Statement}

  This paper presents work whose goal is to advance the field of Machine Learning. There are many potential societal consequences of our work, none of which we feel must be specifically highlighted here.}{}

\bibliography{neural_networks_notes}
\bibliographystyle{icml2024}

\newpage
\appendix
\onecolumn

\section{Overview}\label{sec:appendix-overview}

The appendix provide some omitted details and proofs.

We consider two algorithms: RMSProp and Adam, and two versions of each algorithm (with the numerical stability $\varepsilon$ parameter inside and outside of the square root in the denominator). This means there are four main theorems: \cref{th:global-error-bound}, \cref{th:mod-rmsprop-global-error-bound}, \cref{th:adam-global-error-bound} and \cref{th:mod-adam-inside-global-error-bound}, each residing in the section completely devoted to one algorithm. The simple induction argument taken from \citet{ghosh2023implicit}, essentially the same for each of these theorems, is based on an auxiliary result whose corresponding versions are \cref{th:local-error-bound}, \cref{th:mod-rmsprop-local-error-bound}, \cref{th:adam-local-error-bound} and \cref{th:adam-inside-local-error-bound}. The proof of this result is also elementary but long, and it is done by a series of lemmas in \cref{sec:technical-bounding-lemmas} and \cref{sec:proof-of-local-error-bound}. Out of these four, we only prove \cref{th:local-error-bound} since the other three results are proven in the same way with obvious changes.

\cref{sec:numerical-experiments} contains some details about the numerical experiments.

\subsection{Notation} We denote the loss of the $k$th minibatch as a function of the network parameters $\param \in \mathbb{R}^p$ by $E_k(\param)$, and in the full-batch setting we omit the index and write $E(\param)$. As usual, $\nabla E$ means the gradient of $E$, and nabla with indices means partial derivatives, e.\,g. $\nabla_{i j s} E$ is a shortcut for $\frac{\partial^3 E}{\partial \paramsc_i \partial \paramsc_j \partial \paramsc_s}$.

The letter $T > 0$ will always denote a finite time horizon of the ODEs, $h$ will always denote the training step size, and we will replace $n h$ with $t_n$ when convenient, where $n \in \left\{0, 1, \ldots\right\}$ is the step number. \textit{We will use the same notation} for the iteration of the discrete algorithm $\left\{\param^{(k)}\right\}_{k \in \mathbb{Z}_{\geq 0}}$, the piecewise ODE solution $\tilde{\param}(t)$ and some auxiliary terms for each of the four algorithms: see \cref{def:rmsprop-outside}, \cref{def:rmsprop-inside}, \cref{def:adam-outside}, \cref{def:adam-inside}. This way, we avoid cluttering the notation significantly. We are careful to reference the relevant definition in all theorem statements.

\section{RMSProp with $\varepsilon$ Outside the Square Root}

\begin{definition}\label{def:rmsprop-outside}
In this section, for some $\param^{(0)} \in \mathbb{R}^p$, $\mathbf{\nu}^{(0)} = \mathbf{0} \in \mathbb{R}^p$, $\rho \in (0, 1)$, let the sequence of $p$-vectors $\left\{\param^{(k)}\right\}_{k \in \mathbb{Z}_{\geq 0}}$ be defined for $n \geq 0$ by
\begin{equation}\label{eq:rmsprop-iteration}
\begin{aligned}
  &\nu_j^{(n + 1)} = \rho \nu_j^{(n)} + (1 - \rho) \left( \nabla_j E_n\left(\param^{(n)}\right) \right)^2,\\
  &\paramsc_j^{(n + 1)} = \paramsc_j^{(n)} - \frac{h}{\sqrt{\nu_j^{(n + 1)}} + \varepsilon} \nabla_j E_n\left(\param^{(n)}\right).
\end{aligned}
\end{equation}

Let $\tilde{\param}(t)$ be defined as a continuous solution to the piecewise ODE
\begin{equation}\label{eq:nth-step-modified-equation}
\begin{aligned}
  &\dot{\tilde{\paramsc}}_j(t) = - \frac{\nabla_j E_n\left(\tilde{\param}(t)\right)}{R^{(n)}_j\left(\tilde{\param}(t)\right) + \varepsilon}\\
  &\qquad + h \left( \frac{\nabla_j E_n \left( \tilde{\param}(t) \right) \left(2 P_j^{(n)}\left(\tilde{\param}(t)\right) + \bar{P}_j^{(n)}\left(\tilde{\param}(t)\right)\right)}{2 \left( R_j^{(n)} \left( \tilde{\param}(t) \right) + \varepsilon \right)^2 R_j^{(n)} \left( \tilde{\param}(t) \right)} - \frac{\sum_{i = 1}^p \nabla_{i j} E_n \left( \tilde{\param}(t) \right) \frac{\nabla_i E_n \left( \tilde{\param}(t) \right)}{R_i^{(n)}\left(\tilde{\param}(t)\right) + \varepsilon}}{2 \left( R_j^{(n)}(\tilde{\param}(t)) + \varepsilon \right)} \right)
\end{aligned}
\end{equation}
for $t \in [t_n, t_{n + 1}]$ with the initial condition $\tilde{\param}(0) = \param^{(0)}$, where $\mathbf{R}^{(n)}(\param)$, $\mathbf{P}^{(n)}(\param)$ and $\bar{\mathbf{P}}^{(n)}(\param)$ are $p$-dimensional functions with components
\begin{align*}
  &R^{(n)}_j(\param) := \sqrt{\sum_{k = 0}^n \rho^{n - k} (1 - \rho) \left( \nabla_j E_k(\param) \right)^2},\\
  &P_j^{(n)}(\param) := \sum_{k = 0}^n \rho^{n - k} (1 - \rho) \nabla_j E_k \left( \param \right) \sum_{i = 1}^p \nabla_{i j} E_k \left( \param \right) \sum_{l = k}^{n - 1} \frac{\nabla_i E_l \left( \param \right)}{R_i^{(l)} \left( \param \right) + \varepsilon},\\
  &\bar{P}_j^{(n)}(\param) := \sum_{k = 0}^n \rho^{n - k} (1 - \rho) \nabla_j E_k \left( \param \right) \sum_{i = 1}^p \nabla_{i j} E_k \left( \param \right) \frac{\nabla_i E_n \left( \param \right)}{R_i^{(n)} \left( \param \right) + \varepsilon}.
\end{align*}
\end{definition}

\begin{assumption}\label{ass:bounds}\leavevmode
  \begin{enumerate}
  \item For some positive constants $M_1$, $M_2$, $M_3$, $M_4$ we have
    \begin{align*}
      &\sup_{i} \sup_k \sup_{\param} \left| \nabla_{i} E_k(\param) \right| \leq M_1,\\
      &\sup_{i, j} \sup_k \sup_{\param} \left| \nabla_{i j} E_k(\param) \right| \leq M_2,\\
      &\sup_{i, j, s} \sup_k \sup_{\param} \left| \nabla_{i j s} E_k(\param) \right| \leq M_3,\\
      &\sup_{i, j, s, r} \sup_k \sup_{\param} \left| \nabla_{i j s r} E_k(\param) \right| \leq M_4.
    \end{align*}

  \item For some $R > 0$ we have for all $n \in \left\{ 0, 1, \dots, \lfloor T / h \rfloor\right\}$
    \begin{equation*}
      R^{(n)}_j \left( \tilde{\param}(t_n) \right) \geq R,\quad \sum_{k = 0}^n \rho^{n - k} (1 - \rho) \left( \nabla_j E_k \left( \tilde{\param}(t_k) \right) \right)^2 \geq R^2,
    \end{equation*}
    where $\tilde{\param}(t)$ is defined in \cref{def:rmsprop-outside}.
  \end{enumerate}
\end{assumption}

\begin{theorem}[RMSProp with $\varepsilon$ outside: local error bound]\label{th:local-error-bound}
  Suppose \cref{ass:bounds} holds. Then for all $n \in \left\{ 0, 1, \dots, \lfloor T / h \rfloor \right\}$, $j \in \left\{ 1, \ldots, p \right\}$
\begin{equation*}
\left| \tilde{\paramsc}_j(t_{n + 1}) - \tilde{\paramsc}_j(t_n) + h \frac{\nabla_j E_n \left( \tilde{\param}(t_n) \right)}{\sqrt{\sum_{k = 0}^n \rho^{n - k} (1 - \rho) \left( \nabla_j E_k \left( \tilde{\param}(t_k) \right) \right)^2} + \varepsilon} \right| \leq \oldbigc{localerrorbound} h^3
\end{equation*}
for a positive constant $\newbigc[localerrorbound]$ depending on $\rho$.
\end{theorem}

The proof of \cref{th:local-error-bound} is conceptually simple but very technical, and we delay it until \cref{sec:proof-of-local-error-bound}. For now assuming it as given and combining it with a simple induction argument gives a global error bound which follows.

\begin{theorem}[RMSProp with $\varepsilon$ outside: global error bound]\label{th:global-error-bound}
  Suppose \cref{ass:bounds} holds, and
\begin{equation*}
\sum_{k = 0}^n \rho^{n - k} (1 - \rho) \left( \nabla_j E_k \left( \param^{(k)} \right) \right)^2 \geq R^2
\end{equation*}
for $\left\{\param^{(k)}\right\}_{k \in \mathbb{Z}_{\geq 0}}$ defined in \cref{def:rmsprop-outside}. Then there exist positive constants $\newd[rmsprop-outside-d]$, $\newd[rmsprop-outside-dd]$, $\newd[rmsprop-outside-ddd]$ such that for all $n \in \left\{ 0, 1, \dots, \lfloor T / h \rfloor \right\}$
\begin{equation*}
\left\| \mathbf{e}_n\right\| \leq \oldd{rmsprop-outside-d} e^{\oldd{rmsprop-outside-dd} n h} h^2\quad\text{and}\quad \left\| \mathbf{e}_{n + 1} - \mathbf{e}_n \right\| \leq \oldd{rmsprop-outside-ddd} e^{\oldd{rmsprop-outside-dd} n h} h^3,
\end{equation*}
where $\mathbf{e}_n := \tilde{\param}(t_n) - \param^{(n)}$. The constants can be defined as
\begin{align*}
  &\oldd{rmsprop-outside-d} := \oldbigc{localerrorbound},\\
  &\oldd{rmsprop-outside-dd} := \left[ 1 + \frac{M_2 \sqrt{p}}{R + \varepsilon} \left(\frac{M_1^2}{R (R + \varepsilon)} + 1\right) \oldd{rmsprop-outside-d} \right] \sqrt{p},\\
  &\oldd{rmsprop-outside-ddd} := \oldbigc{localerrorbound} \oldd{rmsprop-outside-dd}.
\end{align*}
\end{theorem}

\begin{proof}
We will show this by induction over $n$, the same way an analogous bound is shown in~\citet{ghosh2023implicit}.

  The base case is $n = 0$. Indeed, $\mathbf{e}_0 = \tilde{\param}(0) - \param^{(0)} = \mathbf{0}$. Then the $j$th component of $\mathbf{e}_1 - \mathbf{e}_0$ is
\begin{align*}
  &\left[\mathbf{e}_{1} - \mathbf{e}_{0}\right]_j = \left[\mathbf{e}_{1}\right]_j = \tilde{\paramsc}_j(t_1) - \paramsc_j^{(0)} + \frac{h \nabla_j E_0 \left( \param^{(0)} \right)}{\sqrt{(1 - \rho) \left( \nabla_j E_0 \left( \param^{(0)} \right) \right)^2} + \varepsilon}\\
  &\quad = \tilde{\paramsc}_j(t_1) - \tilde{\paramsc}_j(t_0) + \frac{h \nabla_j E_0 \left( \tilde{\param}(t_0) \right)}{\sqrt{(1 - \rho) \left( \nabla_j E_0 \left( \tilde{\param}(t_0) \right) \right)^2} + \varepsilon}.
\end{align*}
By \cref{th:local-error-bound}, the absolute value of the right-hand side does not exceed $\oldbigc{localerrorbound} h^3$, which means $\left\| \mathbf{e}_{1} - \mathbf{e}_{0} \right\| \leq \oldbigc{localerrorbound} h^3 \sqrt{p}$. Since $\oldbigc{localerrorbound} \sqrt{p} \leq \oldd{rmsprop-outside-ddd}$, the base case is proven.

Now suppose that for all $k = 0, 1, \ldots, n - 1$ the claim
\begin{equation*}
\left\| \mathbf{e}_k\right\| \leq \oldd{rmsprop-outside-d} e^{\oldd{rmsprop-outside-dd} k h} h^2\quad\text{and}\quad \left\| \mathbf{e}_{k + 1} - \mathbf{e}_k \right\| \leq \oldd{rmsprop-outside-ddd} e^{\oldd{rmsprop-outside-dd} k h} h^3
\end{equation*}
is proven. Then
\begin{align*}
  &\left\| \mathbf{e}_n \right\| \labrel{1}[\leq] \left\| \mathbf{e}_{n - 1} \right\| + \left\| \mathbf{e}_n - \mathbf{e}_{n - 1} \right\| \leq \oldd{rmsprop-outside-d} e^{\oldd{rmsprop-outside-dd} (n - 1) h} h^2 + \oldd{rmsprop-outside-ddd} e^{\oldd{rmsprop-outside-dd} (n - 1) h} h^3\\
  &\quad = \oldd{rmsprop-outside-d} e^{\oldd{rmsprop-outside-dd} (n - 1) h} h^2 \left( 1 + \frac{\oldd{rmsprop-outside-ddd}}{\oldd{rmsprop-outside-d}} h \right) \labrel{2}[\leq] \oldd{rmsprop-outside-d} e^{\oldd{rmsprop-outside-dd} (n - 1) h} h^2 \left( 1 + \oldd{rmsprop-outside-dd} h \right)\\
  &\quad \labrel{3}[\leq] \oldd{rmsprop-outside-d} e^{\oldd{rmsprop-outside-dd} (n - 1) h} h^2 \cdot e^{\oldd{rmsprop-outside-dd} h} = \oldd{rmsprop-outside-d} e^{\oldd{rmsprop-outside-dd} n h} h^2,
\end{align*}
where \labrel{1} is by the triangle inequality, \labrel{2} is by $\oldd{rmsprop-outside-ddd} / \oldd{rmsprop-outside-d} \leq \oldd{rmsprop-outside-dd}$, in \labrel{3} we used $1 + x \leq e^x$ for all $x \geq 0$.

Next, combining \cref{th:local-error-bound} with~\eqref{eq:rmsprop-iteration}, we have
\begin{equation}\label{eq:global-start-bounding-e-n-pl-one-min-e-n}
  \left| \left[\mathbf{e}_{n + 1} - \mathbf{e}_n\right]_j \right| \leq \oldbigc{localerrorbound} h^3 + h \left| \frac{\nabla_j E_n \left( \tilde{\param}(t_n) \right)}{\sqrt{A} + \varepsilon} - \frac{\nabla_j E_n \left( \param^{(n)} \right)}{\sqrt{B} + \varepsilon} \right|,
\end{equation}
where to simplify notation we put
\begin{align*}
  A := \sum_{k = 0}^n \rho^{n - k} (1 - \rho) \left( \nabla_j E_k \left( \tilde{\param}(t_k) \right) \right)^2,\\
  B := \sum_{k = 0}^n \rho^{n - k} (1 - \rho) \left( \nabla_j E_k \left( \param^{(k)} \right) \right)^2.
\end{align*}

Using $A \geq R^2$, $B \geq R^2$, we have
\begin{equation}\label{eq:global-diff-of-frac-bound}
\left| \frac{1}{\sqrt{A} + \varepsilon} - \frac{1}{\sqrt{B} + \varepsilon}\right| = \frac{\left| A - B \right|}{\left( \sqrt{A} + \varepsilon \right) \left( \sqrt{B} + \varepsilon \right) \left( \sqrt{A} + \sqrt{B} \right)} \leq \frac{\left| A - B \right|}{2 R \left( R + \varepsilon \right)^2}.
\end{equation}
But since
\begin{align*}
  &\left| \left( \nabla_j E_k \left( \tilde{\param}(t_k) \right) \right)^2 - \left( \nabla_j E_k \left( \param^{(k)} \right) \right)^2 \right|\\
  &\quad = \left| \nabla_j E_k \left( \tilde{\param}(t_k) \right) - \nabla_j E_k \left( \param^{(k)} \right) \right| \cdot \left| \nabla_j E_k \left( \tilde{\param}(t_k) \right) + \nabla_j E_k \left( \param^{(k)} \right) \right|\\
  &\quad \leq 2 M_1 \left| \nabla_j E_k \left( \tilde{\param}(t_k) \right) - \nabla_j E_k \left( \param^{(k)} \right) \right| \leq 2 M_1 M_2 \sqrt{p} \left\| \tilde{\param}(t_k) - \param^{(k)} \right\|,
\end{align*}
we have
\begin{equation}\label{eq:global-a-b-bound}
\left| A - B \right| \leq 2 M_1 M_2 \sqrt{p} \sum_{k = 0}^n \rho^{n - k} (1 - \rho) \left\| \tilde{\param}(t_k) - \param^{(k)} \right\|.
\end{equation}
Combining~\eqref{eq:global-diff-of-frac-bound} and~\eqref{eq:global-a-b-bound}, we obtain
\begin{align}
  &\left| \frac{\nabla_j E_n \left( \tilde{\param}(t_n) \right)}{\sqrt{A} + \varepsilon} - \frac{\nabla_j E_n \left( \param^{(n)} \right)}{\sqrt{B} + \varepsilon} \right|\nonumber\\
  &\quad \leq \left| \nabla_j E_n \left( \tilde{\param}(t_n) \right) \right| \cdot \left| \frac{1}{\sqrt{A} + \varepsilon} - \frac{1}{\sqrt{B} + \varepsilon}\right| + \frac{\left| \nabla_j E_n \left( \tilde{\param}(t_n) \right) - \nabla_j E_n \left( \param^{(n)} \right) \right|}{\sqrt{B} + \varepsilon}\nonumber\\
  &\quad \leq M_1 \cdot \frac{2 M_1 M_2 \sqrt{p} \sum_{k = 0}^n \rho^{n - k} (1 - \rho) \left\| \tilde{\param}(t_k) - \param^{(k)} \right\|}{2 R (R + \varepsilon)^2} + \frac{M_2 \sqrt{p} \left\| \tilde{\param}(t_n) - \param^{(n)} \right\|}{R + \varepsilon}\nonumber\\
  &\quad = \frac{M_1^2 M_2 \sqrt{p}}{R (R + \varepsilon)^2} \sum_{k = 0}^n \rho^{n - k} (1 - \rho) \left\| \tilde{\param}(t_k) - \param^{(k)} \right\| + \frac{M_2 \sqrt{p}}{R + \varepsilon} \left\| \tilde{\param}(t_n) - \param^{(n)} \right\|\nonumber\\
  &\quad \labrel{1}[\leq] \frac{M_1^2 M_2 \sqrt{p}}{R (R + \varepsilon)^2} \sum_{k = 0}^n \rho^{n - k} (1 - \rho) \oldd{rmsprop-outside-d} e^{\oldd{rmsprop-outside-dd} k h} h^2 + \frac{M_2 \sqrt{p}}{R + \varepsilon} \oldd{rmsprop-outside-d} e^{\oldd{rmsprop-outside-dd} n h} h^2\label{eq:global-big-frac-dif-stop-point},
\end{align}
where in \labrel{1} we used the induction hypothesis and that the bound on $\left\| \mathbf{e}_n \right\|$ is already proven.

Now note that since $0 < \rho e^{-\oldd{rmsprop-outside-dd} h} \leq \rho$, we have $\sum_{k = 0}^n \left( \rho e^{-\oldd{rmsprop-outside-dd} h} \right)^k \leq \sum_{k = 0}^{\infty} \rho^k = \frac{1}{1 - \rho}$, which is rewritten as
\begin{equation*}
\sum^n_{k = 0} \rho^{n - k} (1 - \rho) e^{\oldd{rmsprop-outside-dd} k h} \leq e^{\oldd{rmsprop-outside-dd} n h}.
\end{equation*}
Then we can continue~\eqref{eq:global-big-frac-dif-stop-point}:
\begin{equation}\label{eq:global-big-frac-dif-final-bound}
\left| \frac{\nabla_j E_n \left( \tilde{\param}(t_n) \right)}{\sqrt{A} + \varepsilon} - \frac{\nabla_j E_n \left( \param^{(n)} \right)}{\sqrt{B} + \varepsilon} \right| \leq \frac{M_2 \sqrt{p}}{R + \varepsilon} \left(\frac{M_1^2}{R (R + \varepsilon)} + 1\right) \oldd{rmsprop-outside-d} e^{\oldd{rmsprop-outside-dd} n h} h^2
\end{equation}
Again using $1 \leq e^{\oldd{rmsprop-outside-dd} n h}$, we conclude from~\eqref{eq:global-start-bounding-e-n-pl-one-min-e-n} and~\eqref{eq:global-big-frac-dif-final-bound} that
\begin{equation*}
\left\| \mathbf{e}_{n + 1} - \mathbf{e}_n \right\| \leq \underbrace{\left( \oldbigc{localerrorbound} + \frac{M_2 \sqrt{p}}{R + \varepsilon} \left(\frac{M_1^2}{R (R + \varepsilon)} + 1\right) \oldd{rmsprop-outside-d}\right) \sqrt{p}}_{\leq \oldd{rmsprop-outside-ddd}} e^{\oldd{rmsprop-outside-dd} n h} h^3,
\end{equation*}
finishing the induction step.
\end{proof}

\subsection{RMSProp with $\varepsilon$ outside: full-batch}
In the full-batch setting $E_k \equiv E$, the terms in~\eqref{eq:nth-step-modified-equation} simplify to
\begin{align*}
  &R_j^{(n)}(\param) = \left| \nabla_j E(\param) \right| \sqrt{1 - \rho^{n + 1}},\\
  &P_j^{(n)}(\param) = \sum_{k = 0}^n \rho^{n - k} (1 - \rho) \nabla_j E(\param) \sum_{i = 1}^p \nabla_{i j} E(\param) \sum_{l = k}^{n - 1} \frac{\nabla_i E(\param)}{\left| \nabla_i E(\param) \right| \sqrt{1 - \rho^{l + 1}} + \varepsilon},\\
  &\bar{P}_j^{(n)}(\param) = \left(1 - \rho^{n + 1}\right) \nabla_j E(\param) \sum_{i = 1}^p \nabla_{i j} E(\param) \frac{\nabla_i E(\param)}{\left| \nabla_i E(\param) \right| \sqrt{1 - \rho^{n + 1}} + \varepsilon}.
\end{align*}

If $\varepsilon$ is small and the iteration number $n$ is large, \eqref{eq:nth-step-modified-equation} simplifies to
\begin{align*}
  &\dot{\tilde{\paramsc}}_j(t) = -\sign \nabla_j E(\tilde{\param}(t)) + h \frac{\rho}{1 - \rho} \cdot \frac{\sum_{i = 1}^p \nabla_{i j} E(\tilde{\param}(t)) \sign \nabla_i E(\tilde{\param}(t))}{\left| \nabla_j E(\tilde{\param}(t)) \right|}\\
  &\quad = \left| \nabla_j E(\tilde{\param}(t)) \right|^{-1} \left[ - \nabla_j E(\tilde{\param}(t)) + h \frac{\rho}{1 - \rho} \nabla_j \left\| \nabla E(\tilde{\param}(t)) \right\|_1 \right].
\end{align*}

\section{RMSProp with $\varepsilon$ Inside the Square Root}\label{sec:appendix-about-rmsprop-with-eps-inside}

\begin{definition}\label{def:rmsprop-inside}
  In this section, for some $\param^{(0)} \in \mathbb{R}^p$, $\mathbf{\nu}^{(0)} = \mathbf{0} \in \mathbb{R}^p$, $\rho \in (0, 1)$, let the sequence of $p$-vectors $\left\{\param^{(k)}\right\}_{k \in \mathbb{Z}_{\geq 0}}$ be defined for $n \geq 0$ by
\begin{equation}\label{eq:mod-rmsprop-iteration}
\begin{aligned}
  &\nu_j^{(n + 1)} = \rho \nu_j^{(n)} + (1 - \rho) \left( \nabla_j E_n\left(\param^{(n)}\right) \right)^2,\\
  &\paramsc_j^{(n + 1)} = \paramsc_j^{(n)} - \frac{h}{\sqrt{\nu_j^{(n + 1)} + \varepsilon}} \nabla_j E_n\left(\param^{(n)}\right).
\end{aligned}
\end{equation}

Let $\tilde{\param}(t)$ be defined as a continuous solution to the piecewise ODE
\begin{equation}\label{eq:mod-nth-step-modified-equation}
\begin{aligned}
  &\dot{\tilde{\paramsc}}_j(t) = - \frac{\nabla_j E_n\left(\tilde{\param}(t)\right)}{R^{(n)}_j\left(\tilde{\param}(t)\right)}\\
  &\qquad + h \left( \frac{\nabla_j E_n \left( \tilde{\param}(t) \right) \left(2 P_j^{(n)}\left(\tilde{\param}(t)\right) + \bar{P}_j^{(n)}\left(\tilde{\param}(t)\right)\right)}{2 R_j^{(n)} \left( \tilde{\param}(t) \right)^3} - \frac{\sum_{i = 1}^p \nabla_{i j} E_n \left( \tilde{\param}(t) \right) \frac{\nabla_i E_n \left( \tilde{\param}(t) \right)}{R_i^{(n)}\left(\tilde{\param}(t)\right)}}{2 R_j^{(n)}(\tilde{\param}(t))} \right).
\end{aligned}
\end{equation}
for $t \in [t_n, t_{n + 1}]$ with the initial condition $\tilde{\param}(0) = \param^{(0)}$, where $\mathbf{R}^{(n)}(\param)$, $\mathbf{P}^{(n)}(\param)$ and $\bar{\mathbf{P}}^{(n)}(\param)$ are $p$-dimensional functions with components
\begin{equation}\label{eq:r-p-modified-rmsprop}
\begin{aligned}
  &R^{(n)}_j(\param) := \sqrt{\sum_{k = 0}^n \rho^{n - k} (1 - \rho) \left( \nabla_j E_k(\param) \right)^2 + \varepsilon},\\
  &P_j^{(n)}(\param) := \sum_{k = 0}^n \rho^{n - k} (1 - \rho) \nabla_j E_k \left( \param \right) \sum_{i = 1}^p \nabla_{i j} E_k \left( \param \right) \sum_{l = k}^{n - 1} \frac{\nabla_i E_l \left( \param \right)}{R_i^{(l)} \left( \param \right)},\\
  &\bar{P}_j^{(n)}(\param) := \sum_{k = 0}^n \rho^{n - k} (1 - \rho) \nabla_j E_k \left( \param \right) \sum_{i = 1}^p \nabla_{i j} E_k \left( \param \right) \frac{\nabla_i E_n \left( \param \right)}{R_i^{(n)} \left( \param \right)}.
\end{aligned}
\end{equation}
\end{definition}

\begin{assumption}\label{ass:mod-rmsprop-bounds}\leavevmode
  For some positive constants $M_1$, $M_2$, $M_3$, $M_4$ we have
  \begin{align*}
    &\sup_{i} \sup_k \sup_{\param} \left| \nabla_{i} E_k(\param) \right| \leq M_1,\\
    &\sup_{i, j} \sup_k \sup_{\param} \left| \nabla_{i j} E_k(\param) \right| \leq M_2,\\
    &\sup_{i, j, s} \sup_k \sup_{\param} \left| \nabla_{i j s} E_k(\param) \right| \leq M_3,\\
    &\sup_{i, j, s, r} \sup_k \sup_{\param} \left| \nabla_{i j s r} E_k(\param) \right| \leq M_4.
  \end{align*}
\end{assumption}

\begin{theorem}[RMSProp with $\varepsilon$ inside: local error bound]\label{th:mod-rmsprop-local-error-bound}
  Suppose \cref{ass:mod-rmsprop-bounds} holds. Then for all $n \in \left\{ 0, 1, \dots, \lfloor T / h \rfloor \right\}$, $j \in \left\{ 1, \ldots, p \right\}$
\begin{equation*}
\left| \tilde{\paramsc}_j(t_{n + 1}) - \tilde{\paramsc}_j(t_n) + h \frac{\nabla_j E_n \left( \tilde{\param}(t_n) \right)}{\sqrt{\sum_{k = 0}^n \rho^{n - k} (1 - \rho) \left( \nabla_j E_k \left( \tilde{\param}(t_k) \right) \right)^2 + \varepsilon}} \right| \leq \oldbigc{mod-rmsprop-localerrorbound} h^3
\end{equation*}
for a positive constant $\newbigc[mod-rmsprop-localerrorbound]$ depending on $\rho$, where $\tilde{\param}(t)$ is defined in \cref{def:rmsprop-inside}.
\end{theorem}

The argument is the same as for \cref{th:local-error-bound}.

\begin{theorem}[RMSProp with $\varepsilon$ inside: global error bound]\label{th:mod-rmsprop-global-error-bound}
  Suppose \cref{ass:mod-rmsprop-bounds} holds. Then there exist positive constants $\newd[rmsprop-inside-d]$, $\newd[rmsprop-inside-dd]$, $\newd[rmsprop-inside-ddd]$ such that for all $n \in \left\{ 0, 1, \dots, \lfloor T / h \rfloor \right\}$
\begin{equation*}
\left\| \mathbf{e}_n\right\| \leq \oldd{rmsprop-inside-d} e^{\oldd{rmsprop-inside-dd} n h} h^2\quad\text{and}\quad \left\| \mathbf{e}_{n + 1} - \mathbf{e}_n \right\| \leq \oldd{rmsprop-inside-ddd} e^{\oldd{rmsprop-inside-dd} n h} h^3,
\end{equation*}
where $\mathbf{e}_n := \tilde{\param}(t_n) - \param^{(n)}$; $\tilde{\param}(t)$ and $\left\{\param^{(k)}\right\}_{k \in \mathbb{Z}_{\geq 0}}$ are defined in \cref{def:rmsprop-inside}. The constants can be defined as
\begin{align*}
  &\oldd{rmsprop-inside-d} := \oldbigc{mod-rmsprop-localerrorbound},\\
  &\oldd{rmsprop-inside-dd} := \left[ 1 + \frac{M_2 \sqrt{p}}{\sqrt{\varepsilon}} \left(\frac{M_1^2}{\varepsilon} + 1\right) \oldd{rmsprop-inside-d} \right] \sqrt{p},\\
  &\oldd{rmsprop-inside-ddd} := \oldbigc{mod-rmsprop-localerrorbound} \oldd{rmsprop-inside-dd}.
\end{align*}
\end{theorem}

The argument is the same as for \cref{th:global-error-bound}.

\subsection{RMSProp with $\varepsilon$ Inside: Full-Batch}
In the full-batch setting $E_k \equiv E$, the terms in~\eqref{eq:mod-nth-step-modified-equation} simplify to
\begin{align*}
  &R_j^{(n)}(\param) = \sqrt{\left| \nabla_j E(\param) \right|^2 (1 - \rho^{n + 1}) + \varepsilon},\\
  &P_j^{(n)}(\param) = \sum_{k = 0}^n \rho^{n - k} (1 - \rho) \nabla_j E(\param) \sum_{i = 1}^p \nabla_{i j} E(\param) \sum_{l = k}^{n - 1} \frac{\nabla_i E(\param)}{\sqrt{| \nabla_i E(\param) |^2 (1 - \rho^{l + 1}) + \varepsilon}},\\
  &\bar{P}_j^{(n)}(\param) = (1 - \rho^{n + 1}) \nabla_j E(\param) \sum_{i = 1}^p \nabla_{i j} E(\param) \frac{\nabla_i E(\param)}{\sqrt{| \nabla_i E(\param)|^2 (1 - \rho^{n + 1}) + \varepsilon}}.
\end{align*}
If the iteration number $n$ is large, \eqref{eq:mod-nth-step-modified-equation} rapidly becomes
\begin{equation*}
  \dot{\tilde{\paramsc}}_j(t) = - \frac{1}{\sqrt{| \nabla_j E(\tilde{\param}(t))|^2 + \varepsilon}} \big( \nabla_j E(\tilde{\param}(t)) + \text{correction}_j\big(\tilde{\param}(t)\big) \big),
\end{equation*}
where
\begin{equation*}
\text{correction}_j\big(\tilde{\param}(t)\big) := \frac{h}{2} \left\{ -\frac{2 \rho}{1 - \rho} + \frac{1 + \rho}{1 - \rho} \cdot \frac{\varepsilon}{| \nabla_j E(\tilde{\param}(t))|^2 + \varepsilon} \right\} \nabla_j \big\| \nabla E(\tilde{\param}(t)) \big\|_{1, \varepsilon}.
\end{equation*}

\section{Adam with $\varepsilon$ Outside the Square Root}

\begin{definition}\label{def:adam-outside}
  In this section, for some $\param^{(0)} \in \mathbb{R}^p$, $\mathbf{\nu}^{(0)} = \mathbf{0} \in \mathbb{R}^p$, $\beta, \rho \in (0, 1)$, let the sequence of $p$-vectors $\left\{\param^{(k)}\right\}_{k \in \mathbb{Z}_{\geq 0}}$ be defined for $n \geq 0$ by
\begin{equation*}
\begin{aligned}
  &\nu_j^{(n + 1)} = \rho \nu_j^{(n)} + (1 - \rho) \left( \nabla_j E_n\left(\param^{(n)}\right) \right)^2,\\
  &m_j^{(n + 1)} = \beta m_j^{(n)} + (1 - \beta) \nabla_j E_n \left( \param^{(n)} \right),\\
  &\paramsc_j^{(n + 1)} = \paramsc_j^{(n)} - h \frac{m_j^{(n + 1)} / \left(1 - \beta^{n + 1}\right)}{\sqrt{\nu_j^{(n + 1)} / \left( 1 - \rho^{n + 1} \right)} + \varepsilon}
\end{aligned}
\end{equation*}
or, rewriting,
\begin{equation}\label{eq:adam-iteration}
  \paramsc_j^{(n + 1)} = \paramsc_j^{(n)} - h \frac{\frac{1}{1 - \beta^{n + 1}} \sum_{k = 0}^n \beta^{n - k} \left(1 - \beta\right) \nabla_j E_k \left( \param^{(k)} \right)}{\sqrt{\frac{1}{1 - \rho^{n + 1}}\sum_{k = 0}^n \rho^{n - k} (1 - \rho) \left( \nabla_j E_k \left( \param^{(k)} \right) \right)^2} + \varepsilon}.
\end{equation}

Let $\tilde{\param}(t)$ be defined as a continuous solution to the piecewise ODE
\begin{equation}\label{eq:adam-nth-step-modified-equation}
\begin{aligned}
  &\dot{\tilde{\paramsc}}_j(t) = - \frac{M_j^{(n)} \left( \tilde{\param}(t) \right)}{R^{(n)}_j\left(\tilde{\param}(t)\right) + \varepsilon}\\
  &\qquad + h \left( \frac{M_j^{(n)}\left( \tilde{\param}(t) \right) \left(2 P_j^{(n)}\left(\tilde{\param}(t)\right) + \bar{P}_j^{(n)}\left(\tilde{\param}(t)\right)\right)}{2 \left(R_j^{(n)} \left( \tilde{\param}(t) \right) + \varepsilon\right)^2 R_j^{(n)} \left( \tilde{\param}(t) \right)} - \frac{2 L_j^{(n)} \left( \tilde{\param}(t) \right) + \bar{L}_j^{(n)} \left( \tilde{\param}(t) \right)}{2 \left(R_j^{(n)}\left(\tilde{\param}(t)\right) + \varepsilon\right)} \right).
\end{aligned}
\end{equation}
for $t \in [t_n, t_{n + 1}]$ with the initial condition $\tilde{\param}(0) = \param^{(0)}$, where $\mathbf{R}^{(n)}(\param)$, $\mathbf{P}^{(n)}(\param)$, $\bar{\mathbf{P}}^{(n)}(\param)$, $\mathbf{M}^{(n)}(\param)$, $\mathbf{L}^{(n)}(\param)$, $\bar{\mathbf{L}}^{(n)}(\param)$ are $p$-dimensional functions with components
\begin{equation}\label{eq:r-p-adam}
\begin{aligned}
  &R^{(n)}_j(\param) := \sqrt{\sum_{k = 0}^n \rho^{n - k} (1 - \rho) \left( \nabla_j E_k(\param) \right)^2 / \left(1 - \rho^{n + 1}\right)},\\
  &M^{(n)}_j(\param) := \frac{1}{1 - \beta^{n + 1}} \sum_{k = 0}^n \beta^{n - k} \left(1 - \beta\right) \nabla_j E_k \left( \param \right),\\
  &L_j^{(n)}(\param) := \frac{1}{1 - \beta^{n + 1}} \sum_{k = 0}^n \beta^{n - k} (1 - \beta) \sum_{i = 1}^p \nabla_{i j} E_k(\param) \sum_{l = k}^{n - 1} \frac{M^{(l)}_i(\param)}{R_i^{(l)}(\param) + \varepsilon},\\
  &\bar{L}_j^{(n)}(\param) := \frac{1}{1 - \beta^{n + 1}} \sum_{k = 0}^n \beta^{n - k} (1 - \beta) \sum_{i = 1}^p \nabla_{i j} E_k(\param) \frac{M^{(n)}_i(\param)}{R_i^{(n)}(\param) + \varepsilon},\\
  &P_j^{(n)}(\param) := \frac{1}{1 - \rho^{n + 1}} \sum_{k = 0}^n \rho^{n - k} (1 - \rho) \nabla_j E_k(\param) \sum_{i = 1}^p \nabla_{i j} E_k(\param) \sum_{l = k}^{n - 1} \frac{M^{(l)}_i(\param)}{R_i^{(l)}(\param) + \varepsilon},\\
  &\bar{P}_j^{(n)}(\param) := \frac{1}{1 - \rho^{n + 1}} \sum_{k = 0}^n \rho^{n - k} (1 - \rho) \nabla_j E_k(\param) \sum_{i = 1}^p \nabla_{i j} E_k(\param) \frac{M^{(n)}_i(\param)}{R_i^{(n)}(\param) + \varepsilon}.
\end{aligned}
\end{equation}
\end{definition}

\begin{assumption}\label{ass:adam-bounds}\leavevmode
  \begin{enumerate}
  \item For some positive constants $M_1$, $M_2$, $M_3$, $M_4$ we have
    \begin{align*}
      &\sup_{i} \sup_k \sup_{\param} \left| \nabla_{i} E_k(\param) \right| \leq M_1,\\
      &\sup_{i, j} \sup_k \sup_{\param} \left| \nabla_{i j} E_k(\param) \right| \leq M_2,\\
      &\sup_{i, j, s} \sup_k \sup_{\param} \left| \nabla_{i j s} E_k(\param) \right| \leq M_3,\\
      &\sup_{i, j, s, r} \sup_k \sup_{\param} \left| \nabla_{i j s r} E_k(\param) \right| \leq M_4.
    \end{align*}

  \item For some $R > 0$ we have for all $n \in \left\{ 0, 1, \dots, \lfloor T / h \rfloor \right\}$
    \begin{equation*}
      R^{(n)}_j \left( \tilde{\param}(t_n) \right) \geq R,\quad \frac{1}{1 - \rho^{n + 1}} \sum_{k = 0}^n \rho^{n - k} (1 - \rho) \left( \nabla_j E_k \left( \tilde{\param}(t_k) \right) \right)^2 \geq R^2,
    \end{equation*}
    where $\tilde{\param}(t)$ is defined in \cref{def:adam-outside}.
  \end{enumerate}
\end{assumption}

\begin{theorem}[Adam with $\varepsilon$ outside: local error bound]\label{th:adam-local-error-bound}
  Suppose \cref{ass:adam-bounds} holds. Then for all $n \in \left\{ 0, 1, \dots, \lfloor T / h \rfloor \right\}$, $j \in \left\{ 1, \ldots, p \right\}$
\begin{equation*}
\left| \tilde{\paramsc}_j(t_{n + 1}) - \tilde{\paramsc}_j(t_n) + h \frac{\frac{1}{1 - \beta^{n + 1}} \sum_{k = 0}^n \beta^{n - k} \left(1 - \beta\right) \nabla_j E_k \left( \tilde{\param}(t_k) \right)}{\sqrt{\frac{1}{1 - \rho^{n + 1}}\sum_{k = 0}^n \rho^{n - k} (1 - \rho) \left( \nabla_j E_k \left( \tilde{\param}(t_k) \right) \right)^2} + \varepsilon} \right| \leq \oldbigc{adamlocalerrorbound} h^3
\end{equation*}
for a positive constant $\newbigc[adamlocalerrorbound]$ depending on $\beta$ and $\rho$.
\end{theorem}

The argument is the same as for \cref{th:local-error-bound}.

\begin{theorem}[Adam with $\varepsilon$ outside: global error bound]\label{th:adam-global-error-bound}
  Suppose \cref{ass:adam-bounds} holds, and
\begin{equation*}
\frac{1}{1 - \rho^{n + 1}} \sum_{k = 0}^n \rho^{n - k} (1 - \rho) \left( \nabla_j E_k \left( \param^{(k)} \right) \right)^2 \geq R^2
\end{equation*}
for $\left\{\param^{(k)}\right\}_{k \in \mathbb{Z}_{\geq 0}}$ defined in \cref{def:adam-outside}. Then there exist positive constants $\newd[adam-outside-d]$, $\newd[adam-outside-dd]$, $\newd[adam-outside-ddd]$ such that for all $n \in \left\{ 0, 1, \dots, \lfloor T / h \rfloor \right\}$
\begin{equation*}
\left\| \mathbf{e}_n\right\| \leq \oldd{adam-outside-d} e^{\oldd{adam-outside-dd} n h} h^2\quad\text{and}\quad \left\| \mathbf{e}_{n + 1} - \mathbf{e}_n \right\| \leq \oldd{adam-outside-ddd} e^{\oldd{adam-outside-dd} n h} h^3,
\end{equation*}
where $\mathbf{e}_n := \tilde{\param}(t_n) - \param^{(n)}$. The constants can be defined as
\begin{align*}
  &\oldd{adam-outside-d} := \oldbigc{adamlocalerrorbound},\\
  &\oldd{adam-outside-dd} := \left[ 1 + \frac{M_2 \sqrt{p}}{R + \varepsilon} \left(\frac{M_1^2}{R (R + \varepsilon)} + 1\right) \oldd{adam-outside-d} \right] \sqrt{p},\\
  &\oldd{adam-outside-ddd} := \oldbigc{adamlocalerrorbound} \oldd{adam-outside-dd}.
\end{align*}
\end{theorem}

\begin{proof}
  Analogously to \cref{th:global-error-bound}, we will prove this by induction over $n$.

  The base case is $n = 0$. Indeed, $\mathbf{e}_0 = \tilde{\param}(0) - \param^{(0)} = \mathbf{0}$. Then the $j$th component of $\mathbf{e}_1 - \mathbf{e}_0$ is
\begin{align*}
  &\left[\mathbf{e}_{1} - \mathbf{e}_{0}\right]_j = \left[\mathbf{e}_{1}\right]_j = \tilde{\paramsc}_j(t_1) - \paramsc_j^{(0)} + \frac{h \nabla_j E_0 \left( \param^{(0)} \right)}{\left| \nabla_j E_0 \left( \param^{(0)} \right) \right| + \varepsilon}\\
  &\quad = \tilde{\paramsc}_j(t_1) - \tilde{\paramsc}_j(t_0) + \frac{h \nabla_j E_0 \left( \tilde{\param}(t_0) \right)}{\sqrt{\left( \nabla_j E_0 \left( \tilde{\param}(t_0) \right) \right)^2} + \varepsilon}.
\end{align*}
By \cref{th:adam-local-error-bound}, the absolute value of the right-hand side does not exceed $\oldbigc{adamlocalerrorbound} h^3$, which means $\left\|\mathbf{e}_1 - \mathbf{e}_0\right\| \leq \oldbigc{adamlocalerrorbound} h^3 \sqrt{p}$. Since $\oldbigc{adamlocalerrorbound} \sqrt{p} \leq \oldd{adam-outside-ddd}$, the base case is proven.

Now suppose that for all $k = 0, 1, \ldots, n - 1$ the claim
\begin{equation*}
\left\| \mathbf{e}_k\right\| \leq \oldd{adam-outside-d} e^{\oldd{adam-outside-dd} k h} h^2\quad\text{and}\quad \left\| \mathbf{e}_{k + 1} - \mathbf{e}_k \right\| \leq \oldd{adam-outside-ddd} e^{\oldd{adam-outside-dd} k h} h^3
\end{equation*}
is proven. Then
\begin{align*}
  &\left\| \mathbf{e}_n \right\| \labrel{1}[\leq] \left\| \mathbf{e}_{n - 1} \right\| + \left\| \mathbf{e}_n - \mathbf{e}_{n - 1} \right\| \leq \oldd{adam-outside-d} e^{\oldd{adam-outside-dd} (n - 1) h} h^2 + \oldd{adam-outside-ddd} e^{\oldd{adam-outside-dd} (n - 1) h} h^3\\
  &\quad = \oldd{adam-outside-d} e^{\oldd{adam-outside-dd} (n - 1) h} h^2 \left( 1 + \frac{\oldd{adam-outside-ddd}}{\oldd{adam-outside-d}} h \right) \labrel{2}[\leq] \oldd{adam-outside-d} e^{\oldd{adam-outside-dd} (n - 1) h} h^2 \left( 1 + \oldd{adam-outside-dd} h \right)\\
  &\quad \labrel{3}[\leq] \oldd{adam-outside-d} e^{\oldd{adam-outside-dd} (n - 1) h} h^2 \cdot e^{\oldd{adam-outside-dd} h} = \oldd{adam-outside-d} e^{\oldd{adam-outside-dd} n h} h^2,
\end{align*}
where \labrel{1} is by the triangle inequality, \labrel{2} is by $\oldd{adam-outside-ddd} / \oldd{adam-outside-d} \leq \oldd{adam-outside-dd}$, in \labrel{3} we used $1 + x \leq e^x$ for all $x \geq 0$.

Next, combining \cref{th:adam-local-error-bound} with~\eqref{eq:adam-iteration}, we have
\begin{equation}\label{eq:adam-global-start-bounding-e-n-pl-one-min-e-n}
  \left| \left[\mathbf{e}_{n + 1} - \mathbf{e}_n\right]_j \right| \leq \oldbigc{adamlocalerrorbound} h^3 + h \left| \frac{N'}{\sqrt{D'} + \varepsilon} - \frac{N''}{\sqrt{D''} + \varepsilon} \right|,
\end{equation}
where to simplify notation we put
\begin{align*}
  &N' := \frac{1}{1 - \beta^{n + 1}} \sum_{k = 0}^n \beta^{n - k} (1 - \beta) \nabla_j E_k \left( \param^{(k)} \right),\\
  &N'' := \frac{1}{1 - \beta^{n + 1}} \sum_{k = 0}^n \beta^{n - k} (1 - \beta) \nabla_j E_k \left( \tilde{\param}(t_k) \right),\\
  &D' := \frac{1}{1 - \rho^{n + 1}} \sum_{k = 0}^n \rho^{n - k} (1 - \rho) \left( \nabla_j E_k \left( \param^{(k)} \right) \right)^2,\\
  &D'' := \frac{1}{1 - \rho^{n + 1}} \sum_{k = 0}^n \rho^{n - k} (1 - \rho) \left( \nabla_j E_k \left( \tilde{\param}(t_k) \right) \right)^2.
\end{align*}

Using $D' \geq R^2$, $D'' \geq R^2$, we have
\begin{equation}\label{eq:adam-global-diff-of-frac-bound}
\left| \frac{1}{\sqrt{D'} + \varepsilon} - \frac{1}{\sqrt{D''} + \varepsilon}\right| = \frac{\left| D' - D'' \right|}{\left( \sqrt{D'} + \varepsilon \right) \left( \sqrt{D''} + \varepsilon \right) \left( \sqrt{D'} + \sqrt{D''} \right)} \leq \frac{\left| D' - D'' \right|}{2 R \left( R + \varepsilon \right)^2}.
\end{equation}
But since
\begin{align*}
  &\left| \left( \nabla_j E_k \left( \param^{(k)} \right) \right)^2 - \left( \nabla_j E_k \left( \tilde{\param}(t_k) \right) \right)^2 \right|\\
  &\quad = \left| \nabla_j E_k \left( \param^{(k)} \right) - \nabla_j E_k \left( \tilde{\param}(t_k) \right) \right| \cdot \left| \nabla_j E_k \left( \param^{(k)} \right) + \nabla_j E_k \left( \tilde{\param}(t_k) \right) \right|\\
  &\quad \leq 2 M_1 \left| \nabla_j E_k \left( \param^{(k)} \right) - \nabla_j E_k \left( \tilde{\param}(t_k) \right) \right| \leq 2 M_1 M_2 \sqrt{p} \left\| \param^{(k)} - \tilde{\param}(t_k) \right\|,
\end{align*}
we have
\begin{equation}\label{eq:adam-d-diff-bound}
\left| D' - D'' \right| \leq \frac{2 M_1 M_2 \sqrt{p}}{1 - \rho^{n + 1}} \sum_{k = 0}^n \rho^{n - k} (1 - \rho) \left\|  \param^{(k)} - \tilde{\param}(t_k) \right\|.
\end{equation}

Similarly,
\begin{equation}\label{eq:adam-n-diff-bound}
\begin{aligned}
  &\left| N' - N'' \right| \leq \frac{1}{1 - \beta^{n + 1}} \sum_{k = 0}^n \beta^{n - k} (1 - \beta) \left| \nabla_j E_k \left( \param^{(k)} \right) - \nabla_j E_k \left( \tilde{\param}(t_k) \right) \right|\\
  &\quad \leq \frac{1}{1 - \beta^{n + 1}} \sum_{k = 0}^n \beta^{n - k} (1 - \beta) M_2 \sqrt{p} \left\| \param^{(k)} - \tilde{\param}(t_k) \right\|.
\end{aligned}
\end{equation}

Combining \eqref{eq:adam-global-diff-of-frac-bound}, \eqref{eq:adam-d-diff-bound} and \eqref{eq:adam-n-diff-bound}, we get
\begin{align}
  &\left| \frac{N'}{\sqrt{D'} + \varepsilon} - \frac{N''}{\sqrt{D''} + \varepsilon} \right| \leq \left| N' \right| \cdot \left| \frac{1}{\sqrt{D'} + \varepsilon} - \frac{1}{\sqrt{D''} + \varepsilon}\right| + \frac{\left| N' - N'' \right|}{\sqrt{D''} + \varepsilon}\nonumber\\
  &\quad \leq \frac{1}{1 - \beta^{n + 1}} \sum_{k = 0}^n \beta^{n - k} (1 - \beta) M_1 \cdot \frac{2 M_1 M_2 \sqrt{p}}{2 R (R + \varepsilon)^2 \left(1 - \rho^{n + 1}\right)} \sum_{k = 0}^n \rho^{n - k} (1 - \rho) \left\|  \param^{(k)} - \tilde{\param}(t_k) \right\|\nonumber\\
  &\qquad + \frac{M_2 \sqrt{p}}{(R + \varepsilon) \left(1 - \beta^{n + 1}\right)} \sum_{k = 0}^n \beta^{n - k} (1 - \beta) \left\| \param^{(k)} - \tilde{\param}(t_k) \right\|\nonumber\\
  &\quad = \frac{M_1^2 M_2 \sqrt{p}}{R (R + \varepsilon)^2 \left(1 - \rho^{n + 1}\right)} \sum_{k = 0}^n \rho^{n - k} (1 - \rho) \left\|  \param^{(k)} - \tilde{\param}(t_k) \right\|\nonumber\\
  &\qquad + \frac{M_2 \sqrt{p}}{(R + \varepsilon) \left(1 - \beta^{n + 1}\right)} \sum_{k = 0}^n \beta^{n - k} (1 - \beta) \left\| \param^{(k)} - \tilde{\param}(t_k) \right\|\nonumber\\
  &\quad \labrel{1}[\leq] \frac{M_1^2 M_2 \sqrt{p}}{R (R + \varepsilon)^2 \left(1 - \rho^{n + 1}\right)} \sum_{k = 0}^n \rho^{n - k} (1 - \rho) \oldd{adam-outside-d} e^{\oldd{adam-outside-dd} k h} h^2\nonumber\\
  &\qquad + \frac{M_2 \sqrt{p}}{(R + \varepsilon) \left(1 - \beta^{n + 1}\right)} \sum_{k = 0}^n \beta^{n - k} (1 - \beta) \oldd{adam-outside-d} e^{\oldd{adam-outside-dd} k h} h^2,\label{eq:adam-diff-of-two-frac-before-applying-induction}
\end{align}
where in \labrel{1} we used the induction hypothesis and that the bound on $\left\| \mathbf{e}_n \right\|$ is already proven.

Now note that since $0 < \rho e^{-\oldd{adam-outside-dd} h} < \rho$, we have $\sum_{k = 0}^n \left( \rho e^{-\oldd{adam-outside-dd} h} \right)^k \leq \sum_{k = 0}^n \rho^k = \left(1 - \rho^{n + 1}\right) / \left(1 - \rho\right)$, which is rewritten as
\begin{equation*}
\frac{1}{1 - \rho^{n + 1}} \sum_{k = 0}^n \rho^{n - k} (1 - \rho) e^{\oldd{adam-outside-dd} k h} \leq e^{\oldd{adam-outside-dd} n h}.
\end{equation*}
By the same logic,
\begin{equation*}
\frac{1}{1 - \beta^{n + 1}} \sum_{k = 0}^n \beta^{n - k} (1 - \beta) e^{\oldd{adam-outside-dd} k h} \leq e^{\oldd{adam-outside-dd} n h}.
\end{equation*}
Then we can continue~\eqref{eq:adam-diff-of-two-frac-before-applying-induction}:
\begin{equation}\label{eq:adam-global-big-frac-dif-final-bound}
\left| \frac{N'}{\sqrt{D'} + \varepsilon} - \frac{N''}{\sqrt{D''} + \varepsilon} \right| \leq \frac{M_2 \sqrt{p}}{R + \varepsilon} \left( \frac{M_1^2}{R (R + \varepsilon)} + 1 \right) \oldd{adam-outside-d} e^{\oldd{adam-outside-dd} n h} h^2
\end{equation}
Again using $1 \leq e^{\oldd{adam-outside-dd} n h}$, we conclude from~\eqref{eq:adam-global-start-bounding-e-n-pl-one-min-e-n} and~\eqref{eq:adam-global-big-frac-dif-final-bound} that
\begin{equation*}
\left\| \mathbf{e}_{n + 1} - \mathbf{e}_n \right\| \leq \underbrace{\left( \oldbigc{adamlocalerrorbound} + \frac{M_2 \sqrt{p}}{R + \varepsilon} \left(\frac{M_1^2}{R (R + \varepsilon)} + 1\right) \oldd{adam-outside-d}\right) \sqrt{p}}_{\leq \oldd{adam-outside-ddd}} e^{\oldd{adam-outside-dd} n h} h^3,
\end{equation*}
finishing the induction step.
\end{proof}

\section{Adam with $\varepsilon$ Inside the Square Root}

\begin{definition}\label{def:adam-inside}
  In this section, for some $\param^{(0)} \in \mathbb{R}^p$, $\mathbf{\nu}^{(0)} = \mathbf{0} \in \mathbb{R}^p$, $\beta, \rho \in (0, 1)$, let the sequence of $p$-vectors $\left\{\param^{(k)}\right\}_{k \in \mathbb{Z}_{\geq 0}}$ be defined for $n \geq 0$ by
\begin{equation}\label{eq:mod-adam-iteration}
\begin{aligned}
  &\nu_j^{(n + 1)} = \rho \nu_j^{(n)} + (1 - \rho) \left( \nabla_j E_n\left(\param^{(n)}\right) \right)^2,\\
  &m_j^{(n + 1)} = \beta m_j^{(n)} + (1 - \beta) \nabla_j E_n \left( \param^{(n)} \right),\\
  &\paramsc_j^{(n + 1)} = \paramsc_j^{(n)} - h \frac{m_j^{(n + 1)} / \left(1 - \beta^{n + 1}\right)}{\sqrt{\nu_j^{(n + 1)} / \left( 1 - \rho^{n + 1} \right) + \varepsilon}}.
\end{aligned}
\end{equation}

Let $\tilde{\param}(t)$ be defined as a continuous solution to the piecewise ODE
\begin{equation}\label{eq:mod-adam-nth-step-modified-equation}
\begin{aligned}
  &\dot{\tilde{\paramsc}}_j(t) = - \frac{M_j^{(n)} \left( \tilde{\param}(t) \right)}{R^{(n)}_j\left(\tilde{\param}(t)\right)}\\
  &\qquad + h \left( \frac{M_j^{(n)}\left( \tilde{\param}(t) \right) \left(2 P_j^{(n)}\left(\tilde{\param}(t)\right) + \bar{P}_j^{(n)}\left(\tilde{\param}(t)\right)\right)}{2 R_j^{(n)} \left( \tilde{\param}(t) \right)^3} - \frac{2 L_j^{(n)} \left( \tilde{\param}(t) \right) + \bar{L}_j^{(n)} \left( \tilde{\param}(t) \right)}{2 R_j^{(n)}\left(\tilde{\param}(t)\right)} \right)
\end{aligned}
\end{equation}
for $t \in [t_n, t_{n + 1}]$ with the initial condition $\tilde{\param}(0) = \param^{(0)}$, where $\mathbf{R}^{(n)}(\param)$, $\mathbf{P}^{(n)}(\param)$, $\bar{\mathbf{P}}^{(n)}(\param)$, $\mathbf{M}^{(n)}(\param)$, $\mathbf{L}^{(n)}(\param)$, $\bar{\mathbf{L}}^{(n)}(\param)$ are $p$-dimensional functions with components
\begin{equation}\label{eq:r-p-modified-adam}
\begin{aligned}
  &R^{(n)}_j(\param) := \sqrt{\sum_{k = 0}^n \rho^{n - k} (1 - \rho) \left( \nabla_j E_k(\param) \right)^2 / \left(1 - \rho^{n + 1}\right) + \varepsilon},\\
  &M^{(n)}_j(\param) := \frac{1}{1 - \beta^{n + 1}} \sum_{k = 0}^n \beta^{n - k} \left(1 - \beta\right) \nabla_j E_k \left( \param \right),\\
  &L_j^{(n)}(\param) := \frac{1}{1 - \beta^{n + 1}} \sum_{k = 0}^n \beta^{n - k} (1 - \beta) \sum_{i = 1}^p \nabla_{i j} E_k(\param) \sum_{l = k}^{n - 1} \frac{M^{(l)}_i(\param)}{R_i^{(l)}(\param)},\\
  &\bar{L}_j^{(n)}(\param) := \frac{1}{1 - \beta^{n + 1}} \sum_{k = 0}^n \beta^{n - k} (1 - \beta) \sum_{i = 1}^p \nabla_{i j} E_k(\param) \frac{M^{(n)}_i(\param)}{R_i^{(n)}(\param)},\\
  &P_j^{(n)}(\param) := \frac{1}{1 - \rho^{n + 1}} \sum_{k = 0}^n \rho^{n - k} (1 - \rho) \nabla_j E_k(\param) \sum_{i = 1}^p \nabla_{i j} E_k(\param) \sum_{l = k}^{n - 1} \frac{M^{(l)}_i(\param)}{R_i^{(l)}(\param)},\\
  &\bar{P}_j^{(n)}(\param) := \frac{1}{1 - \rho^{n + 1}} \sum_{k = 0}^n \rho^{n - k} (1 - \rho) \nabla_j E_k(\param) \sum_{i = 1}^p \nabla_{i j} E_k(\param) \frac{M^{(n)}_i(\param)}{R_i^{(n)}(\param)}.
\end{aligned}
\end{equation}
\end{definition}

\begin{assumption}\label{ass:mod-adam-bounds} For some positive constants $M_1$, $M_2$, $M_3$, $M_4$ we have
  \begin{align*}
    &\sup_{i} \sup_k \sup_{\param} \left| \nabla_{i} E_k(\param) \right| \leq M_1,\\
    &\sup_{i, j} \sup_k \sup_{\param} \left| \nabla_{i j} E_k(\param) \right| \leq M_2,\\
    &\sup_{i, j, s} \sup_k \sup_{\param} \left| \nabla_{i j s} E_k(\param) \right| \leq M_3,\\
    &\sup_{i, j, s, r} \sup_k \sup_{\param} \left| \nabla_{i j s r} E_k(\param) \right| \leq M_4.
  \end{align*}
\end{assumption}

\begin{theorem}[Adam with $\varepsilon$ inside: local error bound]\label{th:adam-inside-local-error-bound}
  Suppose \cref{ass:mod-adam-bounds} holds. Then for all $n \in \left\{ 0, 1, \dots, \lfloor T / h \rfloor \right\}$, $j \in \left\{ 1, \ldots, p \right\}$
\begin{equation*}
\left| \tilde{\paramsc}_j(t_{n + 1}) - \tilde{\paramsc}_j(t_n) + h \frac{\frac{1}{1 - \beta^{n + 1}} \sum_{k = 0}^n \beta^{n - k} \left(1 - \beta\right) \nabla_j E_k \left( \tilde{\param}(t_k) \right)}{\sqrt{\frac{1}{1 - \rho^{n + 1}}\sum_{k = 0}^n \rho^{n - k} (1 - \rho) \left( \nabla_j E_k \left( \tilde{\param}(t_k) \right) \right)^2 + \varepsilon}} \right| \leq \oldbigc{modadamlocalerrorbound} h^3
\end{equation*}
for a positive constant $\newbigc[modadamlocalerrorbound]$ depending on $\beta$ and $\rho$.
\end{theorem}

The argument is the same as for \cref{th:local-error-bound}.

\begin{theorem}[Adam with $\varepsilon$ inside: global error bound]\label{th:mod-adam-inside-global-error-bound}
  Suppose \cref{ass:mod-adam-bounds} holds for $\left\{\param^{(k)}\right\}_{k \in \mathbb{Z}_{\geq 0}}$ defined in \cref{def:adam-inside}. Then there exist positive constants $\newd[adam-inside-d]$, $\newd[adam-inside-dd]$, $\newd[adam-inside-ddd]$ such that for all $n \in \left\{ 0, 1, \dots, \lfloor T / h \rfloor \right\}$
\begin{equation*}
\left\| \mathbf{e}_n\right\| \leq \oldd{adam-inside-d} e^{\oldd{adam-inside-dd} n h} h^2\quad\text{and}\quad \left\| \mathbf{e}_{n + 1} - \mathbf{e}_n \right\| \leq \oldd{adam-inside-ddd} e^{\oldd{adam-inside-dd} n h} h^3,
\end{equation*}
where $\mathbf{e}_n := \tilde{\param}(t_n) - \param^{(n)}$. The constants can be defined as
\begin{align*}
  &\oldd{adam-inside-d} := \oldbigc{modadamlocalerrorbound},\\
  &\oldd{adam-inside-dd} := \left[ 1 + \frac{M_2 \sqrt{p}}{\sqrt{\varepsilon}} \left(\frac{M_1^2}{\varepsilon} + 1\right) \oldd{adam-inside-d} \right] \sqrt{p},\\
  &\oldd{adam-inside-ddd} := \oldbigc{modadamlocalerrorbound} \oldd{adam-inside-dd}.
\end{align*}
\end{theorem}

The argument is the same as for \cref{th:adam-global-error-bound}.

\section{Bounding the Derivatives of the ODE Solution}\label{sec:technical-bounding-lemmas}

Our first goal is to argue that the first derivative of $t \mapsto \tilde{\paramsc}_j(t)$ is uniformly bounded in absolute value. To achieve this, we just need to bound all the terms on the right-hand side of the ODE~\eqref{eq:nth-step-modified-equation}.

\begin{lemma}\label{lem:trivial-p-bounds}
  Suppose \cref{ass:bounds} holds. Then for all $n \in \left\{ 0, 1, \dots, \lfloor T / h \rfloor \right\}$
  \begin{align}
    &\sup_{\param} \left| P_j^{(n)}(\param) \right| \leq \oldbigc{p-bound},\label{eq:p-bound}\\
    &\sup_{\param} \left| \bar{P}_j^{(n)}(\param)  \right| \leq \oldbigc{p-bar-bound},\label{eq:p-bar-bound}
  \end{align}
  with constants $\newbigc[p-bound]$, $\newbigc[p-bar-bound]$ defined as follows:
  \begin{align*}
    &\oldbigc{p-bound} := p \frac{M_1^2 M_2}{R + \varepsilon} \cdot \frac{\rho}{1 - \rho},\\
    &\oldbigc{p-bar-bound} := p \frac{M_1^2 M_2}{R + \varepsilon}.
  \end{align*}
\end{lemma}

\begin{proof}[Proof of \cref{lem:trivial-p-bounds}] Both bounds are straightforward:
  \begin{align*}
    &\sup_{\param} \left| P_j^{(n)}(\param) \right| = \sup_{\param} \left| \sum_{k = 0}^n \rho^{n - k} (1 - \rho) \nabla_j E_k \left( \param \right) \sum_{i = 1}^p \nabla_{i j} E_k \left( \param \right) \sum_{l = k}^{n - 1} \frac{\nabla_i E_l \left( \param \right)}{R_i^{(l)} \left( \param \right) + \varepsilon} \right|\\
    &\quad \leq p \frac{M_1^2 M_2}{R + \varepsilon} (1 - \rho) \sum_{k = 0}^n \rho^{n - k} (n - k) \leq p \frac{M_1^2 M_2}{R + \varepsilon} (1 - \rho) \sum_{k = 0}^\infty \rho^k k = \oldbigc{p-bound}.
  \end{align*}
and
  \begin{align*}
    &\sup_{\param} \left| \bar{P}_j^{(n)}(\param)  \right| = \sup_{\param} \left| \sum_{k = 0}^n \rho^{n - k} (1 - \rho) \nabla_j E_k \left( \param \right) \sum_{i = 1}^p \nabla_{i j} E_k \left( \param \right) \frac{\nabla_i E_n \left( \param \right)}{R_i^{(n)} \left( \param \right) + \varepsilon} \right|\\
    &\quad \leq p \frac{M_1^2 M_2}{R + \varepsilon} (1 - \rho) \sum_{k = 0}^n \rho^{n - k} \leq p \frac{M_1^2 M_2}{R + \varepsilon} = \oldbigc{p-bar-bound},
  \end{align*}
concluding the proof of \cref{lem:trivial-p-bounds}.
\end{proof}

\begin{lemma}
Suppose \cref{ass:bounds} holds. Then the first derivative of $t \mapsto \tilde{\paramsc}_j(t)$ is uniformly over $j$ and $t \in [0, T]$ bounded in absolute value by some positive constant, say $D_1$.
\end{lemma}

\begin{proof}
This follows immediately from $h \leq T$, \eqref{eq:p-bound}, \eqref{eq:p-bar-bound} and the definition of $\tilde{\param}(t)$ given in~\eqref{eq:nth-step-modified-equation}.
\end{proof}

Our next goal is to argue that the \textit{second} derivative of $t \mapsto \tilde{\paramsc}_j(t)$ is bounded in absolute value. For this, we need to bound the first derivatives of all the three additive terms on the right-hand side of~\eqref{eq:nth-step-modified-equation}.

\begin{lemma}\label{lem:first-derivatives}
  Suppose \cref{ass:bounds} holds. Then for all $n, k \in \left\{ 0, 1, \dots, \lfloor T / h \rfloor \right\}$, $j \in \left\{ 1, \ldots, p \right\}$ we have
\begin{align}
  &\sup_{t \in [0, T]} \left| \timeder{\nabla_j E_n \left( \tilde{\param}(t) \right)} \right| \leq \oldbigc{e-first-der},\label{eq:e-first-der-bound}\\
  &\sup_{t \in [t_n, t_{n + 1}]} \left| \sum_{i = 1}^p \nabla_{i j} E_k \left( \tilde{\param}(t) \right) \left[ \dot{\tilde{\paramsc}}_i(t) + \frac{\nabla_i E_n \left( \tilde{\param}(t) \right)}{R_i^{(n)} \left( \tilde{\param}(t) \right) + \varepsilon} \right] \right| \leq \oldbigc{sol-der-first-order} h\label{eq:sol-der-first-order},\\
  &\sup_{t \in [0, T]} \left| \sum_{i = 1}^p \nabla_{i j} E_k \left( \tilde{\param}(t) \right) \sum_{l = k}^{n - 1} \frac{\nabla_i E_l \left( \tilde{\param}(t) \right)}{R_i^{(l)} \left( \tilde{\param}(t) \right) + \varepsilon} \right| \leq (n - k) \oldbigc{second-first-grad-over-r-partial-sum}\quad\text{for $k < n$,}\label{eq:second-first-grad-over-r-partial-sum-bound}\\
  &\sup_{t \in [0, T]} \left| \timeder{P_j^{(n)} \left( \tilde{\param}(t) \right)} \right| \leq \oldbigc{der-p-first-term} + \oldbigc{der-p-second-term},\label{eq:der-p-bound}\\
  &\sup_{t \in [0, T]} \left| \timeder{\bar{P}_j^{(n)}(\tilde{\param}(t))}\right| \leq \oldbigc{der-bar-p},\label{eq:der-p-bar-bound}\\
    &\sup_{t \in [0, T]} \left| \timeder{\sum_{i = 1}^p \nabla_{i j} E_k \left( \tilde{\param}(t) \right) \frac{\nabla_i E_n\left(\tilde{\param}(t)\right)}{R^{(n)}_i\left(\tilde{\param}(t)\right)  + \varepsilon}} \right| \leq \oldbigc{der-p-second-term-aux-two}\label{eq:nabla-two-nabla-one-over-r-eps},\\
  &\sup_{t \in [0, T]} \left|\timeder{\frac{\nabla_j E_n \left( \tilde{\param}(t) \right) \left(2 P_j^{(n)}\left(\tilde{\param}(t)\right) + \bar{P}_j^{(n)}\left(\tilde{\param}(t)\right)\right)}{2 \left( R_j^{(n)} \left( \tilde{\param}(t) \right) + \varepsilon \right)^2 R_j^{(n)} \left( \tilde{\param}(t) \right)}}\right| \leq \oldbigc{der-huge-func-first-term},\label{eq:der-huge-func-first-term-bound}\\
  &\sup_{t \in [0, T]} \left| \timeder{\frac{\sum_{i = 1}^p \nabla_{i j} E_n \left( \tilde{\param}(t) \right) \frac{\nabla_i E_n \left( \tilde{\param}(t) \right)}{R_i^{(n)}\left(\tilde{\param}(t)\right) + \varepsilon}}{2 \left( R_j^{(n)}(\tilde{\param}(t)) + \varepsilon \right)}} \right| \leq \oldbigc{der-huge-func-second-term},\label{eq:der-huge-func-second-term-bound}
\end{align}
with constants $\newbigc[e-first-der]$, $\newbigc[sol-der-first-order]$, $\newbigc[second-first-grad-over-r-partial-sum]$, $\newbigc[der-p-first-term]$, $\newbigc[der-r]$, $\newbigc[der-p-second-term-aux-one]$, $\newbigc[der-p-second-term-aux-two]$, $\newbigc[der-p-second-term]$, $\newbigc[der-bar-p]$, $\newbigc[der-r-pl-eps-sq-r-inverse]$, $\newbigc[der-huge-func-first-term]$, $\newbigc[der-huge-func-second-term]$ defined as follows:
  \begin{align*}
    &\oldbigc{e-first-der} := p M_2 D_1,\\
    &\oldbigc{sol-der-first-order} := p M_2 \left[\frac{M_1 \left( 2 \oldbigc{p-bound} + \oldbigc{p-bar-bound} \right)}{2 (R + \varepsilon)^2 R} + \frac{p M_1 M_2}{2 (R + \varepsilon)^2} \right],\\
    &\oldbigc{second-first-grad-over-r-partial-sum} := p \frac{M_1 M_2}{R + \varepsilon},\\
    &\oldbigc{der-p-first-term} := D_1 p^2 \frac{M_1 M_2^2}{R + \varepsilon} \cdot \frac{\rho}{1 - \rho},\\
    &\oldbigc{der-r} := \frac{D_1 p M_1 M_2}{R},\\
    &\oldbigc{der-p-second-term-aux-one} := D_1 p^2 \frac{M_1 M_3}{R + \varepsilon},\\
    &\oldbigc{der-p-second-term-aux-two} := \oldbigc{der-p-second-term-aux-one} + p M_2 \left( \frac{D_1 p M_2}{R + \varepsilon} + \frac{M_1}{(R + \varepsilon)^2} \oldbigc{der-r} \right)\\
    &\quad = \frac{D_1 p^2}{R + \varepsilon} \left( M_1 M_3 + M_2^2 + \frac{M_1^2 M_2^2}{(R + \varepsilon) R} \right),\\
    &\oldbigc{der-p-second-term} := M_1 \oldbigc{der-p-second-term-aux-two} \frac{\rho}{1 - \rho},\\
    &\oldbigc{der-bar-p} := \frac{D_1 p^2 M_1 M_2^2}{R + \varepsilon} + \frac{D_1 p^2 M_1^2 M_3}{R + \varepsilon} + \frac{D_1 p^2 M_1 M_2^2}{R + \varepsilon} + \frac{p M_1^2 M_2 \oldbigc{der-r}}{(R + \varepsilon)^2},\\
    &\oldbigc{der-r-pl-eps-sq-r-inverse} := \frac{2 \oldbigc{der-r}}{R (R + \varepsilon)^3} + \frac{\oldbigc{der-r}}{(R + \varepsilon)^4},\\
    &\oldbigc{der-huge-func-first-term} := \frac{D_1 p M_2 \cdot \left( 2 \oldbigc{p-bound} + \oldbigc{p-bar-bound} \right)}{2 \left( R + \varepsilon \right)^2 R} + \frac{M_1 \left( 2 \left( \oldbigc{der-p-first-term} + \oldbigc{der-p-second-term} \right) + \oldbigc{der-bar-p} \right)}{2 \left( R + \varepsilon \right)^2 R} + \frac{M_1 \left( 2 \oldbigc{p-bound} + \oldbigc{p-bar-bound} \right) \oldbigc{der-r-pl-eps-sq-r-inverse}}{2},\\
    &\oldbigc{der-huge-func-second-term} := \frac{1}{2(R + \varepsilon)} \left( \frac{p^2 D_1 M_1 M_3}{R + \varepsilon} + \frac{p^2 D_1 M_2^2}{R + \varepsilon} + \frac{p M_1 M_2 \oldbigc{der-r}}{\left( R + \varepsilon \right)^2} \right) + \frac{1}{2} \cdot \frac{p M_1 M_2}{R + \varepsilon} \cdot \frac{\oldbigc{der-r}}{(R + \varepsilon)^2}.
  \end{align*}
\end{lemma}

\begin{proof}[Proof of \cref{lem:first-derivatives}] We prove the inequalities one by one.

  The bound~\eqref{eq:e-first-der-bound} is straightforward:
\begin{equation*}
\left| \timeder{\nabla_j E_n \left( \tilde{\param}(t) \right)} \right| = \left| \sum_{i = 1}^p \nabla_{i j} E_n \left( \tilde{\param}(t) \right) \dot{\tilde{\paramsc}}_i(t) \right| \leq \oldbigc{e-first-der}.
\end{equation*}

  The inequality \eqref{eq:sol-der-first-order} follows immediately from the fact that by~\eqref{eq:nth-step-modified-equation} we have for $t \in [t_n, t_{n + 1}]$
\begin{equation*}
\left| \dot{\tilde{\paramsc}}_j(t) + \frac{\nabla_j E_n\left(\tilde{\param}(t)\right)}{R_j^{(n)} \left( \tilde{\param}(t) \right) + \varepsilon} \right| \le h \left[\frac{M_1 \left( 2 \oldbigc{p-bound} + \oldbigc{p-bar-bound} \right)}{2 (R + \varepsilon)^2 R} + \frac{p M_1 M_2}{2 (R + \varepsilon)^2} \right].
\end{equation*}

The bound~\eqref{eq:second-first-grad-over-r-partial-sum-bound} follows from the assumptions immediately.

We will prove~\eqref{eq:der-p-bound} by bounding the two additive terms on the right-hand side of the equality
  \begin{equation}\label{eq:der-p-into-two-terms}
    \begin{aligned}
      &\frac{\mathrm{d}}{\mathrm{d} t} P_j^{(n)} \left( \tilde{\param}(t) \right)\\
      &\quad = \sum_{k = 0}^n \rho^{n - k} (1 - \rho) \sum_{u = 1}^p \nabla_{j u} E_k \left( \tilde{\param}(t) \right) \dot{\tilde{\paramsc}}_u(t) \sum_{i = 1}^p \nabla_{i j} E_k \left( \tilde{\param}(t) \right) \sum_{l = k}^{n - 1} \frac{\nabla_i E_l \left( \tilde{\param}(t) \right)}{R_i^{(l)} \left( \tilde{\param}(t) \right) + \varepsilon}\\
      &\qquad + \sum_{k = 0}^n \rho^{n - k} (1 - \rho) \nabla_j E_k \left( \tilde{\param}(t) \right) \sum_{i = 1}^p \frac{\mathrm{d}}{\mathrm{d} t} \left\{ \nabla_{i j} E_k \left( \tilde{\param}(t) \right) \sum_{l = k}^{n - 1} \frac{\nabla_i E_l \left( \tilde{\param}(t) \right)}{R_i^{(l)} \left( \tilde{\param}(t) \right) + \varepsilon} \right\}.
    \end{aligned}
  \end{equation}

  It is easily shown that the first term in~\eqref{eq:der-p-into-two-terms} is bounded in absolute value by $\oldbigc{der-p-first-term}$:
  \begin{align*}
    &\left| \sum_{k = 0}^n \rho^{n - k} (1 - \rho) \sum_{u = 1}^p \nabla_{j u} E_k \left( \tilde{\param}(t) \right) \dot{\tilde{\paramsc}}_u(t) \sum_{i = 1}^p \nabla_{i j} E_k \left( \tilde{\param}(t) \right) \sum_{l = k}^{n - 1} \frac{\nabla_i E_l \left( \tilde{\param}(t) \right)}{R_i^{(l)} \left( \tilde{\param}(t) \right) + \varepsilon}\right|\\
    &\quad \leq D_1 p^2 \frac{M_1 M_2^2}{R + \varepsilon} (1 - \rho) \sum_{k = 0}^n \rho^k k\\
    &\quad \le D_1 p^2 \frac{M_1 M_2^2}{R + \varepsilon} (1 - \rho) \sum_{k = 0}^\infty \rho^k k\\
    &\quad = \oldbigc{der-p-first-term}.
  \end{align*}

  For the proof of~\eqref{eq:der-p-bound}, it is left to show that the second term in~\eqref{eq:der-p-into-two-terms} is bounded in absolute value by $\oldbigc{der-p-second-term}$.

  To bound $\sum_{i = 1}^p \frac{\mathrm{d}}{\mathrm{d} t} \left\{ \nabla_{i j} E_k \left( \tilde{\param}(t) \right) \sum_{l = k}^{n - 1} \frac{\nabla_i E_l \left( \tilde{\param}(t) \right)}{R_i^{(l)} \left( \tilde{\param}(t) \right) + \varepsilon} \right\}$, we can use
  \begin{align*}
    &\left| \sum_{i = 1}^p \frac{\mathrm{d}}{\mathrm{d} t} \left\{ \nabla_{i j} E_k \left( \tilde{\param}(t) \right) \sum_{l = k}^{n - 1} \frac{\nabla_i E_l \left( \tilde{\param}(t) \right)}{R_i^{(l)} \left( \tilde{\param}(t) \right) + \varepsilon} \right\} \right|\\
    &\quad \leq \left|\sum_{i = 1}^p \frac{\mathrm{d}}{\mathrm{d} t} \left\{ \nabla_{i j} E_k \left( \tilde{\param}(t) \right) \right\} \sum_{l = k}^{n - 1} \frac{\nabla_i E_l \left( \tilde{\param}(t) \right)}{R_i^{(l)} \left( \tilde{\param}(t) \right) + \varepsilon}\right|\\
    &\qquad + \left|\sum_{i = 1}^p \nabla_{i j} E_k \left( \tilde{\param}(t) \right) \sum_{l = k}^{n - 1} \frac{\mathrm{d}}{\mathrm{d} t} \left\{ \frac{\nabla_i E_l \left( \tilde{\param}(t) \right)}{R_i^{(l)} \left( \tilde{\param}(t) \right) + \varepsilon} \right\}\right|
  \end{align*}

  By the Cauchy-Schwarz inequality applied twice,
  \begin{align*}
    &\left|\sum_{i = 1}^p \frac{\mathrm{d}}{\mathrm{d} t} \left\{ \nabla_{i j} E_k \left( \tilde{\param}(t) \right) \right\} \sum_{l = k}^{n - 1} \frac{\nabla_i E_l \left( \tilde{\param}(t) \right)}{R_i^{(l)} \left( \tilde{\param}(t) \right) + \varepsilon}\right|\\
    &\quad \leq \sqrt{\sum_{i = 1}^p \sum_{s = 1}^p \left( \nabla_{i j s} E_k \left( \tilde{\param}(t) \right) \right)^2} \sqrt{\sum_{u = 1}^p \dot{\tilde{\paramsc}}_u(t)^2} \sqrt{\sum_{i = 1}^p \left| \sum_{l = k}^{n - 1} \frac{\nabla_i E_l \left( \tilde{\param}(t) \right)}{R_i^{(l)} \left( \tilde{\param}(t) \right) + \varepsilon}\right|^2}\\
    &\quad \leq M_3 p \cdot D_1\sqrt{p} \cdot \sqrt{\sum_{i = 1}^p \left| \sum_{l = k}^{n - 1} \frac{\nabla_i E_l \left( \tilde{\param}(t) \right)}{R_i^{(l)} \left( \tilde{\param}(t) \right) + \varepsilon}\right|^2} \leq (n - k) \oldbigc{der-p-second-term-aux-one}.
  \end{align*}

  Next, for any $n$ and $j$
  \begin{equation}\label{eq:der-r-bound}
    \begin{aligned}
      &\left| \frac{\mathrm{d}}{\mathrm{d} t} R_j^{(n)} \left( \tilde{\param}(t) \right) \right| = \frac{1}{R_j^{(n)} \left( \tilde{\param}(t) \right)} \left|\sum_{k = 0}^n \rho^{n - k} (1 - \rho) \nabla_j E_k \left( \tilde{\param}(t) \right) \sum_{i = 1}^p \nabla_{i j} E_k \left( \tilde{\param}(t) \right) \dot{\tilde{\paramsc}}_i(t)\right|\\
      &\quad \leq \frac{1}{R_j^{(n)} \left( \tilde{\param}(t) \right)} D_1 p M_1 M_2 \sum_{k = 0}^n \rho^{n - k}(1 - \rho) \leq \oldbigc{der-r}.
    \end{aligned}
  \end{equation}

  This gives
  \begin{align*}
    &\left| \frac{\mathrm{d}}{\mathrm{d} t} \left\{ \frac{\nabla_i E_l \left( \tilde{\param}(t) \right)}{R_i^{(l)} \left( \tilde{\param}(t) \right) + \varepsilon} \right\} \right| \leq \frac{\left| \sum_{s = 1}^p \nabla_{i s} E_l \left( \tilde{\param}(t) \right) \dot{\tilde{\paramsc}}_s(t) \right|}{R_i^{(l)} \left( \tilde{\param}(t) \right) + \varepsilon} + \frac{\left| \nabla_i E_l \left( \tilde{\param}(t) \right)\right| \cdot \left| \frac{\mathrm{d}}{\mathrm{d} t} R_i^{(l)} \left( \tilde{\param}(t) \right) \right|}{\left( R_i^{(l)} \left( \tilde{\param}(t) \right) + \varepsilon \right)^2}\\
    &\quad \leq \frac{D_1 p M_2}{R + \varepsilon} + \frac{M_1}{(R + \varepsilon)^2} \oldbigc{der-r}.
  \end{align*}

  We have obtained
  \begin{equation}\label{eq:der-p-second-term-aux-two-definition}
    \left| \sum_{i = 1}^p \frac{\mathrm{d}}{\mathrm{d} t} \left\{ \nabla_{i j} E_k \left( \tilde{\param}(t) \right) \sum_{l = k}^{n - 1} \frac{\nabla_i E_l \left( \tilde{\param}(t) \right)}{R_i^{(l)} \left( \tilde{\param}(t) \right) + \varepsilon} \right\} \right| \leq (n - k) \oldbigc{der-p-second-term-aux-two}.
  \end{equation}

  This gives a bound on the second term in~\eqref{eq:der-p-into-two-terms}:
  \begin{align*}
    &\left| \sum_{k = 0}^n \rho^{n - k} (1 - \rho) \nabla_j E_k \left( \tilde{\param}(t) \right) \sum_{i = 1}^p \frac{\mathrm{d}}{\mathrm{d} t} \left\{ \nabla_{i j} E_k \left( \tilde{\param}(t) \right) \sum_{l = k}^{n - 1} \frac{\nabla_i E_l \left( \tilde{\param}(t) \right)}{R_i^{(l)} \left( \tilde{\param}(t) \right) + \varepsilon} \right\} \right|\\
    &\quad \leq M_1 \sum_{k = 0}^n \rho^{n - k} (1 - \rho) (n - k) \oldbigc{der-p-second-term-aux-two} \leq \oldbigc{der-p-second-term},
  \end{align*}
  concluding the proof of~\eqref{eq:der-p-bound}.

  We will prove~\eqref{eq:der-p-bar-bound} by bounding the four terms in the expression
\begin{align*}
  &\frac{\mathrm{d}}{\mathrm{d} t} \left\{ \sum_{k = 0}^n \rho^{n - k} (1 - \rho) \nabla_j E_k \left( \tilde{\param}(t) \right) \sum_{i = 1}^p \nabla_{i j} E_k \left( \tilde{\param}(t) \right) \frac{\nabla_i E_n \left( \tilde{\param}(t) \right)}{R_i^{(n)} \left( \tilde{\param}(t) \right) + \varepsilon} \right\}\\
  &\quad = \mathrm{Term1} + \mathrm{Term2} + \mathrm{Term3} + \mathrm{Term4},
\end{align*}
where
\begin{align*}
  &\mathrm{Term1}\\
  &\quad := \sum_{k = 0}^n \rho^{n - k} (1 - \rho) \frac{\mathrm{d}}{\mathrm{d} t} \left\{ \nabla_j E_k \left( \tilde{\param}(t) \right) \right\} \sum_{i = 1}^p \nabla_{i j} E_k \left( \tilde{\param}(t) \right) \frac{\nabla_i E_n \left( \tilde{\param}(t) \right)}{R_i^{(n)} \left( \tilde{\param}(t) \right) + \varepsilon},\\
  &\mathrm{Term2}\\
  &\quad := \sum_{k = 0}^n \rho^{n - k} (1 - \rho) \nabla_j E_k \left( \tilde{\param}(t) \right) \sum_{i = 1}^p \frac{\mathrm{d}}{\mathrm{d} t} \left\{ \nabla_{i j} E_k \left( \tilde{\param}(t) \right) \right\} \frac{\nabla_i E_n \left( \tilde{\param}(t) \right)}{R_i^{(n)} \left( \tilde{\param}(t) \right) + \varepsilon},\\
  &\mathrm{Term3}\\
  &\quad := \sum_{k = 0}^n \rho^{n - k} (1 - \rho) \nabla_j E_k \left( \tilde{\param}(t) \right) \sum_{i = 1}^p \nabla_{i j} E_k \left( \tilde{\param}(t) \right) \frac{\frac{\mathrm{d}}{\mathrm{d} t} \left\{\nabla_i E_n \left( \tilde{\param}(t) \right)\right\}}{R_i^{(n)} \left( \tilde{\param}(t) \right) + \varepsilon},\\
  &\mathrm{Term4}\\
  &\quad := - \sum_{k = 0}^n \rho^{n - k} (1 - \rho) \nabla_j E_k \left( \tilde{\param}(t) \right) \sum_{i = 1}^p \nabla_{i j} E_k \left( \tilde{\param}(t) \right) \frac{\nabla_i E_n \left( \tilde{\param}(t) \right) \frac{\mathrm{d}}{\mathrm{d} t} R_i^{(n)} \left( \tilde{\param}(t) \right)}{\left(R_i^{(n)} \left( \tilde{\param}(t) \right) + \varepsilon\right)^2}.
\end{align*}

To bound Term1, use $\left| \frac{\mathrm{d}}{\mathrm{d} t} \left\{ \nabla_j E_k \left( \tilde{\param}(t) \right) \right\} \right| \leq D_1 p M_2$, giving
\begin{equation*}
\left| \mathrm{Term1} \right| \leq \frac{D_1 p^2 M_1 M_2^2}{R + \varepsilon} \sum_{k = 0}^n \rho^{n - k} (1 - \rho) \leq \frac{D_1 p^2 M_1 M_2^2}{R + \varepsilon}.
\end{equation*}

To bound Term2, use $\left| \frac{\mathrm{d}}{\mathrm{d} t} \left\{ \nabla_{i j} E_k \left( \tilde{\param}(t) \right) \right\} \right| \leq D_1 p M_3$, giving
\begin{equation*}
\left| \mathrm{Term2} \right| \leq \frac{D_1 p^2 M_1^2 M_3}{R + \varepsilon} \sum_{k = 0}^n \rho^{n - k} (1 - \rho) \leq \frac{D_1 p^2 M_1^2 M_3}{R + \varepsilon}.
\end{equation*}

To bound Term3, use $\left| \frac{\mathrm{d}}{\mathrm{d} t} \left\{\nabla_i E_n \left( \tilde{\param}(t) \right)\right\} \right| \leq D_1 p M_2$, giving
\begin{equation*}
  \left| \mathrm{Term3} \right| \leq \frac{D_1 p^2 M_1 M_2^2}{R + \varepsilon} \sum_{k = 0}^n \rho^{n - k} (1 - \rho) \leq \frac{D_1 p^2 M_1 M_2^2}{R + \varepsilon}.
\end{equation*}

To bound Term4, use~\eqref{eq:der-r-bound}, giving
\begin{equation*}
\left| \mathrm{Term4} \right| \leq \frac{p M_1^2 M_2 \oldbigc{der-r}}{(R + \varepsilon)^2} \sum_{k = 0}^n \rho^{n - k} (1 - \rho) \leq \frac{p M_1^2 M_2 \oldbigc{der-r}}{(R + \varepsilon)^2}.
\end{equation*}
The proof of~\eqref{eq:der-p-bar-bound} is finished.

The inequality \eqref{eq:nabla-two-nabla-one-over-r-eps} is already proven in~\eqref{eq:der-p-second-term-aux-two-definition}.

To prove~\eqref{eq:der-huge-func-first-term-bound}, note that the bound \eqref{eq:der-r-bound} gives
\begin{align}
  &\left| \frac{\mathrm{d}}{\mathrm{d} t} \left\{ \frac{1}{R_j^{(n)} \left( \tilde{\param}(t) \right)} \right\} \right| = \frac{\left| \frac{\mathrm{d}}{\mathrm{d} t} R_j^{(n)} \left( \tilde{\param}(t) \right) \right|}{R_j^{(n)} \left( \tilde{\param}(t) \right)^2} \leq \frac{\oldbigc{der-r}}{R^2},\label{eq:der-r-inverse-bound}\\
  &\left| \frac{\mathrm{d}}{\mathrm{d} t} \left\{ \frac{1}{R_j^{(n)} \left( \tilde{\param}(t)\right) + \varepsilon} \right\} \right| = \frac{\left| \frac{\mathrm{d}}{\mathrm{d} t} R_j^{(n)} \left( \tilde{\param}(t) \right) \right|}{\left(R_j^{(n)} \left( \tilde{\param}(t) \right) + \varepsilon\right)^2} \leq \frac{\oldbigc{der-r}}{(R + \varepsilon)^2},\\
  &\left| \frac{\mathrm{d}}{\mathrm{d} t} \left\{ \frac{1}{\left(R_j^{(n)} \left( \tilde{\param}(t)\right) + \varepsilon\right)^2} \right\} \right| = \frac{2 \left| \frac{\mathrm{d}}{\mathrm{d} t} R_j^{(n)} \left( \tilde{\param}(t) \right) \right|}{\left(R_j^{(n)} \left( \tilde{\param}(t) \right) + \varepsilon\right)^3} \leq \frac{2 \oldbigc{der-r}}{(R + \varepsilon)^3}.
\end{align}

Combining two bounds above, we have
\begin{align*}
  &\left| \frac{\mathrm{d}}{\mathrm{d} t} \left\{ \left( R_j^{(n)} \left( \tilde{\param}(t) \right) + \varepsilon \right)^{-2} R_j^{(n)}(\tilde{\param}(t))^{-1} \right\} \right|\\
  &\quad \leq \frac{\left| \frac{\mathrm{d}}{\mathrm{d} t} \left\{ \left( R_j^{(n)} \left( \tilde{\param}(t) \right) + \varepsilon \right)^{-2}\right\} \right|}{R_j^{(n)}(\tilde{\param}(t))} + \frac{\left|\frac{\mathrm{d}}{\mathrm{d} t} \left\{ R_j^{(n)}(\tilde{\param}(t))^{-1} \right\}\right|}{\left( R_j^{(n)} \left( \tilde{\param}(t) \right) + \varepsilon \right)^{2}} \leq \oldbigc{der-r-pl-eps-sq-r-inverse}.
\end{align*}

We are ready to conclude
\begin{align*}
  &\left|\timeder{\frac{\nabla_j E_n \left( \tilde{\param}(t) \right) \left(2 P_j^{(n)}\left(\tilde{\param}(t)\right) + \bar{P}_j^{(n)}\left(\tilde{\param}(t)\right)\right)}{2 \left( R_j^{(n)} \left( \tilde{\param}(t) \right) + \varepsilon \right)^2 R_j^{(n)} \left( \tilde{\param}(t) \right)}}\right|\\
  &\quad \leq \left| \frac{\timeder{\nabla_j E_n \left( \tilde{\param}(t) \right)} \left(2 P_j^{(n)}\left(\tilde{\param}(t)\right) + \bar{P}_j^{(n)}\left(\tilde{\param}(t)\right)\right)}{2 \left( R_j^{(n)} \left( \tilde{\param}(t) \right) + \varepsilon \right)^2 R_j^{(n)} \left( \tilde{\param}(t) \right)}\right| +\\
  &\qquad + \left|\frac{\nabla_j E_n \left( \tilde{\param}(t) \right) \timeder{2 P_j^{(n)}\left(\tilde{\param}(t)\right) + \bar{P}_j^{(n)}\left(\tilde{\param}(t)\right)}}{2 \left( R_j^{(n)} \left( \tilde{\param}(t) \right) + \varepsilon \right)^2 R_j^{(n)} \left( \tilde{\param}(t) \right)}\right|\\
  &\qquad + \left|\frac{\nabla_j E_n \left( \tilde{\param}(t) \right) \left(2 P_j^{(n)}\left(\tilde{\param}(t)\right) + \bar{P}_j^{(n)}\left(\tilde{\param}(t)\right)\right)}{2}\right.\\
  &\qquad \times \left.\timeder{\left( R_j^{(n)} \left( \tilde{\param}(t) \right) + \varepsilon \right)^{-2} R_j^{(n)}(\tilde{\param}(t))^{-1}}\right| \leq \oldbigc{der-huge-func-first-term}.
\end{align*}

It is left to prove~\eqref{eq:der-huge-func-second-term-bound}. Since
\begin{equation*}
\left| \sum_{i = 1}^p \nabla_{i j} E_n \left( \tilde{\param}(t) \right) \frac{\nabla_i E_n \left( \tilde{\param}(t) \right)}{R_i^{(n)} \left( \tilde{\param}(t) \right) + \varepsilon} \right| \leq \frac{p M_1 M_2}{R + \varepsilon}
\end{equation*}
and, as we have already seen in the argument for~\eqref{eq:der-p-bar-bound},
\begin{equation*}
\left| \timeder{\sum_{i = 1}^p \nabla_{i j} E_n \left( \tilde{\param}(t) \right) \frac{\nabla_i E_n \left( \tilde{\param}(t) \right)}{R_i^{(n)} \left( \tilde{\param}(t) \right) + \varepsilon}} \right| \leq \frac{p^2 D_1 M_1 M_3}{R + \varepsilon} + \frac{p^2 D_1 M_2^2}{R + \varepsilon} + \frac{p M_1 M_2 \oldbigc{der-r}}{\left( R + \varepsilon \right)^2},
\end{equation*}
we are ready to bound
\begin{equation*}
  \left| \timeder{\frac{\sum_{i = 1}^p \nabla_{i j} E_n \left( \tilde{\param}(t) \right) \frac{\nabla_i E_n \left( \tilde{\param}(t) \right)}{R_i^{(n)}\left(\tilde{\param}(t)\right) + \varepsilon}}{2 \left( R_j^{(n)}(\tilde{\param}(t)) + \varepsilon \right)}} \right| \leq \oldbigc{der-huge-func-second-term}.
\end{equation*}

The proof of \cref{lem:first-derivatives} is concluded.
\end{proof}

\begin{lemma}
Suppose \cref{ass:bounds} holds. Then the second derivative of $t \mapsto \tilde{\paramsc}_j(t)$ is uniformly over $j$ and $t \in [0, T]$ bounded in absolute value by some positive constant, say $D_2$.
\end{lemma}

\begin{proof}
This follows from the definition of $\tilde{\param}(t)$ given in~\eqref{eq:nth-step-modified-equation}, $h \leq T$ and that the first derivatives of all three terms in~\eqref{eq:nth-step-modified-equation} are bounded by \cref{lem:first-derivatives}.
\end{proof}

Finally, we need to argue that the \textit{third} derivative of $t \mapsto \tilde{\paramsc}_j(t)$ is bounded in absolute value. To achieve this, we need to bound the second derivatives of the terms on the right-hand side of~\eqref{eq:nth-step-modified-equation}.

\begin{lemma}\label{lem:second-derivatives}
  Suppose \cref{ass:bounds} holds. Then for all $n, k \in \left\{ 0, 1, \dots, \lfloor T / h \rfloor \right\}$, $j \in \left\{ 1, \ldots, p \right\}$
  \begin{align}
    &\sup_{t \in [0, T]} \left| \timederder{\nabla_j E_n \left( \tilde{\param}(t) \right)} \right| \leq \oldbigc{e-second-der},\label{eq:e-second-der-bound}\\
    &\sup_{t \in [0, T]} \left| \timederder{R_j^{(n)} \left( \tilde{\param}(t) \right)}\right| \leq \oldbigc{der-der-r},\label{eq:der-der-r}\\
    &\sup_{t \in [0, T]} \left| \timederder{\left(R_j^{(n)} \left( \tilde{\param}(t) \right) + \varepsilon\right)^{-2}} \right| \leq \oldbigc{der-der-r-inv-sq},\label{eq:der-der-r-inv-sq}\\
    &\sup_{t \in [0, T]} \left| \timederder{R_j^{(n)} \left( \tilde{\param}(t) \right)^{-1}} \right| \leq \oldbigc{der-der-r-inv},\label{eq:der-der-r-inv}\\
    &\sup_{t \in [0, T]} \left| \timederder{\left(R_j^{(n)} \left( \tilde{\param}(t) \right) + \varepsilon\right)^{-2} R_j^{(n)} \left( \tilde{\param}(t) \right)^{-1}} \right| \leq \oldbigc{der-der-r-sq-r-inv},\label{eq:der-der-r-sq-r-inv}\\
    &\sup_{t \in [0, T]} \left| \timederder{\sum_{i = 1}^p \nabla_{i j} E_k \left( \tilde{\param}(t) \right) \sum_{l = k}^{n - 1} \frac{\nabla_i E_l \left( \tilde{\param}(t) \right)}{R_i^{(l)} \left( \tilde{\param}(t) \right) + \varepsilon}} \right| \leq (n - k) \oldbigc{der-der-second-first-grad-over-r-partial-sum-bound}\quad\text{for $k < n$},\label{eq:der-der-second-first-grad-over-r-partial-sum-bound}
  \end{align}
  with constants $\newbigc[e-second-der]$, $\newbigc[der-der-r]$, $\newbigc[der-der-r-inv-sq]$, $\newbigc[der-der-r-inv]$, $\newbigc[der-der-r-sq-r-inv]$, $\newbigc[der-der-second-first-grad-over-r-partial-sum-bound]$ defined as follows:
  \begin{align*}
    &\oldbigc{e-second-der} := p^2 M_3 D_1^2 + p M_2 D_2,\\
    &\oldbigc{der-der-r} := \frac{\oldbigc{der-r}}{R^2} p M_1 M_2 D_1 + \frac{1}{R} p^2 M_2^2 D_1^2 + \frac{1}{R} p^2 M_1 M_3 D_1^2 + \frac{1}{R} p M_1 M_2 D_2,\\
    &\oldbigc{der-der-r-inv-sq} := \frac{6 \oldbigc{der-r}^2}{(R + \varepsilon)^4} + \frac{2 \oldbigc{der-der-r}}{(R + \varepsilon)^3},\\
    &\oldbigc{der-der-r-inv} := \frac{2 \oldbigc{der-r}^2}{R^3} + \frac{\oldbigc{der-der-r}}{R^2},\\
    &\oldbigc{der-der-r-sq-r-inv} := \frac{\oldbigc{der-der-r-inv-sq}}{R} + \frac{4 \oldbigc{der-r}^2}{R^2 (R + \varepsilon)^3} + \frac{\oldbigc{der-der-r-inv}}{(R + \varepsilon)^2},\\
    &\oldbigc{der-der-second-first-grad-over-r-partial-sum-bound} := p \left[ \frac{2 \oldbigc{der-r} \left(D_1 M_2^2 p+D_1 M_1 M_3 p\right)}{(R+\varepsilon )^2}+M_1 M_2
      \left(\frac{2 \oldbigc{der-r}^2}{(R+\varepsilon )^3}+\frac{\oldbigc{der-der-r}}{(R+\varepsilon )^2}\right)\right.\\
    &\qquad +\left.\frac{2
      D_1^2 M_2 M_3 p^2+M_2 \left(D_1^2 M_3 p^2+D_2 M_2 p\right)+M_1 \left(D_1^2 M_4
      p^2+D_2 M_3 p\right)}{R+\varepsilon} \right].
  \end{align*}
\end{lemma}

\begin{proof}[Proof of \cref{lem:second-derivatives}] We prove the inequalities one by one.

The proof of~\eqref{eq:e-second-der-bound} is straightforward:
\begin{equation*}
  \left| \timederder{\nabla_j E_n \left( \tilde{\param}(t) \right)} \right| = \left| \sum_{i = 1}^p \sum_{s = 1}^p \nabla_{i j s} E_n \left( \tilde{\param}(t) \right) \dot{\tilde{\paramsc}}_s(t) \dot{\tilde{\paramsc}}_i(t) + \sum_{i = 1}^p \nabla_{i j} E_n \left( \tilde{\param}(t) \right) \ddot{\tilde{\paramsc}}_t(t) \right| \leq \oldbigc{e-second-der}.
\end{equation*}

To prove~\eqref{eq:der-der-r}, note that
\begin{align*}
  &\timederder{R_j^{(n)} \left( \tilde{\param}(t) \right)} = \timeder{R_j^{(n)} \left( \tilde{\param}(t) \right)^{-1}} \sum_{k = 0}^n \rho^{n - k} (1 - \rho) \nabla_j E_k \left( \tilde{\param}(t) \right) \sum_{i = 1}^p \nabla_{i j} E_k \left( \tilde{\param}(t) \right) \dot{\tilde{\paramsc}}_i(t)\\
  &\qquad + R_j^{(n)} \left( \tilde{\param}(t) \right)^{-1} \sum_{k = 0}^n \rho^{n - k} (1 - \rho) \timeder{\nabla_j E_k \left( \tilde{\param}(t) \right)} \sum_{i = 1}^p \nabla_{i j} E_k \left( \tilde{\param}(t) \right) \dot{\tilde{\paramsc}}_i(t)\\
  &\qquad + R_j^{(n)} \left( \tilde{\param}(t) \right)^{-1} \sum_{k = 0}^n \rho^{n - k} (1 - \rho) \nabla_j E_k \left( \tilde{\param}(t) \right) \sum_{i = 1}^p \timeder{\nabla_{i j} E_k \left( \tilde{\param}(t) \right)} \dot{\tilde{\paramsc}}_i(t)\\
  &\qquad + R_j^{(n)} \left( \tilde{\param}(t) \right)^{-1} \sum_{k = 0}^n \rho^{n - k} (1 - \rho) \nabla_j E_k \left( \tilde{\param}(t) \right) \sum_{i = 1}^p \nabla_{i j} E_k \left( \tilde{\param}(t) \right) \ddot{\tilde{\paramsc}}_i(t),
\end{align*}
giving by~\eqref{eq:der-r-inverse-bound}
\begin{align*}
  &\left| \timederder{R_j^{(n)} \left( \tilde{\param}(t) \right)}\right| \leq \frac{\oldbigc{der-r}}{R^2} p M_1 M_2 D_1 \sum_{k = 0}^n \rho^{n - k} (1 - \rho) + \frac{1}{R} p^2 M_2^2 D_1^2 \sum_{k = 0}^n \rho^{n - k} (1 - \rho)\\
  &\qquad + \frac{1}{R} p^2 M_1 M_3 D_1^2 \sum_{k = 0}^n \rho^{n - k} (1 - \rho) + \frac{1}{R} p M_1 M_2 D_2 \sum_{k = 0} \rho^{n - k} (1 - \rho)\\
  &\quad \leq \oldbigc{der-der-r}.
\end{align*}

To prove~\eqref{eq:der-der-r-inv-sq}, note that
\begin{equation*}
\timederder{\left(R_j^{(n)} \left( \tilde{\param}(t) \right) + \varepsilon\right)^{-2}} = \frac{6 \left(\timeder{R_j^{(n)} \left( \tilde{\param}(t) \right)}\right)^2}{\left( R_j^{(n)} \left( \tilde{\param}(t) \right) + \varepsilon \right)^4} - \frac{2 \timederder{R_j^{(n)} \left( \tilde{\param}(t) \right)}}{\left( R_j^{(n)} \left( \tilde{\param}(t) \right) + \varepsilon \right)^3},
\end{equation*}
giving by~\eqref{eq:der-r-bound} and~\eqref{eq:der-der-r}
\begin{equation*}
\left| \timederder{\left(R_j^{(n)} \left( \tilde{\param}(t) \right) + \varepsilon\right)^{-2}} \right| \leq \oldbigc{der-der-r-inv-sq}.
\end{equation*}

The bound~\eqref{eq:der-der-r-inv} follows from~~\eqref{eq:der-r-bound}, \eqref{eq:der-der-r} and
\begin{equation*}
\timederder{R_j^{(n)} \left( \tilde{\param}(t) \right)^{-1}} = \frac{2 \left(\timeder{R_j^{(n)} \left( \tilde{\param}(t) \right)}\right)^2}{R_j^{(n)} \left( \tilde{\param}(t) \right)^3} - \frac{\timederder{R_j^{(n)} \left( \tilde{\param}(t) \right)}}{R_j^{(n)} \left( \tilde{\param}(t) \right)^2}.
\end{equation*}

To justify~\eqref{eq:der-der-r-sq-r-inv}, put temporarily $a := \left(R_j^{(n)} \left( \tilde{\param}(t) \right) + \varepsilon\right)^{-2}$, $b := R_j^{(n)} \left( \tilde{\param}(t) \right)^{-1}$ and use
  \begin{align*}
    &\left| a \right| \leq \frac{1}{(R + \varepsilon)^2},\quad \left| b \right| \leq \frac{1}{R},\\
    &\left| \dot{a} \right| \leq \frac{2 \oldbigc{der-r}}{(R + \varepsilon)^3},\quad \left| \dot{b} \right| \leq \frac{\oldbigc{der-r}}{R^2},\\
&\left| \ddot{a} \right| \leq \oldbigc{der-der-r-inv-sq},\quad \left| \ddot{b} \right| \leq \oldbigc{der-der-r-inv}
  \end{align*}
combined with
\begin{equation*}
\timederder{a b} = \ddot{a} b + 2 \dot{a} \dot{b} + a \ddot{b}.
\end{equation*}

To justify~\eqref{eq:der-der-second-first-grad-over-r-partial-sum-bound}, put temporarily
\begin{align*}
  &a := \nabla_{i j} E_k \left( \tilde{\param}(t) \right),\\
  &b := \nabla_i E_l \left( \tilde{\param}(t) \right),\\
  &c := \left( R_i^{(l)} \left( \tilde{\param}(t) \right) + \varepsilon \right)^{-1},
\end{align*}
and use
\begin{align*}
  &\left| a \right| \leq M_2,\quad \left| \dot{a} \right| \leq p M_3 D_1,\quad \left| \ddot{a} \right| \leq p^2 M_4 D_1^2 + p M_3 D_2,\\
  &\left| b \right| \leq M_1,\quad \left| \dot{b} \right| \leq p M_2 D_1,\quad \left| \ddot{b} \right| \leq p^2 M_3 D_1^2 + p M_2 D_2,\\
  &\left| c \right| \leq \frac{1}{R + \varepsilon},\quad \left| \dot{c} \right| \leq \frac{\oldbigc{der-r}}{(R + \varepsilon)^2},\quad \left| \ddot{c} \right| \leq \frac{2 \oldbigc{der-r}^2}{(R + \varepsilon)^3} + \frac{\oldbigc{der-der-r}}{(R + \varepsilon)^2},
\end{align*}
from which \eqref{eq:der-der-second-first-grad-over-r-partial-sum-bound} follows.

The proof of \cref{lem:second-derivatives} is concluded.
\end{proof}

\begin{lemma}
Suppose \cref{ass:bounds} holds. Then the third derivative of $t \mapsto \tilde{\paramsc}_j(t)$ is uniformly over $j$ and $t \in [0, T]$ bounded in absolute value by some positive constant, say $D_3$.
\end{lemma}

\begin{proof}
By~\eqref{eq:second-first-grad-over-r-partial-sum-bound}, \eqref{eq:der-p-second-term-aux-two-definition} and \eqref{eq:der-der-second-first-grad-over-r-partial-sum-bound}
\begin{align*}
  &\left| \sum_{i = 1}^p \nabla_{i j} E_k \left( \tilde{\param}(t) \right) \sum_{l = k}^{n - 1} \frac{\nabla_i E_l \left( \tilde{\param}(t) \right)}{R_i^{(l)} \left( \tilde{\param}(t) \right) + \varepsilon} \right| \leq (n - k) \oldbigc{second-first-grad-over-r-partial-sum},\\
  &\left| \timeder{\sum_{i = 1}^p \nabla_{i j} E_k \left( \tilde{\param}(t) \right) \sum_{l = k}^{n - 1} \frac{\nabla_i E_l \left( \tilde{\param}(t) \right)}{R_i^{(l)} \left( \tilde{\param}(t) \right) + \varepsilon}} \right| \leq (n - k) \oldbigc{der-p-second-term-aux-two},\\
  &\left| \timederder{\sum_{i = 1}^p \nabla_{i j} E_k \left( \tilde{\param}(t) \right) \sum_{l = k}^{n - 1} \frac{\nabla_i E_l \left( \tilde{\param}(t) \right)}{R_i^{(l)} \left( \tilde{\param}(t) \right) + \varepsilon}} \right| \leq (n - k) \oldbigc{der-der-second-first-grad-over-r-partial-sum-bound}.
\end{align*}

From the definition of $t \mapsto P_j^{(n)} \left( \tilde{\param}(t) \right)$, it means that its derivatives up to order two are bounded. Similarly, the same is true for $t \mapsto \bar{P}_j^{(n)} \left( \tilde{\param}(t) \right)$.

It follows from~\eqref{eq:der-der-r-inv-sq} and its proof that the derivatives up to order two of
\begin{equation*}
t \mapsto \left(R_j^{(n)} \left( \tilde{\param}(t) \right) + \varepsilon\right)^{-2} R_j^{(n)} \left( \tilde{\param}(t) \right)^{-1}
\end{equation*}
are also bounded.

These considerations give the boundedness of the second derivative of the term
\begin{equation*}
t \mapsto \frac{\nabla_j E_n \left( \tilde{\param}(t) \right) \left(2 P_j^{(n)}\left(\tilde{\param}(t)\right) + \bar{P}_j^{(n)}\left(\tilde{\param}(t)\right)\right)}{2 \left( R_j^{(n)} \left( \tilde{\param}(t) \right) + \varepsilon \right)^2 R_j^{(n)} \left( \tilde{\param}(t) \right)}
\end{equation*}
in~\eqref{eq:nth-step-modified-equation}. The boundedness of the second derivatives of the other two terms is shown analogously. By~\eqref{eq:nth-step-modified-equation} and since $h \leq T$, this means
\begin{equation*}
\sup_j \sup_{t \in [0, T]} \left| \dddot{\tilde{\paramsc}}_j(t) \right| \leq D_3
\end{equation*}
for some positive constant $D_3$.
\end{proof}

\section{Proof of Theorem~\ref{th:local-error-bound}}\label{sec:proof-of-local-error-bound}

Our next objective is proving and identifying the constant in the equality
\begin{align*}
&\frac{1}{\sqrt{\sum_{k = 0}^n \rho^{n - k} (1 - \rho) \left( \nabla_j E_k \left( \tilde{\param}(t_k) \right) \right)^2} + \varepsilon}\\
&\quad = \frac{1}{R^{(n)}_j \left( \tilde{\param}(t_n) \right) + \varepsilon} - h \frac{P^{(n)}_j \left( \tilde{\param}(t_n) \right)}{\left( R^{(n)}_j \left( \tilde{\param}(t_n) \right) + \varepsilon \right)^2 R^{(n)}_j \left( \tilde{\param}(t_n) \right)} + O(h^2).
\end{align*}
We will make some preparations and achieve this objective in \cref{lem:roots-in-denominators-first-order}. Then we will conclude the proof of \cref{th:local-error-bound}.

\begin{lemma}\label{lem:grad-e-to-the-nth-point-n-k}
  Suppose \cref{ass:bounds} holds. Then for all $n \in \left\{ 0, 1, \dots, \lfloor T / h \rfloor \right\}$, $k \in \left\{ 0, 1, \ldots, n - 1 \right\}$, $j \in \left\{ 1, \ldots, p \right\}$ we have
  \begin{equation}\label{eq:grad-e-k-k-grad-e-k-n-bound}
    \left| \nabla_j E_k \left( \tilde{\param}(t_k) \right) - \nabla_j E_k \left( \tilde{\param}(t_n) \right) \right| \leq \oldbigc{e-first-der} (n - k) h
  \end{equation}
\end{lemma}

\begin{proof}
  \eqref{eq:grad-e-k-k-grad-e-k-n-bound} follows from the mean value theorem applied $n - k$ times.
\end{proof}

\begin{lemma}\label{lem:c1-c2-n-l-1-bound-for-e}
  In the setting of \cref{lem:grad-e-to-the-nth-point-n-k}, for any $l \in \left\{ k, k + 1, \ldots, n - 1 \right\}$ we have
  \begin{align*}
  &\left| \nabla_j E_k \left( \tilde{\param}(t_l) \right) - \nabla_j E_k \left( \tilde{\param}(t_{l + 1}) \right) - h \sum_{i = 1}^p \nabla_{i j} E_k \left( \tilde{\param}(t_n) \right) \frac{\nabla_i E_l\left(\tilde{\param}(t_n)\right)}{R^{(l)}_i\left(\tilde{\param}(t_n)\right) + \varepsilon} \right|\\
  &\quad \leq \left(\oldbigc{e-second-der} / 2 + \oldbigc{sol-der-first-order} + (n - l - 1) \oldbigc{der-p-second-term-aux-two}\right) h^2.
\end{align*}
\end{lemma}

\begin{proof}
  By the Taylor expansion of $t \mapsto \nabla_j E_k \left( \tilde{\param}(t) \right)$ on the segment $[t_l, t_{l + 1}]$ at $t_{l + 1}$ on the left
\begin{equation*}
  \left| \nabla_j E_k \left( \tilde{\param}(t_l) \right) - \nabla_j E_k \left( \tilde{\param}(t_{l + 1}) \right) + h \sum_{i = 1}^p \nabla_{i j} E_k \left( \tilde{\param}(t_{l + 1}) \right) \dot{\tilde{\param}}_i\left(t_{l + 1}^-\right) \right| \leq \frac{\oldbigc{e-second-der}}{2} h^2.
\end{equation*}

Combining this with~\eqref{eq:sol-der-first-order} gives
\begin{equation}\label{eq:c1-c2-bound}
\begin{aligned}
  &\left| \nabla_j E_k \left( \tilde{\param}(t_l) \right) - \nabla_j E_k \left( \tilde{\param}(t_{l + 1}) \right) - h \sum_{i = 1}^p \nabla_{i j} E_k \left( \tilde{\param}(t_{l + 1}) \right) \frac{\nabla_i E_l\left(\tilde{\param}(t_{l + 1})\right)}{R^{(l)}_i\left(\tilde{\param}(t_{l + 1})\right) + \varepsilon} \right|\\
  &\quad \leq \left(\oldbigc{e-second-der} / 2 + \oldbigc{sol-der-first-order}\right) h^2.
\end{aligned}
\end{equation}

Now applying the mean-value theorem $n - l - 1$ times, we have by~\eqref{eq:der-p-second-term-aux-two-definition}
\begin{align*}
  &\left| \sum_{i = 1}^p \nabla_{i j} E_k \left( \tilde{\param}(t_{l + 1}) \right) \frac{\nabla_i E_l\left(\tilde{\param}(t_{l + 1})\right)}{R^{(l)}_i\left(\tilde{\param}(t_{l + 1})\right) + \varepsilon} - \sum_{i = 1}^p \nabla_{i j} E_k \left( \tilde{\param}(t_{l + 2}) \right) \frac{\nabla_i E_l\left(\tilde{\param}(t_{l + 2})\right)}{R^{(l)}_i\left(\tilde{\param}(t_{l + 2})\right) + \varepsilon}\right| \leq \oldbigc{der-p-second-term-aux-two} h,\\
  &\cdots\\
  &\left| \sum_{i = 1}^p \nabla_{i j} E_l \left( \tilde{\param}(t_{n - 1}) \right) \frac{\nabla_i E_k\left(\tilde{\param}(t_{n - 1})\right)}{R^{(l)}_i\left(\tilde{\param}(t_{n - 1})\right) + \varepsilon} - \sum_{i = 1}^p \nabla_{i j} E_k \left( \tilde{\param}(t_{n}) \right) \frac{\nabla_i E_l\left(\tilde{\param}(t_{n})\right)}{R^{(l)}_i\left(\tilde{\param}(t_{n})\right) + \varepsilon}\right| \leq \oldbigc{der-p-second-term-aux-two} h,
\end{align*}
and in particular
\begin{align*}
  &\left| \sum_{i = 1}^p \nabla_{i j} E_k \left( \tilde{\param}(t_{l + 1}) \right) \frac{\nabla_i E_l\left(\tilde{\param}(t_{l + 1})\right)}{R^{(l)}_i\left(\tilde{\param}(t_{l + 1})\right) + \varepsilon} - \sum_{i = 1}^p \nabla_{i j} E_k \left( \tilde{\param}(t_{n}) \right) \frac{\nabla_i E_l\left(\tilde{\param}(t_{n})\right)}{R^{(l)}_i\left(\tilde{\param}(t_{n})\right) + \varepsilon}\right|\\
  &\quad \leq (n - l - 1) \oldbigc{der-p-second-term-aux-two} h.
\end{align*}

Combining this with~\eqref{eq:c1-c2-bound}, we conclude the proof of \cref{lem:c1-c2-n-l-1-bound-for-e}.
\end{proof}

\begin{lemma}\label{lem:grad-e-to-the-nth-point}
In the setting of \cref{lem:grad-e-to-the-nth-point-n-k},
  \begin{align*}
    &\left| \nabla_j E_k \left( \tilde{\param}(t_k) \right) - \nabla_j E_k \left( \tilde{\param}(t_n) \right) - h \sum_{i = 1}^p \nabla_{i j} E_k \left( \tilde{\param}(t_n) \right) \sum_{l = k}^{n - 1} \frac{\nabla_i E_l \left( \tilde{\param}(t_n) \right)}{R_i^{(l)} \left( \tilde{\param}(t_n) \right) + \varepsilon} \right|\\
    &\quad \leq \left( (n - k) (\oldbigc{e-second-der} / 2 + \oldbigc{sol-der-first-order}) + \frac{(n - k)(n - k - 1)}{2} \oldbigc{der-p-second-term-aux-two} \right) h^2.
  \end{align*}
\end{lemma}

\begin{proof}
  Fix $n \in \mathbb{Z}_{\geq 0}$.

Note that
\begin{align*}
  &\left| \nabla_j E_k \left( \tilde{\param}(t_k) \right) - \nabla_j E_k \left( \tilde{\param}(t_n) \right) - h \sum_{i = 1}^p \nabla_{i j} E_k \left( \tilde{\param}(t_n) \right) \sum_{l = k}^{n - 1} \frac{\nabla_i E_l \left( \tilde{\param}(t_n) \right)}{R_i^{(l)} \left( \tilde{\param}(t_n) \right) + \varepsilon} \right|\\
  &\quad = \left| \sum_{l = k}^{n - 1} \left\{ \nabla_j E_k \left( \tilde{\param}(t_l) \right) - \nabla_j E_k \left( \tilde{\param}(t_{l + 1}) \right) - h \sum_{i = 1}^p \nabla_{i j} E_k \left( \tilde{\param}(t_n) \right) \frac{\nabla_i E_l\left(\tilde{\param}(t_n)\right)}{R^{(l)}_i\left(\tilde{\param}(t_n)\right) + \varepsilon} \right\} \right|\\
  &\quad \leq \sum_{l = k}^{n - 1} \left| \nabla_j E_k \left( \tilde{\param}(t_l) \right) - \nabla_j E_k \left( \tilde{\param}(t_{l + 1}) \right) - h \sum_{i = 1}^p \nabla_{i j} E_k \left( \tilde{\param}(t_n) \right) \frac{\nabla_i E_l\left(\tilde{\param}(t_n)\right)}{R^{(l)}_i\left(\tilde{\param}(t_n)\right) + \varepsilon} \right|\\
  &\quad \labelrel\leq{eq:e-k-e-n-h-term-approx-one} \sum_{l = k}^{n - 1} \left(\oldbigc{e-second-der} / 2 + \oldbigc{sol-der-first-order} + (n - l - 1) \oldbigc{der-p-second-term-aux-two}\right) h^2 = \left( (n - k) (\oldbigc{e-second-der} / 2 + \oldbigc{sol-der-first-order}) + \frac{(n - k)(n - k - 1)}{2} \oldbigc{der-p-second-term-aux-two} \right) h^2,
\end{align*}
where~\eqref{eq:e-k-e-n-h-term-approx-one} is by \cref{lem:c1-c2-n-l-1-bound-for-e}.
\end{proof}

\begin{lemma}\label{cor:grad-e-squared-to-the-nth-point}
  Suppose \cref{ass:bounds} holds. Then for all $n \in \left\{ 0, 1, \dots, \lfloor T / h \rfloor \right\}$, $j \in \left\{ 1, \ldots, p \right\}$
\begin{equation}\label{eq:diff-of-sums-of-squared-norms-zeroth-order}
\left| \sum_{k = 0}^n \rho^{n - k} (1 - \rho) \left( \nabla_j E_k \left( \tilde{\param}(t_k) \right) \right)^2 - R^{(n)}_j \left( \tilde{\param}(t_n) \right)^2 \right| \leq \oldbigc{firstorderdif} h
\end{equation}
and
\begin{equation}\label{eq:diff-of-sums-of-squared-norms}
  \left| \sum_{k = 0}^n \rho^{n - k} (1 - \rho) \left( \nabla_j E_k \left( \tilde{\param}(t_k) \right) \right)^2 - R^{(n)}_j \left( \tilde{\param}(t_n) \right)^2 - 2 h P^{(n)}_j \left( \tilde{\param}(t_n) \right) \right| \leq \oldbigc{secondorderdif} h^2
\end{equation}
with $\newbigc[firstorderdif]$ and $\newbigc[secondorderdif]$ defined as follows:
\begin{align*}
  &\oldbigc{firstorderdif}(\rho) := 2 M_1 \oldbigc{e-first-der} \frac{\rho}{1 - \rho},\\
  &\oldbigc{secondorderdif}(\rho) := M_1 \left| \oldbigc{e-second-der} + 2 \oldbigc{sol-der-first-order} - \oldbigc{der-p-second-term-aux-two} \right| \frac{\rho}{1 - \rho}\\
  &\qquad + \left( M_1 \oldbigc{der-p-second-term-aux-two} + \left| \oldbigc{e-second-der} + 2 \oldbigc{sol-der-first-order} - \oldbigc{der-p-second-term-aux-two} \right| \oldbigc{second-first-grad-over-r-partial-sum} + \frac{\left( \oldbigc{e-second-der} + 2 \oldbigc{sol-der-first-order} - \oldbigc{der-p-second-term-aux-two} \right)^2}{4} \right) \frac{\rho (1 + \rho)}{(1 - \rho)^2}\\
  &\qquad + \left(\oldbigc{der-p-second-term-aux-two} \oldbigc{second-first-grad-over-r-partial-sum} + \frac{\oldbigc{der-p-second-term-aux-two}}{2} \left| \oldbigc{e-second-der} + 2 \oldbigc{sol-der-first-order} - \oldbigc{der-p-second-term-aux-two} \right|\right) \frac{\rho \left( 1 + 4 \rho + \rho^2 \right)}{(1 - \rho)^3} + \frac{\oldbigc{der-p-second-term-aux-two}^2}{4} \cdot \frac{\rho \left( 1 + 11 \rho + 11 \rho^2 + \rho^3 \right)}{(1 - \rho)^4}.
\end{align*}
\end{lemma}

\begin{proof}
  Note that
\begin{align*}
  &\left| \left(\nabla_j E_k \left( \tilde{\param}(t_k) \right)\right)^2 - \left(\nabla_j E_k \left( \tilde{\param}(t_n) \right)\right)^2 \right|\\
  &\quad \leq \left| \nabla_j E_k \left( \tilde{\param}(t_k) \right) - \nabla_j E_k \left( \tilde{\param}(t_n) \right) \right| \cdot \left| \nabla_j E_k \left( \tilde{\param}(t_k) \right) + \nabla_j E_k \left( \tilde{\param}(t_n) \right) \right|\\
  &\quad \labelrel\leq{eq:c-0-n-k-two-m} \oldbigc{e-first-der} (n - k) h \cdot 2 M_1,
\end{align*}
where~\eqref{eq:c-0-n-k-two-m} is by~\eqref{eq:grad-e-k-k-grad-e-k-n-bound}. Using the triangle inequality, we can conclude
\begin{align*}
  &\left| \sum_{k = 0}^n \rho^{n - k} (1 - \rho) \left( \nabla_j E_k \left( \tilde{\param}(t_k) \right) \right)^2 - R^{(n)}_j \left( \tilde{\param}(t_n) \right)^2 \right|\\
  &\quad \leq 2 M_1 \oldbigc{e-first-der} h (1 - \rho) \sum_{k = 0}^n (n - k) \rho^{n - k} = 2 M_1 \oldbigc{e-first-der} h (1 - \rho) \sum_{k = 0}^n k \rho^k = 2 M_1 \oldbigc{e-first-der} \frac{\rho}{1 - \rho} h.
\end{align*}
\eqref{eq:diff-of-sums-of-squared-norms-zeroth-order} is proven.

  We continue by showing
  \begin{equation}\label{eq:diff-of-squared-norms}
    \begin{aligned}
      &\left| \left(\nabla_j E_k \left( \tilde{\param}(t_k) \right)\right)^2 - \left( \nabla_j E_k \left( \tilde{\param}(t_n) \right) \right)^2\right.\\
      &\qquad \left. - 2 \nabla_j E_k \left( \tilde{\param}(t_n) \right) h \sum_{i = 1}^p \nabla_{i j} E_k \left( \tilde{\param}(t_n) \right) \sum_{l = k}^{n - 1} \frac{\nabla_i E_l \left( \tilde{\param}(t_n) \right)}{R_i^{(l)} \left( \tilde{\param}(t_n) \right) + \varepsilon} \right|\\
      &\quad \leq 2 M_1 \left( (n - k) (\oldbigc{e-second-der} / 2 + \oldbigc{sol-der-first-order}) + \frac{(n - k)(n - k - 1)}{2} \oldbigc{der-p-second-term-aux-two} \right) h^2\\
      &\qquad + 2 (n - k) \oldbigc{second-first-grad-over-r-partial-sum} \left( (n - k) (\oldbigc{e-second-der} / 2 + \oldbigc{sol-der-first-order}) + \frac{(n - k)(n - k - 1)}{2} \oldbigc{der-p-second-term-aux-two} \right) h^3\\
      &\qquad + \left( (n - k) (\oldbigc{e-second-der} / 2 + \oldbigc{sol-der-first-order}) + \frac{(n - k)(n - k - 1)}{2} \oldbigc{der-p-second-term-aux-two} \right)^2 h^4.
    \end{aligned}
  \end{equation}
  To prove this, use
\begin{equation*}
\left| a^2 - b^2 - 2 b K h \right| \leq 2 |b| \cdot \left| a - b - K h \right| + 2 |K| \cdot h \cdot \left| a - b - K h \right| + (a - b - K h )^2
\end{equation*}
with
\begin{equation*}
a := \nabla_j E_k \left( \tilde{\param}(t_k) \right),\quad b := \nabla_j E_k \left( \tilde{\param}(t_n) \right),\quad K := \sum_{i = 1}^p \nabla_{i j} E_k \left( \tilde{\param}(t_n) \right) \sum_{l = k}^{n - 1} \frac{\nabla_i E_l\left(\tilde{\param}(t_n)\right)}{R^{(l)}_i\left(\tilde{\param}(t_n)\right) + \varepsilon},
\end{equation*}
and bounding
\begin{align*}
  &\left| a - b - K h \right| \labelrel\leq{eq:a-b-kh-bounding-one} \left( (n - k) (\oldbigc{e-second-der} / 2 + \oldbigc{sol-der-first-order}) + \frac{(n - k)(n - k - 1)}{2} \oldbigc{der-p-second-term-aux-two} \right) h^2,\\
  &|b| \leq M_1,\quad |K| \leq (n - k) \oldbigc{second-first-grad-over-r-partial-sum},
\end{align*}
where~\eqref{eq:a-b-kh-bounding-one} is by Lemma~\ref{lem:grad-e-to-the-nth-point}. \eqref{eq:diff-of-squared-norms} is proven.

We turn to the proof of~\eqref{eq:diff-of-sums-of-squared-norms}. By~\eqref{eq:diff-of-squared-norms} and the triangle inequality
\begin{align*}
  &\left| \sum_{k = 0}^n \rho^{n - k} (1 - \rho) \left( \nabla_j E_k \left( \tilde{\param}(t_k) \right) \right)^2 - R^{(n)}_j \left( \tilde{\param}(t_n) \right)^2 - 2 h P^{(n)}_j \left( \tilde{\param}(t_n) \right) \right|\\
  &\quad \leq (1 - \rho) \sum_{k = 0}^{n} \rho^{n - k} \left( \mathrm{Poly}_1(n - k) h^2 + \mathrm{Poly}_2(n - k) h^3 + \mathrm{Poly}_3(n - k) h^4 \right)\\
  &\quad = (1 - \rho) \sum_{k = 0}^n \rho^{k} \left( \mathrm{Poly}_1(k) h^2 + \mathrm{Poly}_2(k) h^3 + \mathrm{Poly}_3(k) h^4 \right),
\end{align*}
where
\begin{align*}
  &\mathrm{Poly}_1(k) := 2 M_1 \left( k (\oldbigc{e-second-der} / 2 + \oldbigc{sol-der-first-order}) + \frac{k (k - 1)}{2} \oldbigc{der-p-second-term-aux-two} \right) = M_1 \oldbigc{der-p-second-term-aux-two} k^2 + M_1 (\oldbigc{e-second-der} + 2 \oldbigc{sol-der-first-order} - \oldbigc{der-p-second-term-aux-two}) k,\\
  &\mathrm{Poly}_2(k) := 2 k \oldbigc{second-first-grad-over-r-partial-sum} \left( k (\oldbigc{e-second-der} / 2 + \oldbigc{sol-der-first-order}) + \frac{k (k - 1)}{2} \oldbigc{der-p-second-term-aux-two} \right) = \oldbigc{der-p-second-term-aux-two} \oldbigc{second-first-grad-over-r-partial-sum} k^3 + \left( \oldbigc{e-second-der} + 2 \oldbigc{sol-der-first-order} - \oldbigc{der-p-second-term-aux-two} \right) \oldbigc{second-first-grad-over-r-partial-sum} k^2,\\
  &\mathrm{Poly}_3(k) := \left( k (\oldbigc{e-second-der} / 2 + \oldbigc{sol-der-first-order}) + \frac{k (k - 1)}{2} \oldbigc{der-p-second-term-aux-two} \right)^2\\
  &\quad = \frac{\oldbigc{der-p-second-term-aux-two}^2}{4} k^4 + \frac{\oldbigc{der-p-second-term-aux-two}}{2} \left( \oldbigc{e-second-der} + 2 \oldbigc{sol-der-first-order} - \oldbigc{der-p-second-term-aux-two} \right) k^3 + \frac{1}{4} \left( \oldbigc{e-second-der} + 2 \oldbigc{sol-der-first-order} - \oldbigc{der-p-second-term-aux-two} \right)^2 k^2.
\end{align*}
It is left to combine this with
\begin{align*}
  &\sum_{k = 0}^n k \rho^k \leq \sum_{k = 0}^{\infty} k \rho^k = \frac{\rho}{(1 - \rho)^2},\\
  &\sum_{k = 0}^n k^2 \rho^k \leq \sum_{k = 0}^{\infty} k^2 \rho^k = \frac{\rho (1 + \rho)}{(1 - \rho)^3},\\
  &\sum_{k = 0}^n k^3 \rho^k \leq \sum_{k = 0}^{\infty} k^3 \rho^k = \frac{\rho \left( 1 + 4 \rho + \rho^2 \right)}{(1 - \rho)^4},\\
  &\sum_{k = 0}^n k^4 \rho^k \leq \sum_{k = 0}^{\infty} k^4 \rho^k = \frac{\rho \left( 1 + 11 \rho + 11 \rho^2 + \rho^3 \right)}{(1 - \rho)^5}.
\end{align*}
This gives
\begin{align*}
  &\left| \sum_{k = 0}^n \rho^{n - k} (1 - \rho) \left( \nabla_j E_k \left( \tilde{\param}(t_k) \right) \right)^2 - R^{(n)}_j \left( \tilde{\param}(t_n) \right)^2 - 2 h P^{(n)}_j \left( \tilde{\param}(t_n) \right) \right|\\
  &\quad \leq \left( M_1 \oldbigc{der-p-second-term-aux-two} \frac{\rho (1 + \rho)}{\left( 1 - \rho \right)^2} + M_1 \left| \oldbigc{e-second-der} + 2 \oldbigc{sol-der-first-order} - \oldbigc{der-p-second-term-aux-two} \right| \frac{\rho}{1 - \rho}\right) h^2\\
  &\qquad + \left( \oldbigc{der-p-second-term-aux-two} \oldbigc{second-first-grad-over-r-partial-sum} \frac{\rho \left( 1 + 4 \rho + \rho^2 \right)}{(1 - \rho)^3} + \left| \oldbigc{e-second-der} + 2 \oldbigc{sol-der-first-order} - \oldbigc{der-p-second-term-aux-two} \right| \oldbigc{second-first-grad-over-r-partial-sum} \frac{\rho (1 + \rho)}{(1 - \rho)^2} \right) h^3\\
  &\qquad + \left( \frac{\oldbigc{der-p-second-term-aux-two}^2}{4} \cdot \frac{\rho \left( 1 + 11 \rho + 11 \rho^2 + \rho^3 \right)}{(1 - \rho)^4} + \frac{\oldbigc{der-p-second-term-aux-two}}{2} \left| \oldbigc{e-second-der} + 2 \oldbigc{sol-der-first-order} - \oldbigc{der-p-second-term-aux-two} \right| \frac{\rho \left( 1 + 4 \rho + \rho^2 \right)}{(1 - \rho)^3} \right.\\
  &\quad\qquad + \left.\frac{1}{4} \left( \oldbigc{e-second-der} + 2 \oldbigc{sol-der-first-order} - \oldbigc{der-p-second-term-aux-two} \right)^2 \frac{\rho (1 + \rho)}{(1 - \rho)^2} \right) h^4\\
  &\quad \labelrel\leq{eq:diff-of-sums-of-squared-norms-proof-conclusion-one} \left[ M_1 \left| \oldbigc{e-second-der} + 2 \oldbigc{sol-der-first-order} - \oldbigc{der-p-second-term-aux-two} \right| \frac{\rho}{1 - \rho}\right.\\
  &\qquad + \left( M_1 \oldbigc{der-p-second-term-aux-two} + \left| \oldbigc{e-second-der} + 2 \oldbigc{sol-der-first-order} - \oldbigc{der-p-second-term-aux-two} \right| \oldbigc{second-first-grad-over-r-partial-sum} + \frac{\left( \oldbigc{e-second-der} + 2 \oldbigc{sol-der-first-order} - \oldbigc{der-p-second-term-aux-two} \right)^2}{4} \right) \frac{\rho (1 + \rho)}{(1 - \rho)^2}\\
  &\qquad + \left(\oldbigc{der-p-second-term-aux-two} \oldbigc{second-first-grad-over-r-partial-sum} + \frac{\oldbigc{der-p-second-term-aux-two}}{2} \left| \oldbigc{e-second-der} + 2 \oldbigc{sol-der-first-order} - \oldbigc{der-p-second-term-aux-two} \right|\right) \frac{\rho \left( 1 + 4 \rho + \rho^2 \right)}{(1 - \rho)^3}\\
  &\qquad\left. + \frac{\oldbigc{der-p-second-term-aux-two}^2}{4} \cdot \frac{\rho \left( 1 + 11 \rho + 11 \rho^2 + \rho^3 \right)}{(1 - \rho)^4}\right] h^2,
\end{align*}
where in~\eqref{eq:diff-of-sums-of-squared-norms-proof-conclusion-one} we used that $h < 1$. \eqref{eq:diff-of-sums-of-squared-norms} is proven.
\end{proof}

\newcommand{\RhsLemRootsInDenominatorsFirstOrder}{\frac{\oldbigc{firstorderdif}^2 + R^2 \oldbigc{secondorderdif}}{2 R^3 (R + \varepsilon)^2}}
\begin{lemma}\label{lem:roots-in-denominators-first-order}
  Suppose \cref{ass:bounds} holds. Then
\begin{align*}
  &\left| \left( \sqrt{\sum_{k = 0}^n \rho^{n - k} (1 - \rho) \left( \nabla_j E_k \left( \tilde{\param}(t_k) \right) \right)^2} + \varepsilon \right)^{-1} - \left( R^{(n)}_j \left( \tilde{\param}(t_n) \right) + \varepsilon \right)^{-1} \right.\\
  &\qquad\left. + h \frac{P^{(n)}_j \left( \tilde{\param}(t_n) \right)}{\left( R^{(n)}_j \left( \tilde{\param}(t_n) \right) + \varepsilon \right)^2 R^{(n)}_j \left( \tilde{\param}(t_n) \right)} \right| \leq \frac{\oldbigc{firstorderdif}(\rho)^2 + R^2 \oldbigc{secondorderdif}(\rho)}{2 R^3 (R + \varepsilon)^2} h^2.
\end{align*}
\end{lemma}

\begin{proof}
Note that if $a \geq R^2$, $b \geq R^2$, we have
\begin{align*}
  &\left| \frac{1}{\vphantom{\left( \sqrt{b} + \varepsilon \right)^2 \sqrt{b}}\sqrt{a} + \varepsilon} - \frac{1}{\vphantom{\left( \sqrt{b} + \varepsilon \right)^2 \sqrt{b}}\sqrt{b} + \varepsilon} + \frac{a - b}{2 \left( \sqrt{b} + \varepsilon \right)^2 \sqrt{b}} \right|\\
  &\quad = \frac{(a - b)^2}{2 \sqrt{b} \left( \sqrt{b} + \varepsilon \right) \left( \sqrt{a\vphantom{b}}\vphantom{\sqrt{b}} + \varepsilon \right) \left( \sqrt{a\vphantom{b}} + \sqrt{b} \right)} \underbrace{\left\{ \frac{1}{\sqrt{b} + \varepsilon} + \frac{1}{\sqrt{a\vphantom{b}} + \sqrt{b}} \right\}}_{\leq 2 / R}\\
  &\quad \leq \frac{(a - b)^2}{2 R^3 (R + \varepsilon)^2}.
\end{align*}
By the triangle inequality,
\begin{align*}
  &\left| \frac{1}{\vphantom{\left( \sqrt{b} + \varepsilon \right)^2 \sqrt{b}}\sqrt{a} + \varepsilon} - \frac{1}{\vphantom{\left( \sqrt{b} + \varepsilon \right)^2 \sqrt{b}}\sqrt{b} + \varepsilon} + \frac{c}{2 \left( \sqrt{b} + \varepsilon \right)^2 \sqrt{b}} \right| \leq \frac{(a - b)^2}{2 R^3 (R + \varepsilon)^2} + \frac{\left| a - b - c \right|}{2 \left( \sqrt{b} + \varepsilon \right)^2 \sqrt{b}}\\
  &\quad \leq \frac{(a - b)^2}{2 R^3 (R + \varepsilon)^2} + \frac{\left| a - b - c \right|}{2 R \left( R + \varepsilon \right)^2}.
\end{align*}
Apply this with
\begin{align*}
  &a := \sum_{k = 0}^n \rho^{n - k} (1 - \rho) \left( \nabla_j E_k \left( \tilde{\param}(t_k) \right) \right)^2,\\
  &b := R^{(n)}_j \left( \tilde{\param}(t_n) \right)^2,\\
  &c := 2 h P^{(n)}_j \left( \tilde{\param}(t_n) \right)
\end{align*}
and use bounds
\begin{equation*}
  \left| a - b \right| \leq 2 M_1 \oldbigc{e-first-der} \frac{\rho}{1 - \rho} h,\quad\left| a - b - c \right| \leq \oldbigc{secondorderdif}(\rho) h^2
\end{equation*}
by Lemma~\ref{cor:grad-e-squared-to-the-nth-point}.
\end{proof}

We are finally ready to prove \cref{th:local-error-bound}.
\begin{proof}[Proof of \cref{th:local-error-bound}]
  By~\eqref{eq:der-huge-func-first-term-bound} and~\eqref{eq:der-huge-func-second-term-bound}, the first derivative of the function
\begin{equation*}
t \mapsto \left( \frac{\nabla_j E_n \left( \tilde{\param}(t) \right) \left(2 P_j^{(n)}\left(\tilde{\param}(t)\right) + \bar{P}_j^{(n)}\left(\tilde{\param}(t)\right)\right)}{2 \left( R_j^{(n)} \left( \tilde{\param}(t) \right) + \varepsilon \right)^2 R_j^{(n)} \left( \tilde{\param}(t) \right)} - \frac{\sum_{i = 1}^p \nabla_{i j} E_n \left( \tilde{\param}(t) \right) \frac{\nabla_i E_n \left( \tilde{\param}(t) \right)}{R_i^{(n)}\left(\tilde{\param}(t)\right) + \varepsilon}}{2 \left( R_j^{(n)}(\tilde{\param}(t)) + \varepsilon \right)} \right)
\end{equation*}
is bounded in absolute value by a positive constant $\newbigc[nth-step-modified-h-term-der-bound] = \oldbigc{der-huge-func-first-term} + \oldbigc{der-huge-func-second-term}$. By~\eqref{eq:nth-step-modified-equation}, this means
\begin{equation*}
\left| \ddot{\tilde{\paramsc}}_j(t) + \frac{\mathrm{d}}{\mathrm{d} t} \left( \frac{\nabla_j E_n\left(\tilde{\param}(t)\right)}{R^{(n)}_j\left(\tilde{\param}(t)\right) + \varepsilon} \right) \right| \leq \oldbigc{nth-step-modified-h-term-der-bound} h.
\end{equation*}
Combining this with
\begin{equation*}
\left| \tilde{\paramsc}_j(t_{n + 1}) - \tilde{\paramsc}_j(t_n) - \dot{\tilde{\paramsc}}_j\left(t_n^+\right) h - \frac{\ddot{\tilde{\paramsc}}_j\left(t_n^+\right)}{2} h^2 \right| \leq \frac{D_3}{6}
\end{equation*}
by Taylor expansion, we get
\begin{equation}\label{eq:xdiff-minus-der-minus-d-dt-bound}
\begin{aligned}
  &\left| \tilde{\paramsc}_j(t_{n + 1}) - \tilde{\paramsc}_j(t_n) - \dot{\tilde{\paramsc}}_j\left(t_n^+\right) h + \frac{h^2}{2} \cdot \frac{\mathrm{d}}{\mathrm{d} t} \left.\left( \frac{\nabla_j E_n\left(\tilde{\param}(t)\right)}{R^{(n)}_j\left(\tilde{\param}(t)\right) + \varepsilon} \right)\right|_{t = t_n^+} \right|\\
  &\quad \leq \left( \frac{D_3}{6} + \frac{\oldbigc{nth-step-modified-h-term-der-bound}}{2} \right) h^3.
\end{aligned}
\end{equation}

Using
\begin{equation*}
\left| \dot{\tilde{\paramsc}}_j(t_n) + \frac{\nabla_j E_n \left( \tilde{\param}(t_n) \right)}{R_j^{(n)} \left( \tilde{\param}(t_n) \right) + \varepsilon} \right| \leq \oldbigc{paramsc-der-first-order-bound} h
\end{equation*}
with $\newbigc[paramsc-der-first-order-bound]$ defined as
\begin{equation*}
\oldbigc{paramsc-der-first-order-bound} := \frac{M_1 \left( 2 \oldbigc{p-bound} + \oldbigc{p-bar-bound} \right)}{2 (R + \varepsilon)^2 R} + \frac{p M_1 M_2}{2 (R + \varepsilon)^2}
\end{equation*}
by~\eqref{eq:nth-step-modified-equation}, and calculating the derivative, it is easy to show
\begin{equation}\label{eq:local-error-fr-der-bound}
\left|\frac{\mathrm{d}}{\mathrm{d} t} \left.\left( \frac{\nabla_j E_n\left(\tilde{\param}(t)\right)}{R^{(n)}_j\left(\tilde{\param}(t)\right) + \varepsilon} \right)\right|_{t = t_n^+} - \mathrm{FrDer}\right| \leq \oldbigc{frderhbound} h
\end{equation}
for a positive constant $\newbigc[frderhbound]$, where
\begin{align*}
  &\mathrm{FrDer} := \frac{\mathrm{FrDerNum}}{\left(
    R_j^{(n)} \left(\tilde{\param}(t_n)\right) + \varepsilon\right)^2 R_j^{(n)} \left( \tilde{\param}(t_n)\right)}\\
  &\mathrm{FrDerNum} := \nabla_j E_n \left( \tilde{\param}(t_n) \right) \bar{P}_j^{(n)} \left( \tilde{\param}(t_n) \right)\\
  &\qquad - \left( R_j^{(n)} \left( \tilde{\param}(t_n)\right) + \varepsilon \right) R_j^{(n)} \left( \tilde{\param}(t_n)\right) \sum_{i = 1}^p \nabla_{i j} E_n \left( \tilde{\param}(t_n) \right) \frac{\nabla_i E_n \left( \tilde{\param}(t_n) \right)}{R_i^{(n)} \left( \tilde{\param}(t_n)\right) + \varepsilon},\\
  &\oldbigc{frderhbound} := \left\{ \frac{p M_2}{R + \varepsilon} + \frac{M_1^2 M_2 p}{(R + \varepsilon)^2 R} \right\} \oldbigc{paramsc-der-first-order-bound}.
\end{align*}

From~\eqref{eq:xdiff-minus-der-minus-d-dt-bound} and \eqref{eq:local-error-fr-der-bound}, by the triangle inequality
\begin{equation*}
\left| \tilde{\paramsc}_j(t_{n + 1}) - \tilde{\paramsc}_j(t_n) - \dot{\tilde{\paramsc}}_j\left(t_n^+\right) h + \frac{h^2}{2} \mathrm{FrDer} \right| \leq \left( \frac{D_3}{6} + \frac{\oldbigc{nth-step-modified-h-term-der-bound} + \oldbigc{frderhbound}}{2}\right) h^3,
\end{equation*}
which, using~\eqref{eq:nth-step-modified-equation}, is rewritten as
\begin{align*}
&\left| \tilde{\paramsc}_j(t_{n + 1}) - \tilde{\paramsc}_j(t_n) + h \frac{\nabla_j E_n \left( \tilde{\param}(t_n) \right)}{R_j^{(n)} \left( \tilde{\param}(t_n) \right) + \varepsilon} - h^2 \frac{\nabla_j E_n \left( \tilde{\param}(t_n) \right) P_j^{(n)} \left( \tilde{\param}(t_n) \right)}{\left(
      R_j^{(n)} \left(\tilde{\param}(t_n)\right) + \varepsilon\right)^2 R_j^{(n)} \left( \tilde{\param}(t_n)\right)} \right|\\
&\quad \leq \left( \frac{D_3}{6} + \frac{\oldbigc{nth-step-modified-h-term-der-bound} + \oldbigc{frderhbound}}{2}\right) h^3.
\end{align*}

It is left to combine this with Lemma~\ref{lem:roots-in-denominators-first-order}, giving the assertion of the theorem with
\begin{equation*}
\oldbigc{localerrorbound} = \frac{D_3}{6} + \frac{\oldbigc{nth-step-modified-h-term-der-bound} + \oldbigc{frderhbound}}{2} + M_1 \RhsLemRootsInDenominatorsFirstOrder.\qedhere
\end{equation*}
\end{proof}

\section{Numerical Experiments}\label{sec:numerical-experiments}

\subsection{Models}

We use small modifications of Resnet-50 and Resnet-101 implementations in the \verb|torchvision| library for training on CIFAR-10 and CIFAR-100. The first convolution layer \verb|conv1| has $3 \times 3$ kernel, stride 1 and ``same'' padding. Then comes batch normalization, and relu. Max pooling is removed, and otherwise \verb|conv2_x| to \verb|conv5_x| are as described in~\citet{he2016deep} (see Table~1 there) except downsampling is performed by the middle convolution of each bottleneck block, as in version 1.5\footnote{\url{https://catalog.ngc.nvidia.com/orgs/nvidia/resources/resnet_50_v1_5_for_pytorch}}.
After \verb|conv5| there is global average pooling and 10 or 100-way fully connected layer (for CIFAR-10 and CIFAR-100 respectively).

The MLP that we use for showing the closeness of trajectories in \cref{fig:mlp_gelu_bea_trajectories} consists of two fully connected layers, each with 32 units and GeLU activation, followed by a fully-connected layer with 10 units.

In \cref{fig:mlp_gelu_bea_trajectories}, the curves called ``first order'' plot $\bigl\|\param^{(n)} - \tilde{\param}^{(n)}\bigr\|_{\infty}$ and the curves called ``second order'' plot $\bigl\|\param^{(n)} - \tilde{\tilde{\param}}^{(n)}\bigr\|_{\infty}$, where $\param^{(n)}$ is the Adam iteration defined in \cref{def:adam} and
\begin{equation}\label{eq:bea-closeness-precise-iterations}
\begin{aligned}
  &\tilde{\tilde{\paramsc}}^{(n + 1)}_j = \tilde{\tilde{\paramsc}}^{(n)}_j - h A^{(n)}_j\left(\tilde{\tilde{\param}}^{(n)}\right) + h^2 B^{(n)}_j\left(\tilde{\tilde{\param}}^{(n)}\right),\\
  &\tilde{\paramsc}^{(n + 1)}_j = \tilde{\paramsc}^{(n)}_j - h A^{(n)}_j\left(\tilde{\param}^{(n)}\right)
\end{aligned}
\end{equation}
for $A^{(n)}_j(\cdot)$ and $B^{(n)}_j(\cdot)$ as defined in \cref{sec:main-result}, with the same initial point $\param^{(0)} = \tilde{\param}^{(0)} = \tilde{\tilde{\param}}^{(0)}$.

\subsection{Data Augmentation}
We subtract the per-pixel mean and divide by standard deviation, and we use the data augmentation scheme from \citet{lee2015deeply}, following \citet{he2016deep}, section 4.2. During each pass over the training dataset, each $32 \times 32$ initial image is padded evenly with zeros so that it becomes $40 \times 40$, then random crop is applied so that the picture becomes $32 \times 32$ again, and random (probability $0.5$) horizontal (left to right) flip is used.

\subsection{Experiment Details}
\begin{figure}[h]
 \centerline{\includegraphics[width=\linewidth]{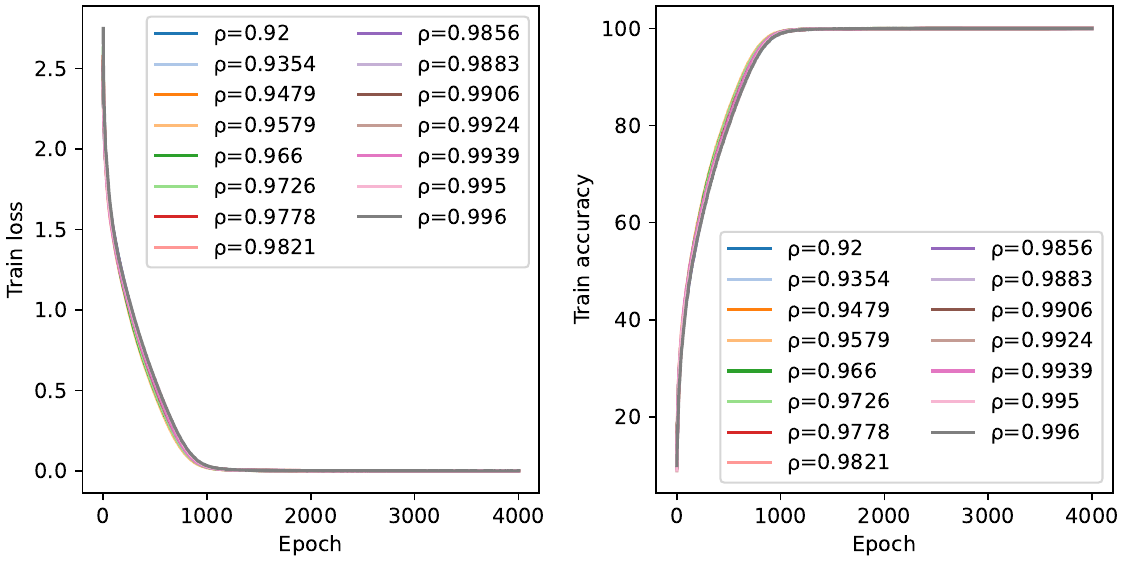}}
  \caption{Train loss and train accuracy curves for full-batch Adam, ResNet-50 on CIFAR-10, $\beta = 0.99$, $\varepsilon = 10^{-8}$, $h = 10^{-4}$.}
  \label{fig:resnet-50-illustration-of-training-dynamics-part-0}
\end{figure}

\begin{figure}[h]
 \centerline{\includegraphics[width=\linewidth]{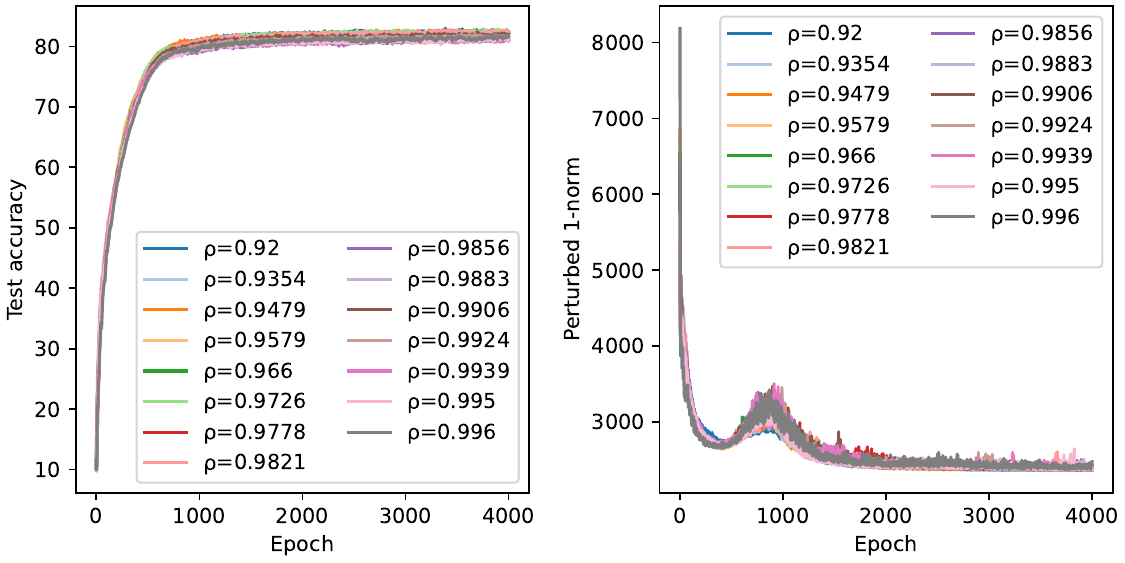}}
  \caption{Test accuracy and $\| \nabla E \|_{1, \varepsilon}$ after each epoch. The setting is the same as in \cref{fig:resnet-50-illustration-of-training-dynamics-part-0}.}
  \label{fig:resnet-50-illustration-of-training-dynamics-part-1}
\end{figure}

In experiments whose results are reported in \cref{fig:resnet-50-rho-increase,fig:resnet-50-beta-increase} of the main paper, we train for a few thousand epochs and stop training when the train accuracy is near-perfect (\cref{fig:resnet-50-illustration-of-training-dynamics-part-0}) and the testing accuracy does not significantly improve (\cref{fig:resnet-50-illustration-of-training-dynamics-part-1}). Therefore, the maximal test accuracies are the final ones reached, and the maximal perturbed one-norms, after excluding the initial fall at the beginning of training, are at peaks of the ``hills'' on the norm curves (\cref{fig:resnet-50-illustration-of-training-dynamics-part-1}).

Since the full dataset does not fit into GPU memory, we divide it into 100 ``ghost batches'' and accumulate the gradients before doing one optimization step. This means that we use ghost batch normalization \citep{hoffer2017train} as opposed to full-dataset batch normalization, similarly to \citet{cohen2021gradient}.

\subsection{Additional Evidence}\label{sec:additional-evidence}

\begin{figure}[h]
  \centerline{\includegraphics[width=\linewidth]{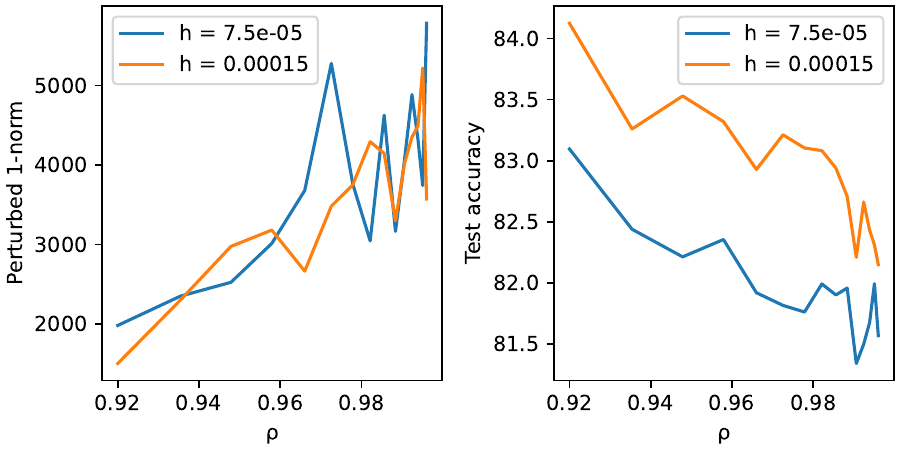}}
  \caption{A simple CNN trained on CIFAR-10 with full-batch Adam, $\beta = 0.99$, $\varepsilon = 10^{-8}$. As $\rho$ increases, the perturbed one-norm rises and the test accuracy falls. Both metrics are calculated as in \cref{fig:resnet-50-rho-increase,fig:resnet-50-beta-increase} of the main paper. All results are averaged across five runs with different initialization seeds.}
  \label{fig:cnn-cifar10-rho-increase}
\end{figure}

\begin{figure}[h]
  \centerline{\includegraphics[width=\linewidth]{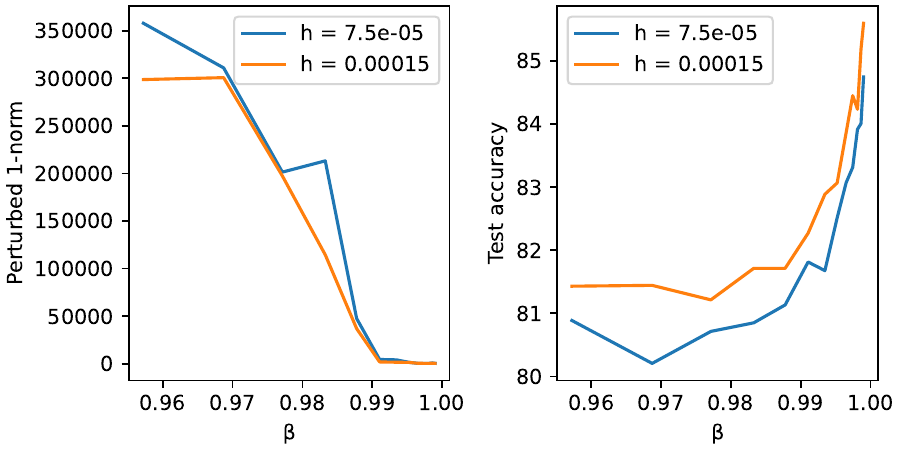}}
  \caption{A simple CNN trained on CIFAR-10 with full-batch Adam, $\rho = 0.999$, $\varepsilon = 10^{-8}$. The perturbed one-norm falls as $\beta$ increases, and the test accuracy rises. Both metrics are calculated as in \cref{fig:resnet-50-rho-increase,fig:resnet-50-beta-increase} of the main paper. All results are averaged across three runs with different initialization seeds.}
  \label{fig:cnn-cifar10-beta-increase}
\end{figure}

\begin{figure}[h]
  \centerline{\includegraphics[width=\linewidth]{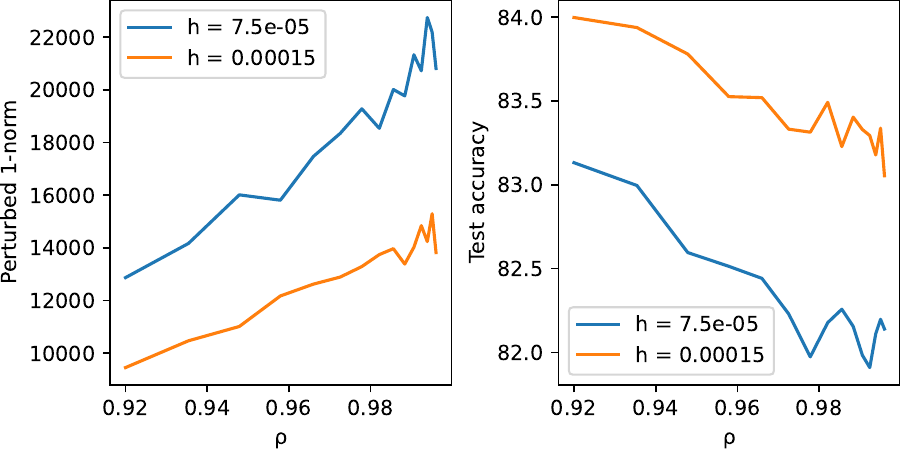}}
  \caption{A vision transformer trained on CIFAR-10 with full-batch Adam. The setting and conclusions are the same as in \cref{fig:cnn-cifar10-rho-increase}.}
  \label{fig:vit-cifar10-rho-increase}
\end{figure}

\begin{figure}[h]
  \centerline{\includegraphics[width=\linewidth]{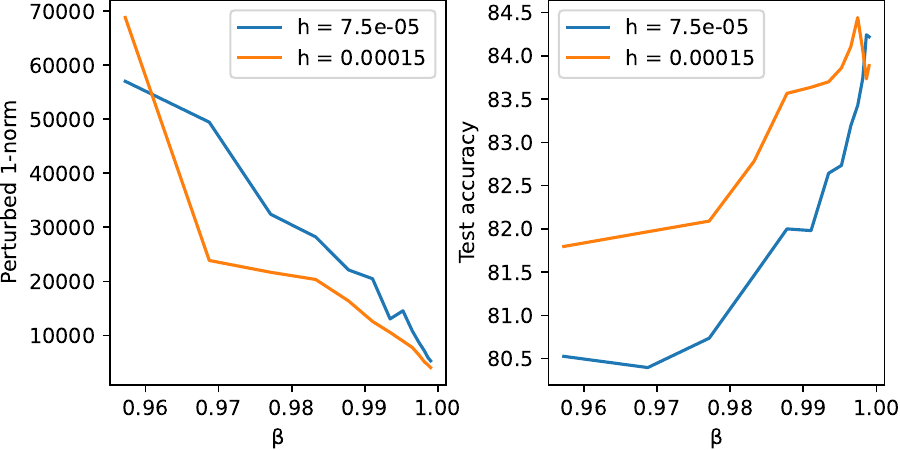}}
  \caption{A vision transformer trained on CIFAR-10 with full-batch Adam. The setting and conclusions are the same as in \cref{fig:cnn-cifar10-beta-increase}.}
  \label{fig:vit-cifar10-beta-increase}
\end{figure}

We provide evidence that the results in \cref{fig:resnet-50-rho-increase,fig:resnet-50-beta-increase} are robust to the change of architectures. In \cref{fig:cnn-cifar10-rho-increase,fig:cnn-cifar10-beta-increase}, we show that the pictures are similar for a simple CNN created by the following code:
\begin{minted}[breaklines]{python}
layers = [
    # First block
    nn.Conv2d(in_channels=3, out_channels=32, kernel_size=3, padding='same'),
    nn.ReLU(),
    nn.Conv2d(in_channels=32, out_channels=32, kernel_size=3, padding='same'),
    nn.ReLU(),
    nn.MaxPool2d(kernel_size=2, stride=2),

    # Second block
    nn.Conv2d(in_channels=32, out_channels=64, kernel_size=3, padding='same'),
    nn.ReLU(),
    nn.Conv2d(in_channels=64, out_channels=64, kernel_size=3, padding='same'),
    nn.ReLU(),
    nn.MaxPool2d(kernel_size=2, stride=2),

    # Third block
    nn.Conv2d(in_channels=64, out_channels=128, kernel_size=3, padding='same'),
    nn.ReLU(),
    nn.Conv2d(in_channels=128, out_channels=128, kernel_size=3, padding='same'),
    nn.ReLU(),
    nn.MaxPool2d(kernel_size=2, stride=2),

    # Flatten and Dense layers
    nn.Flatten(),
    nn.Linear(in_features=128 * 4 * 4, out_features=512),
    nn.ReLU(),
    nn.Linear(in_features=512, out_features=num_classes),
]
return nn.Sequential(*layers)
\end{minted}
In \cref{fig:vit-cifar10-rho-increase,fig:vit-cifar10-beta-increase}, we show that the same conclusions can be made for a Vision Transformer \citep{dosovitskiy2020image,beyer2022better}. In these experiments, we use the SimpleViT architecture from the vit-pytorch library with $4\times4$ patches, 6 transformer blocks with 16 heads, embedding size 512 and MLP dimension of 1024 (following \citet{andriushchenko2023ModernLook}).

\section{Adam with $\varepsilon$ Inside the Square Root: Informal Derivation}\label{sec:informal-derivation}

Our goal is to find such a trajectory $\tilde{\param}(t)$ that
\begin{equation*}
\tilde{\paramsc}_j(t_{n + 1}) = \tilde{\paramsc}_j(t_n) - h \frac{\sum_{k = 0}^n \beta^{n - k} \left(1 - \beta\right) \nabla_j E_k \bigl( \tilde{\param}(t_k) \bigr) / (1 - \beta^{n + 1})}{\sqrt{\sum_{k = 0}^n \rho^{n - k} (1 - \rho) \bigl( \nabla_j E_k \bigl( \tilde{\param}(t_k) \bigr) \bigr)^2 / (1 - \rho^{n + 1}) + \varepsilon}} + O(h^3).
\end{equation*}

\begin{result}\label{lem:mod-adam-paramsc-nth-step-complete}
  For $n \in \left\{ 0, 1, 2, \ldots \right\}$ we have
\begin{equation}\label{eq:mod-adam-paramsc-nth-step-complete}
\begin{aligned}
  &\tilde{\paramsc}_j\left(t_{n + 1}\right) = \tilde{\paramsc}_j(t_n) - h \frac{M_j^{(n)} \left( \tilde{\param}(t_n) \right)}{R_j^{(n)}\left(\tilde{\param}(t_n)\right)}\\
  &\qquad + h^2 \left(\frac{M_j^{(n)} \left( \tilde{\param}(t_n) \right) P_j^{(n)}\left(\tilde{\param}(t_n)\right)}{R_j^{(n)}\left(\tilde{\param}(t_n)\right)^3} - \frac{L_j^{(n)} \left( \tilde{\param}(t_n) \right)}{R_j^{(n)}\left(\tilde{\param}(t_n)\right)}\right) + O \left( h^3 \right).
\end{aligned}
\end{equation}
\end{result}

\begin{proof}[Derivation]
  We take
\begin{equation*}
\tilde{\paramsc}_j\left(t_{n + 1}\right) = \tilde{\paramsc}_j(t_n) - h \frac{M_j^{(n)} \left( \tilde{\param}(t_n) \right)}{R_j^{(n)}\left(\tilde{\param}(t_n)\right)} + O \left( h^2 \right)
\end{equation*}
for granted. Using this and the Taylor series, we can write
\begin{align*}
  &\nabla_j E_k \left( \tilde{\param}(t_{n - 1}) \right)\\
  &\quad = \nabla_j E_k \left( \tilde{\param}(t_n) \right) + \sum_{i = 1}^p \nabla_{i j} E_k \left( \tilde{\param}(t_n) \right) \left\{ \tilde{\paramsc}_i(t_{n - 1}) - \tilde{\paramsc}_i(t_{n}) \right\} + O \left( h^2 \right)\\
  &\quad = \nabla_j E_k \left( \tilde{\param}(t_n) \right) + h \sum_{i = 1}^p \nabla_{i j} E_k \left( \tilde{\param}(t_n) \right) \frac{M_j^{(n - 1)} \left( \tilde{\param}(t_{n - 1}) \right)}{R_j^{(n - 1)}\left(\tilde{\param}(t_{n - 1})\right)} + O \left( h^2 \right)\\
  &\quad = \nabla_j E_k \left( \tilde{\param}(t_n) \right) + h \sum_{i = 1}^p \nabla_{i j} E_k \left( \tilde{\param}(t_n) \right) \frac{M_j^{(n - 1)} \left( \tilde{\param}(t_{n}) \right)}{R_j^{(n - 1)}\left(\tilde{\param}(t_{n})\right)} + O \left( h^2 \right),
\end{align*}
where in the last equality we just replaced $t_{n - 1}$ with $t_n$ in the $h$-term since it only affects higher-order terms. Now doing this again for step $n - 1$ instead of step $n$, we will have
\begin{align*}
  &\nabla_j E_k \left( \tilde{\param}(t_{n - 2}) \right)\\
  &\quad = \nabla_j E_k \left( \tilde{\param}(t_{n - 1}) \right) + h \sum_{i = 1}^p \nabla_{i j} E_k \left( \tilde{\param}(t_{n - 1}) \right) \frac{M_j^{(n - 2)} \left( \tilde{\param}(t_{n - 1}) \right)}{R_j^{(n - 2)}\left(\tilde{\param}(t_{n - 1})\right)} + O \left( h^2 \right)\\
  &\quad = \nabla_j E_k \left( \tilde{\param}(t_{n - 1}) \right) + h \sum_{i = 1}^p \nabla_{i j} E_k \left( \tilde{\param}(t_{n - 1}) \right) \frac{M_j^{(n - 2)} \left( \tilde{\param}(t_{n}) \right)}{R_j^{(n - 2)}\left(\tilde{\param}(t_{n})\right)} + O \left( h^2 \right),
\end{align*}
where in the last equality we again replaced $t_{n - 1}$ with $t_n$ since it only affects higher-order terms. Proceeding like this and adding the resulting equations, we have for $n \in \left\{ 0, 1, \ldots \right\}$, $k \in \left\{ 0, \ldots, n - 1 \right\}$ that
\begin{align*}
  &\nabla_j E_k \left( \tilde{\param}(t_k) \right)\\
  &\quad = \nabla_j E_k \left( \tilde{\param}(t_n) \right) + h \sum_{i = 1}^p \nabla_{i j} E_k \left( \tilde{\param}(t_n) \right) \sum_{l = k}^{n - 1} \frac{M_i^{(l)} \left( \tilde{\param}(t_n) \right)}{R_i^{(l)} \left( \tilde{\param}(t_n) \right)} + O \left( h^2 \right),
\end{align*}
where we ignored the fact that $n - k$ is not bounded (we will get away with this because of exponential averaging). Hence, taking the square of this formal power series,
\begin{align*}
  &\rho^{n - k} (1 - \rho) \left( \nabla_j E_k \left( \tilde{\param}(t_k) \right) \right)^2 = \rho^{n - k} (1 - \rho) \left( \nabla_j E_k \left( \tilde{\param}(t_n) \right) \right)^2\\
  &\qquad + h \cdot 2 \rho^{n - k} (1 - \rho) \nabla_j E_k \left( \tilde{\param}(t_n) \right) \sum_{i = 1}^p \nabla_{i j} E_k \left( \tilde{\param}(t_n) \right) \sum_{l = k}^{n - 1} \frac{M_i^{(l)} \left( \tilde{\param}(t_n) \right)}{R_i^{(l)} \left( \tilde{\param}(t_n) \right)} + O \left( h^2 \right).
\end{align*}

Summing up over $k$, we have
\begin{equation*}
\frac{1}{1 - \rho^{n + 1}} \sum_{k = 0}^n \rho^{n - k} (1 - \rho) \left( \nabla_j E_k \left( \tilde{\param}(t_k) \right) \right)^2 + \varepsilon = R_j^{(n)}\left(\tilde{\param}(t_n)\right)^2 + 2 h P_j^{(n)}\left(\tilde{\param}(t_n)\right) + O \left( h^2 \right),
\end{equation*}
which, using the expression for the inverse square root $\left(\sum_{r = 0}^\infty a_r h^r\right)^{- 1 / 2}$ of a formal power series $\sum_{r = 0}^\infty a_r h^r$, gives us
\begin{align*}
  &\left(\sqrt{\frac{1}{1 - \rho^{n + 1}} \sum_{k = 0}^n \rho^{n - k} (1 - \rho) \left( \nabla_j E_k \left( \tilde{\param}(t_k) \right) \right)^2 + \varepsilon}\right)^{-1}\\
  &\quad = \frac{1}{R_j^{(n)}\left(\tilde{\param}(t_n)\right)} - h \frac{P_j^{(n)}\left(\tilde{\param}(t_n)\right)}{R_j^{(n)}\left(\tilde{\param}(t_n)\right)^3} + O \left( h^2 \right).
\end{align*}

Similarly,
\begin{align*}
  &\frac{1}{1 - \beta^{n + 1}} \sum_{k = 0}^n (1 - \beta) \beta^{n - k} \nabla_j E_k \left( \tilde{\param}(t_k) \right) = \frac{1}{1 - \beta^{n + 1}} \sum_{k = 0}^n (1 - \beta) \beta^{n - k} \nabla_j E_k \left( \tilde{\param}(t_n) \right)\\
  &\qquad + \frac{h}{1 - \beta^{n + 1}} \sum_{k = 0}^n (1 - \beta) \beta^{n - k} \sum_{i = 1}^p \nabla_{i j} E_k \left( \tilde{\param}(t_n) \right) \sum_{l = k}^{n - 1} \frac{M_i^{(l)} \left( \tilde{\param}(t_n) \right)}{R_i^{(l)} \left( \tilde{\param}(t_n) \right)} + O\left(h^2\right)\\
  &\quad = M_j^{(n)}\left(\tilde{\param}(t_n)\right) + h L_j^{(n)} \left( \tilde{\param}(t_n) \right) + O\left(h^2\right).
\end{align*}

We conclude
\begin{align*}
  &\tilde{\paramsc}_j\left(t_{n + 1}\right) = \tilde{\paramsc}_j(t_n) - h \left( M_j^{(n)}\left(\tilde{\param}(t_n)\right) + h L_j^{(n)} \left( \tilde{\param}(t_n) \right) + O \left( h^2 \right)\right) \\
  &\qquad \times \left( \frac{1}{R_j^{(n)}\left(\tilde{\param}(t_n)\right)} - h \frac{P_j^{(n)}\left(\tilde{\param}(t_n)\right)}{R_j^{(n)}\left(\tilde{\param}(t_n)\right)^3} + O \left( h^2 \right) \right) + O \left( h^3 \right)\\
  &\quad = \tilde{\paramsc}_j(t_n) - h \frac{M_j^{(n)} \left( \tilde{\param}(t_n) \right)}{R_j^{(n)}\left(\tilde{\param}(t_n)\right)}\\
  &\qquad + h^2 \left(\frac{M_j^{(n)} \left( \tilde{\param}(t_n) \right) P_j^{(n)}\left(\tilde{\param}(t_n)\right)}{R_j^{(n)}\left(\tilde{\param}(t_n)\right)^3} - \frac{L_j^{(n)} \left( \tilde{\param}(t_n) \right)}{R_j^{(n)}\left(\tilde{\param}(t_n)\right)}\right) + O \left( h^3 \right).\qedhere
\end{align*}
\end{proof}

\begin{result}\label{lem:mod-adam-nth-step-modified-equation}
  For $t_n \leq t < t_{n + 1}$, the modified equation is~\eqref{eq:mod-adam-nth-step-modified-equation}.
\end{result}

\begin{proof}[Derivation]
Assume that the modified flow for $t_n \leq t < t_{n + 1}$ satisfies $\dot{\tilde{\param}} = \tilde{\bof} \left( \tilde{\param}(t) \right)$ where
\begin{equation*}
\tilde{\bof}(\param) = \bof(\param) + h \bof_1(\param) + O\left(h^2\right).
\end{equation*}

By Taylor expansion, we have
\begin{equation}\label{eq:mod-adam-nth-step-taylor-expansion}
\begin{aligned}
  &\tilde{\param}\left(t_{n + 1}\right) = \tilde{\param}(t_n) + h \dot{\tilde{\param}}\left(t_n^+\right) + \frac{h^2}{2} \ddot{\tilde{\param}}\left(t_n^+\right) + O\left(h^3\right)\\
  &\quad = \tilde{\param}(t_n) + h \left[ \bof \left( \tilde{\param}(t_n) \right) + h \bof_1 \left( \tilde{\param}(t_n) \right) + O \left( h^2 \right) \right]\\
  &\qquad + \frac{h^2}{2} \left[ \nabla \bof \left( \tilde{\param}(t_n) \right) \bof \left( \tilde{\param}(t_n) \right) + O(h) \right] + O \left( h^3 \right)\\
  &\quad = \tilde{\param}(t_n) + h \bof \left( \tilde{\param}(t_n) \right) + h^2 \left[ \bof_1 \left( \tilde{\param}(t_n) \right) + \frac{\nabla \bof \left( \tilde{\param}(t_n) \right) \bof \left( \tilde{\param}(t_n) \right)}{2} \right] + O \left( h^3 \right).
\end{aligned}
\end{equation}

Using Lemma~\ref{lem:mod-adam-paramsc-nth-step-complete} and equating the terms before the corresponding powers of $h$ in~\eqref{eq:mod-adam-paramsc-nth-step-complete} and~\eqref{eq:mod-adam-nth-step-taylor-expansion}, we obtain
\begin{equation}\label{eq:mod-adam-nth-step-f-breakdown}
\begin{aligned}
  &f_j(\param) = - \frac{M_j^{(n)}(\param)}{R^{(n)}_j\left(\param\right)},\\
  &f_{1, j}(\param) = - \frac{1}{2} \sum_{i = 1}^p \nabla_i f_j(\param) f_i(\param) + \frac{M_j^{(n)} \left( \param \right) P_j^{(n)}\left(\param\right)}{R_j^{(n)}\left(\param\right)^3} - \frac{L_j^{(n)} \left( \param \right)}{R_j^{(n)}\left(\param\right)}.
\end{aligned}
\end{equation}

It is left to find $\nabla_i f_j(\param)$. Using
\begin{align*}
  &\nabla_i R_j^{(n)}(\param) = \frac{\sum_{k = 0}^n \rho^{n - k} (1 - \rho) \nabla_{i j} E_k(\param) \nabla_j E_k(\param)}{\left( 1 - \rho^{n + 1} \right) R_j^{(n)}(\param)},\\
  &\nabla_i M_j^{(n)}(\param) = \frac{\sum_{k = 0}^n \beta^{n - k} (1 - \beta) \nabla_{i j} E_k(\param)}{1 - \beta^{n + 1}}
\end{align*}
we have
\begin{align*}
  &\nabla_i \left( - \frac{M_j^{(n)}(\param)}{R_j^{(n)}\left(\param\right)} \right)\\
  &\quad = - \frac{\frac{R_j^{(n)}(\param)^2}{1 - \beta^{n + 1}} \sum_{k = 0}^n \beta^{n - k} (1 - \beta) \nabla_{i j} E_k(\param) - \frac{M_j^{(n)}(\param)}{1 - \rho^{n + 1}} \sum_{k = 0}^n \rho^{n - k} (1 - \rho) \nabla_{i j} E_k(\param) \nabla_j E_k(\param)}{R_j^{(n)}\left(\param\right)^3}\\
  &\quad = - \frac{\sum_{k = 0}^n \beta^{n - k} (1 - \beta) \nabla_{i j} E_k(\param)}{\left( 1 - \beta^{n + 1} \right) R_j^{(n)}(\param)} + \frac{M_j^{(n)}(\param) \sum_{k = 0}^n \rho^{n - k} (1 - \rho) \nabla_{i j} E_k(\param) \nabla_j E_k(\param)}{\left( 1 - \rho^{n + 1} \right) R_j^{(n)}\left(\param\right)^3}
\end{align*}

Inserting this into~\eqref{eq:mod-adam-nth-step-f-breakdown} concludes the proof.
\end{proof}

\end{document}